\PassOptionsToPackage{table}{xcolor}
\RequirePackage{fix-cm}
\documentclass[arxiv]{meowreport}
\usepackage{newtxtext}
\ifPDFTeX
  \usepackage[scaled=0.94]{helvet}
\else
\fi
\usepackage{meow-style}
\setcitestyle{authoryear,round,citesep={;},aysep={,},yysep={;}}

\usepackage{amsmath,amsfonts,bm}

\def\eqref#1{\textup{(\ref{#1})}}

\def\1{\bm{1}}

\DeclareMathAlphabet{\mathsfit}{\encodingdefault}{\sfdefault}{m}{sl}
\SetMathAlphabet{\mathsfit}{bold}{\encodingdefault}{\sfdefault}{bx}{n}

\newcommand{\E}{\mathbb{E}}

\newcommand{\KL}{D_{\mathrm{KL}}}

\usepackage{amsmath,amssymb}
\usepackage{graphicx}
\usepackage{adjustbox}
\usepackage{booktabs}
\usepackage{tabularx}
\usepackage{xcolor}
\usepackage{hyperref}
\hypersetup{hidelinks}
\usepackage{url}
\usepackage{comment}
\usepackage{wrapfig}
\usepackage{algorithm}
\usepackage{algpseudocode}
\usepackage{placeins}
\usepackage{needspace}
\usepackage[bottom]{footmisc}

\definecolor{tableSectionTint}{HTML}{F3F0F6}
\definecolor{tablePriorTint}{HTML}{FAF8FC}
\definecolor{tableOursTint}{HTML}{EBE2F5}

\definecolor{metricFDrSixTint}{RGB}{139,236,157}
\definecolor{metricFDrHeldoutTint}{RGB}{209,246,198}
\definecolor{metricFIDTint}{RGB}{255,229,128}
\newcommand{\MetricHighlight}[2]{%
  \begingroup\setlength{\fboxsep}{1pt}%
  \colorbox{#1}{\textcolor{black}{#2}}\endgroup}

\AddToHook{env/table/begin}{\setlength{\belowcaptionskip}{6pt}}
\AddToHook{env/table*/begin}{\setlength{\belowcaptionskip}{6pt}}

\makeatletter
\newcommand{\FinishWrapTable}{%
  \par
  \ifnum\c@WF@wrappedlines>\@ne
    \@tempdima=\baselineskip
    \multiply\@tempdima by \c@WF@wrappedlines
    \advance\@tempdima by -\baselineskip
    \vspace*{\@tempdima}%
  \fi
  \WFclear
}
\makeatother

\meowtitle{Unifying Distributional Training for One-Step \\ Visual Generation}

\meowauthors{\texorpdfstring{%
  Chi Zhang\textsuperscript{1}$^*$, Shi Haoyang\textsuperscript{1,2}$^*$,
  Yueyi Liu\textsuperscript{1,3}$^*$, Ruichuan An\textsuperscript{4},
  Junkang Zhou\textsuperscript{5}, Chang Li\textsuperscript{6}\\[3pt]
  Xiuyuan Lu\textsuperscript{2,7}, Yichi Zhang\textsuperscript{1},
  Bo Wang\textsuperscript{1}, Yuhang Wu\textsuperscript{1},
  Sen Cui\textsuperscript{1}, Miao Liu\textsuperscript{1}$^\dagger$%
}{Chi Zhang, Shi Haoyang, Yueyi Liu, Ruichuan An, Junkang Zhou, Chang Li,
  Xiuyuan Lu, Yichi Zhang, Bo Wang, Yuhang Wu, Sen Cui, Miao Liu}}
\meowaffiliation{%
    \mbox{\textsuperscript{1}Tsinghua University}\quad
    \mbox{\textsuperscript{2}Fudan University}\quad
    \mbox{\textsuperscript{3}Xi'an Jiaotong University}\quad
    \mbox{\textsuperscript{4}Peking University}\quad
    \mbox{\textsuperscript{5}Zhejiang University}\quad
    \mbox{\textsuperscript{6}University of Science and Technology of China}\quad
    \mbox{\textsuperscript{7}University of California, Berkeley}
}
\meowinstitutionmark{%
  \includegraphics[trim=465bp 86bp 492bp 73bp,clip,
    height=\meowInstitutionMarkHeight]{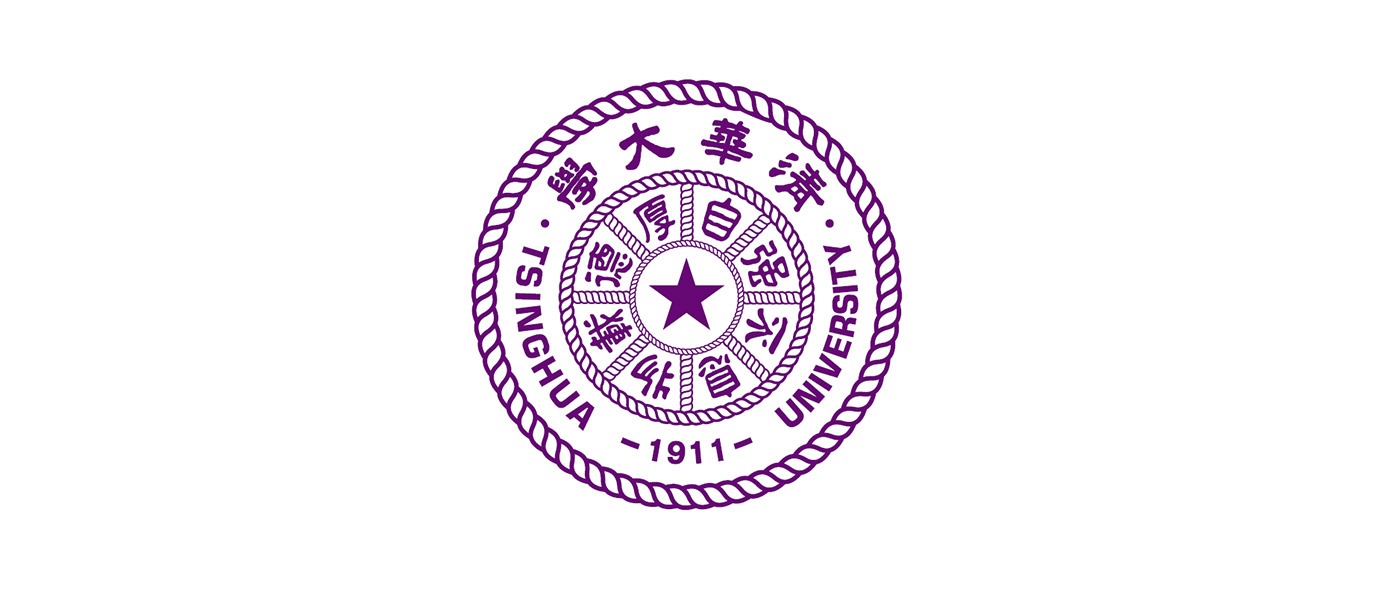}%
}
\meowdate{\strut}
\meowlinks{%
  \meowlink{\raisebox{-0.1em}{\includegraphics[height=0.82em]{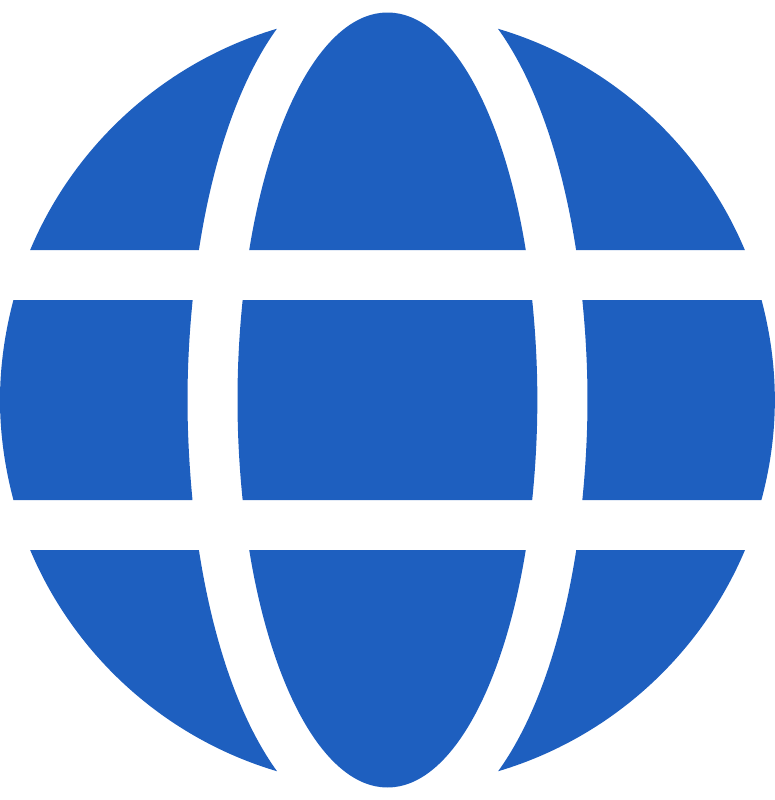}}\enspace Project Page}{https://shihaoyang0423.github.io/MGFlow-website/}
  \hspace{1.8em}
  \meowlink{\raisebox{-0.1em}{\includegraphics[height=0.82em]{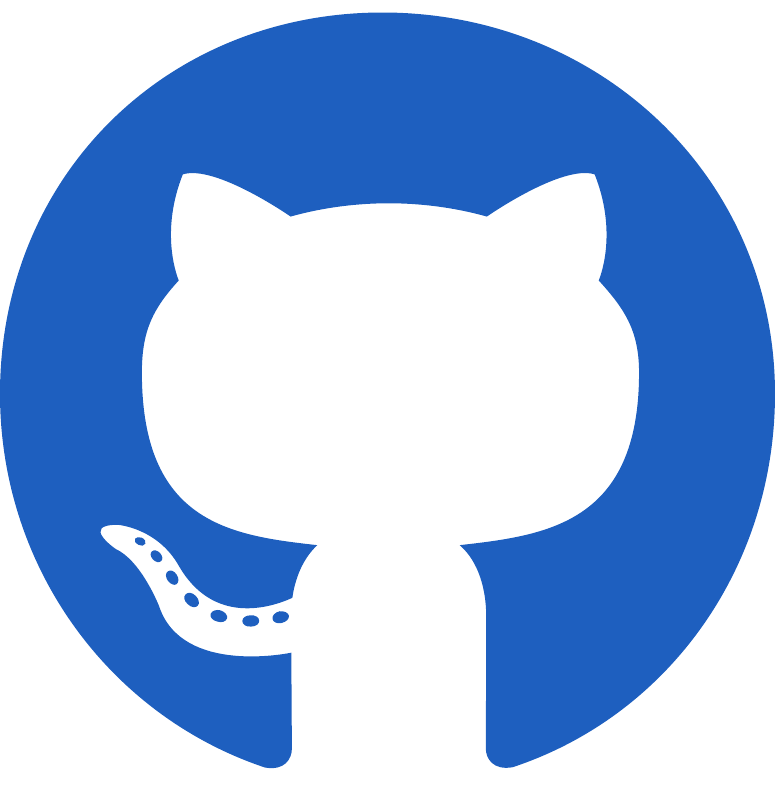}}\enspace Code}{https://github.com/shihaoyang0423/MGFlow}%
}

\begin{document}

\makemeowtitle
\begingroup
  \renewcommand{\thefootnote}{\fnsymbol{footnote}}
  \footnotetext[1]{Equal contribution.\qquad $^\dagger$Corresponding author.}
  \footnotetext[0]{Contact: \href{mailto:imzc.2004@gmail.com}{imzc.2004@gmail.com},
  \href{mailto:miaoliu@mail.tsinghua.edu.cn}{miaoliu@mail.tsinghua.edu.cn}}
\endgroup

\begin{meowteaser}
    \centering
    \captionsetup{justification=centering,singlelinecheck=false}
    \vspace{0pt}
    \includegraphics[width=0.80\linewidth]{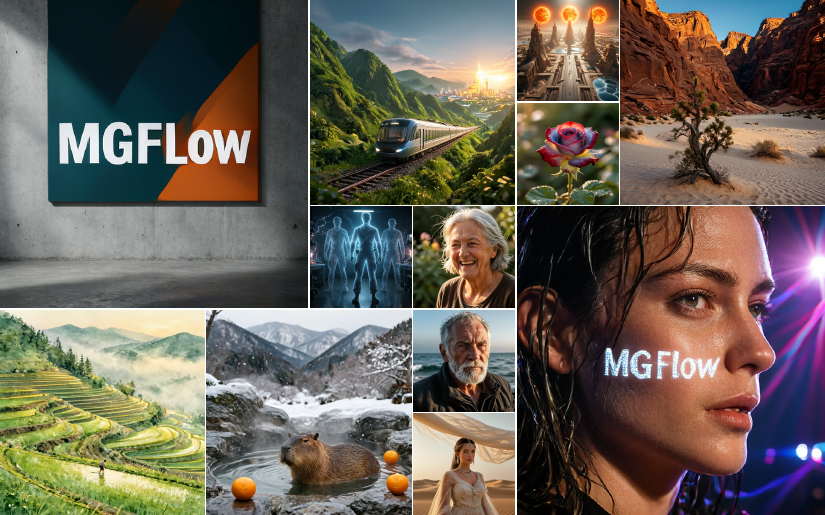}
    \vspace{0pt}
    \caption{\small\textbf{One-step text-to-image samples.}
    FLUX.2 [klein] 4B post-trained with MGFlow.}
    \label{fig:gallery}
\end{meowteaser}

\begin{abstract}
\emph{Distributional training} provides collective supervision for one-step visual generation by matching real and generated features in frozen representation spaces. We introduce \emph{a unified theoretical framework} that separates distribution modeling from matching discrepancy and connects global objectives to pointwise feature updates through Wasserstein gradient flow. Under this framework, FD-Loss and Gaussian-kernel Drifting are recovered through Gaussian optimal transport and kernel-density-based KL matching, respectively.
The framework motivates \textbf{MGFlow}, which models feature distributions with Gaussian mixtures at an adjustable granularity between global moments and sample-based representations. MGFlow supports both optimal transport and score-based matching, and couples mass-constrained sample assignment with paired component updates to address mode collapse that mixture expressivity alone does not resolve. 
On ImageNet $256\times256$, MGFlow substantially surpasses the FD-Loss baseline, achieving state-of-the-art results with \textbf{1.45} FDr$^6$ on pMF-H and \textbf{1.64} on JiT-H. For text-to-image generation, MGFlow post-trains FLUX.2 [klein] 4B into a one-step generator that outperforms the original four-step model on both GenEval and PickScore.\par
\begingroup
  \centering
  \urlstyle{same}
  \textcolor{purple}{Project Page: \nolinkurl{https://shihaoyang0423.github.io/MGFlow-website/}}\par
\endgroup

\end{abstract}

\section{Introduction}

One-step visual generation aims to synthesize high-quality images with a single network evaluation, avoiding iterative refinement at inference. Existing approaches include learning consistent trajectory endpoints~\citep{song2023consistency,song2024ict,kim2024ctm,luo2023lcm,geng2025ect,lu2025scm}, average velocities~\citep{frans2025shortcut,geng2025meanflow,geng2026imf,lu2026pmf}, or distilling pretrained diffusion models~\citep{salimans2022progressive,yin2024dmd,sauer2024ladd,zhou2024sid,yin2024dmd2}.
A prominent recent direction is \emph{distributional training} in frozen representation spaces~\citep{yang2026fdloss,deng2026drifting,feng2026rdm}. Rather than assigning a fixed target image to each output, these methods compare collections of real and generated features and obtain \emph{collective supervision} from their distributional mismatch. Each feature's gradient depends on shared statistics or interactions with other features, rather than only on an individual reconstruction target. This approach raises two fundamental questions: \emph{how should feature distributions be modeled, and how should their mismatch guide learning?}

We introduce \emph{a unified theoretical framework for distributional training} that separates the distribution model from the matching discrepancy. Sampling and frozen encoders determine the feature populations being compared; the distribution model and matching discrepancy determine what distributional information is recorded and how it is aligned. Wasserstein gradient flow (WGF)~\citep{jordan1998variational,peyre2019computational} acts as the theoretical bridge that relates the distributional discrepancy to a pointwise velocity. This establishes a common formulation for global moment losses and local sample interactions, \emph{with existing methods recovered as specific model--discrepancy choices.}

For example, a Gaussian distribution model with the $W_2$ optimal transport (OT) distance recovers \emph{FD-Loss}~\citep{yang2026fdloss}. Let $p_{\text G}$ and $q_{\text G}$ be Gaussians with the respective means and covariances of the real and generated features. FD-Loss optimizes the Fr{\'e}chet loss:
\begin{align}
\mathcal{D}_{\text{FD}}(q_{\text G},p_{\text G})
= W_2^2(q_{\text G},p_{\text G}) = \|\mu_q-\mu_p\|_2^2 + \operatorname{Tr}\!\left( \Sigma_q+\Sigma_p - 2\left(\Sigma_q^{1/2}\Sigma_p\Sigma_q^{1/2}\right)^{1/2} \right).
\label{eq:intro_fd_objective}
\end{align}
Thus, the Fr\'echet discrepancy between global feature statistics is a Gaussian transport cost~\citep{dowson1982frechet,peyre2019computational}. At the population level, differentiating through the moments moves features along the associated affine transport field. The Gaussian model determines which statistics are retained, while the OT discrepancy determines how they guide the movement of generated features.

A sample-based density model with Kullback--Leibler (KL) divergence~\citep{kullback1951information} instead recovers \emph{Gaussian-kernel Drifting}. Drifting~\citep{deng2026drifting} is specified through kernel-weighted attraction and repulsion. For real and generated feature distributions $p$ and $q_\theta$, a positive kernel $k$, and $r\in\{p,q_\theta\}$, its mean-shift and update fields are defined respectively as
\begin{align}
a_r(z)
= \frac{\mathbb{E}_{y\sim r}[k(z,y)(y-z)]}
{\mathbb{E}_{y\sim r}[k(z,y)]},
\quad
v_{\text{drift}}(z)
= a_p(z)-a_{q_\theta}(z).
\label{eq:intro_drifting_field}
\end{align}
The generator regresses samples toward detached targets shifted by this field $v_\text{drift}$. For a Gaussian kernel of bandwidth $h$, each mean shift satisfies $a_r(z)=h^2\nabla_z\log r_h(z)$, where $r_h$ is the Gaussian kernel density estimate (KDE) of $r$~\citep{lai2026unifieddrifting}. Their difference is therefore the density-level KL velocity evaluated at those KDEs, up to a bandwidth-dependent scale~\citep{cao2026gradientflowdrifting}.

Built on top of this theoretical framework, we propose \textbf{Mixture Gradient Flow (MGFlow)}. To refine the modeling granularity while capturing global correlations, MGFlow uses Gaussian mixtures (GMs), which retain componentwise means and covariances at an adjustable resolution between global moment matching and sample-centered KDE. We develop both OT-based and score-based matching for this representation.
Crucially, we note that an expressive modeling does not itself ensure mode coverage or correct transportation~\citep{wenliang2020blindness}: naive posterior assignment and transport may suffer from weight mismatch and mode collapse. MGFlow therefore couples mass-constrained sample assignment with paired component matching, using reference weights to control the mass of each generated component and a shared component correspondence to guide features toward the matched reference components.

We evaluate MGFlow under the same sampling and feature encoder settings as FD-Loss. On ImageNet~\citep{russakovsky2015imagenet} $256\times256$, MGFlow achieves a state-of-the-art $\text{FDr}^6$~\mbox{\citep{yang2026fdloss}} score of $\mathbf{1.45}$ on pMF-H~\citep{lu2026pmf} and $\mathbf{1.64}$ on JiT-H~\citep{li2026jit}, improving over the FD-Loss baseline with \textbf{23\%} and \textbf{38\%} margins, respectively.  For text-to-image generation, MGFlow post-trains FLUX.2 [klein] 4B~\citep{blackforestlabs2026flux2klein} for one-step inference, reaching $\mathbf{0.900}$ GenEval~\citep{ghosh2023geneval} and $\mathbf{21.98}$ PickScore~\citep{kirstain2023pick}, outperforming all previous methods. Together with comparisons across model--discrepancy combinations, these results show how our novel formulation leads to effective alternatives to existing training prescriptions.

\label{sec:method}

\begin{figure}[t]
  \centering
  \includegraphics[width=\linewidth]{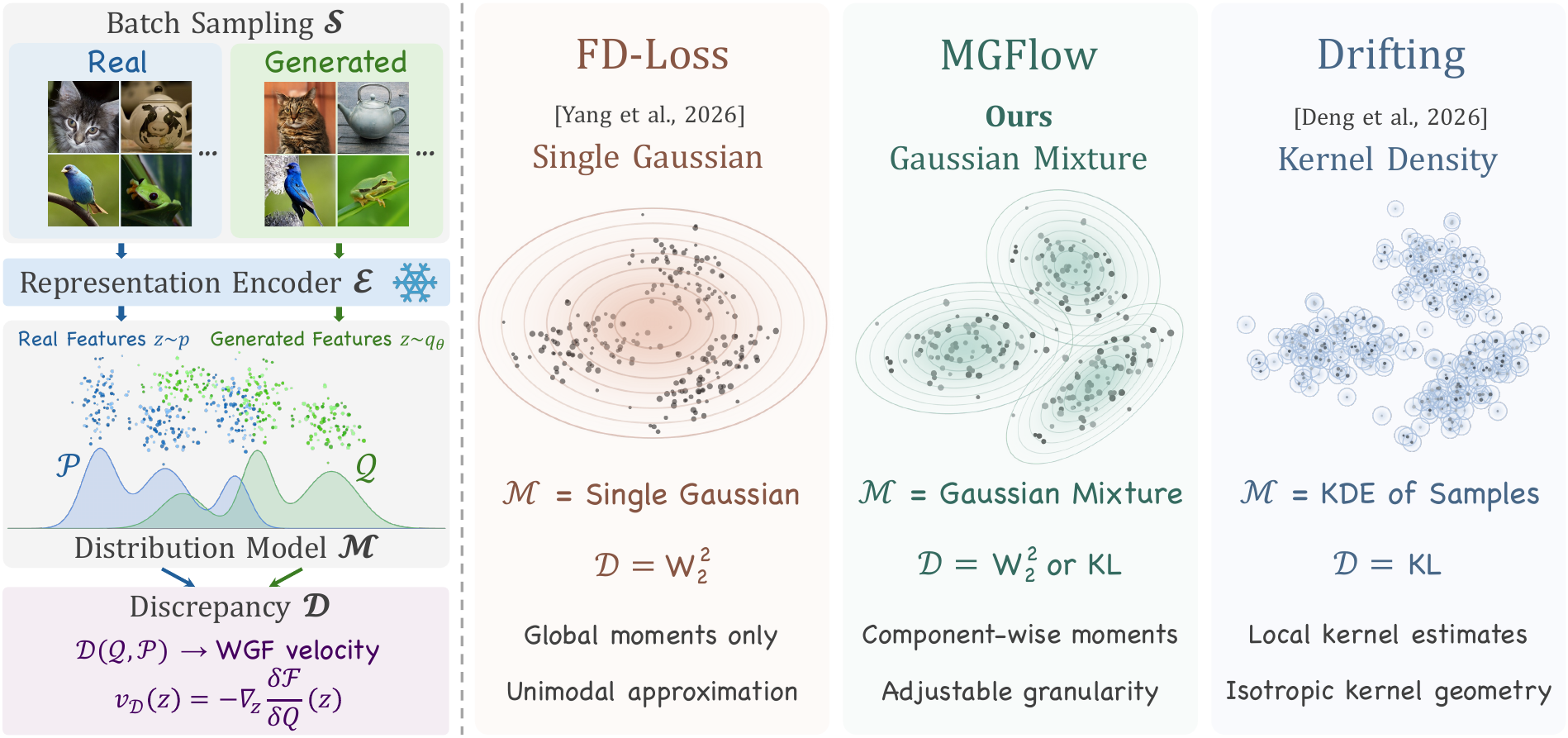}
  \vspace{-5mm}
  \caption{\small\textbf{A unified view of distributional training.}
  Left: sampling $\mathcal S$, encoding $\mathcal E$, distribution modeling
  $\mathcal M$, and matching discrepancy $\mathcal D$ define the paradigm. Right: FD-Loss, MGFlow, and Gaussian-kernel Drifting use
  Gaussian, GM, and KDE representations, respectively. Their matching
  objectives induce feature-space velocity fields through the Wasserstein
  gradient flow of their respective distributional objectives.}
  \label{fig:teaser}
\end{figure}

\section{A Unified View of Distributional Training}
\label{sec:method-unified}

\begin{wraptable}{r}{0.45\textwidth}
  \centering
  \setlength{\belowcaptionskip}{2pt}
  \vspace{-10pt}
  \caption{\small\textbf{Distributional training} with distribution models and discrepancies. All methods train pMF-B with Inception features for 10 epochs. All methods use the aligned recipe in Appendix~\ref{app:short-comparison}.}
  \label{tab:distribution-model-comparison}
  \begingroup
  \fontsize{8.5}{10}\selectfont
  \setlength{\tabcolsep}{2pt}
  \renewcommand{\arraystretch}{1.18}
  \renewcommand{\tabularxcolumn}[1]{m{#1}}
  \begin{adjustbox}{Clip=0pt 0pt 0pt 0pt}
  \begin{tabularx}{\linewidth}{@{}l<{\hspace{2pt}}>{\raggedright\arraybackslash}X@{\hspace{2pt}}r@{}}
    \toprule
    Model & Method & $\text{FDr}^6\!\downarrow$ \\
    \midrule
    \rowcolor{tableSectionTint}
    \multicolumn{3}{@{}l}{\fontsize{7.5}{9}\selectfont\itshape OT-based ($W_2$)} \\
    Gaussian & FD-Loss~\citep{yang2026fdloss} & 13.05 \\
    \rowcolor{tableOursTint}
    GM\, $K=4$ & MGFlow-$W_2$ & \textbf{11.33} \\
    Sample-based & W-Flow~\citep{han2026wflow} & 12.58 \\
    \midrule
    \rowcolor{tableSectionTint}
    \multicolumn{3}{@{}l}{\fontsize{7.5}{9}\selectfont\itshape Score-based (KL)} \\
    Gaussian & Gaussian KL & 12.73 \\
    \rowcolor{tableOursTint}
    GM\, $K=16$ & MGFlow-KL & \textbf{11.32} \\
    Sample-based & Gaussian-kernel Drifting\newline
      \citep{deng2026drifting} & 12.29 \\
    \bottomrule
  \end{tabularx}
  \end{adjustbox}
  \endgroup
  \vspace{-2mm}
\end{wraptable}

\paragraph{A collective feature-space distributional training paradigm.}
We formulate \emph{distributional training} by $(\mathcal S,\mathcal E,\mathcal M,\mathcal D)$ (Figure~\ref{fig:teaser}):
the sampling scheme $\mathcal S$ supplies sample collections that provide collective supervision;
the frozen encoder $\mathcal E$ maps samples to representation spaces where they are compared. For real and generated feature distributions $p$ and $q_\theta$, 
the distribution model $\mathcal M$ constructs parametric approximations $P$ and $Q$,
and the discrepancy $\mathcal D$ measures the mismatch.
This construction provides distributional supervision: each feature update depends on the entire sampled population through the distribution model, rather than only on an individual target.

FD-Loss and Drifting provide collective supervision through shared moments or sample interactions rather than independent reconstruction targets
\citep{yang2026fdloss,deng2026drifting}.
FD-Loss is trained in the feature space of three independent encoders, while Drifting uses a latent MAE encoder.
By default, we adopt FD-Loss's sampling scheme $\mathcal{S}$ and encoders $\mathcal{E}$, and focus on the distribution model and discrepancy $(\mathcal M,\mathcal D)$.

\paragraph{A global discrepancy induces a pointwise descent field.}
To compare loss-based and field-based training prescriptions, we need to connect a scalar distributional mismatch to an update direction for each generated feature. The Wasserstein gradient flow~\citep{jordan1998variational,peyre2019computational} provides this connection by describing how feature locations should move to decrease the chosen discrepancy.
Fix $P$ and let $\mathcal{F}(Q)=\mathcal D(Q,P)$, where $Q_t$ is the evolving generated distribution under continuous flow time $t$.
Let $\delta \mathcal{F}/\delta Q$ denote the first variation of the objective, describing its sensitivity to infinitesimal density changes. The corresponding 2-Wasserstein gradient flow induces a descent
velocity field $v_{\mathcal D,t}$. Under suitable regularity, the velocity and energy dissipation satisfy:
\begin{align}
v_{\mathcal D,t}(z)
=-\nabla_z\left.\frac{\delta \mathcal{F}}{\delta Q}(z)\right|_{Q=Q_t},
\quad\frac{\text{d}z_t}{\text{d}t}=v_{\mathcal D,t}(z_t),
\quad\frac{\text{d}}{\text{d}t}\mathcal{F}(Q_t)
=-\mathbb E_{z\sim Q_t}\!\left[\|v_{\mathcal D,t}(z)\|_2^2\right]\leq0.
\label{eq:mg-wgf}
\end{align}
Thus \textit{moving points along the induced field decreases global discrepancy}. Details are given in Appendix~\ref{app:energy-descent}.

\paragraph{FD-Loss and Drifting are special cases of distributional training.} We now instantiate this construction with specific distribution models and discrepancies to recover the feature-update fields of FD-Loss and Gaussian-kernel Drifting.
For $\mathcal{M}$, let $p_{\mathrm G}=\mathcal{N}(\mu_p,\Sigma_p)$ and $q_{\mathrm G}=\mathcal{N}(\mu_q,\Sigma_q)$ be moment-matched Gaussian models. Alternatively, let $p_h$, $q_h$ be Gaussian kernel density estimates (KDEs) of $p, q_\theta$ using sample-centered kernels with covariance $h^2I$~\citep{parzen1962density}.
For $\mathcal{D}$, we consider an optimal transport (OT) approach using the 2-Wasserstein distance $W_2$ and a score-based approach with the Kullback-Leibler divergence $\KL$; these choices recover the FD-Loss and Gaussian-kernel Drifting fields~\citep{lai2026unifieddrifting,cao2026gradientflowdrifting}:
\begin{align}
\tfrac12W_2^2(q_{\mathrm G},p_{\mathrm G})
&\xrightarrow{\text{WGF}}
T_{q_{\mathrm G}\to p_{\mathrm G}}(z)-z=v_{\text{FD}}(z),\nonumber\\
\KL(q_h\|p_h)
&\xrightarrow{\text{WGF}}
\nabla_z\log p_h(z)-\nabla_z\log q_h(z)
=h^{-2}v_{\text{drift}}(z),
\label{eq:mg-instances}
\end{align}
where $T_{q_{\mathrm G}\to p_{\mathrm G}}$ is the closed-form optimal transport map between Gaussians~\citep{peyre2019computational}. The Fr\'{e}chet distance equals $W_2^2$, and differentiating it through feature moments moves samples along the first descent field. Gaussian-kernel Drifting instead evaluates the KL velocity on the smoothed densities.
The two methods occupy global-moment and sample-local ends of the modeling spectrum, with different discrepancies. Appendices~\ref{app:drifting-derivation}--\ref{app:fd-derivation} give proofs and derivations.

Other methods could also fit within this framework. For instance, calculating an OT-based discrepancy on empirical measures gives W-Flow~\citep{han2026wflow}, which constructs optimal transport between batches with Sinkhorn~\citep{cuturi2013sinkhorn,feydy2019sinkhorn} approximation. Moreover, the framework can inspire new distributional training algorithms.
Keeping the single Gaussian surrogate but replacing $W_2$ with KL divergence gives the following closed-form velocity field:
\begin{align}
    v_{\mathrm{KL}}(z)=\Sigma_p^{-1}(\mu_p-z)-\Sigma_q^{-1}(\mu_q-z)
\end{align}
which we denote as Gaussian KL. We conduct experiments with all methods mentioned above under the same settings, using the same sampling scheme as FD-Loss and a frozen Inception~\citep{szegedy2016inception} encoder. Results are in Table~\ref{tab:distribution-model-comparison}. All alternatives outperform FD-Loss on FDr$^6$, \emph{even without direct optimization of the Fr\'{e}chet distance}, demonstrating the versatility of our formulation.

\section{MGFlow}
\label{sec:method-mixtures}

As shown in Section~\ref{sec:method-unified}, FD-Loss and Gaussian-kernel Drifting take two extremes on the distribution model spectrum: A single Gaussian models only global first and second moments, whereas Gaussian KDE uses sample-centered components and overlooks global covariance structures. In order to improve modeling granularity while capturing global correlations in feature spaces, we introduce \textbf{Mixture Gradient Flow (MGFlow)}, a distributional training algorithm that adopts an intermediate distribution model $\mathcal{M}$ based on a Gaussian mixture (GM). We first introduce Gaussian-mixture representations and direct OT- and KL-based matching constructions. We then show why representing multiple modes alone is insufficient and develop a coupled allocation-and-update procedure that preserves correspondence with the reference components while still bounding the global discrepancies.

\subsection{From a Single Gaussian to Gaussian Mixtures}

For the distribution models $P,Q$, consider using Gaussian Mixtures with $K$ Gaussian components:
\begin{align}
P(z)=\sum_{k=1}^K\pi_kp_k(z),\quad
Q(z)=\sum_{k=1}^K\omega_kq_k(z),
\label{eq:mg-mixtures}
\end{align}
where $p_k=\mathcal N(\mu_{p,k},\Sigma_{p,k})$ and $q_k=\mathcal N(\mu_{q,k},\Sigma_{q,k})$. $P$ is fit offline by the EM algorithm~\citep{dempster1977em} and remains fixed, while $Q$ summarizes the changing generated feature distribution, which requires updating through training. \textit{The case $K=1$ recovers a single Gaussian; sample-centered components with a common isotropic covariance ($K=B$) recover Gaussian KDE at the representation level.} Table~\ref{tab:distribution-model-comparison} summarizes the model--discrepancy combinations.

Both discrepancies, $W_2$ and KL, extend to this representation. Though Wasserstein Distances between GMs are intractable, restricting transport to Gaussian component pairs gives a tractable upper bound, namely the Mixture Wasserstein distance~\citep{delon2020gmmwasserstein}:
\begin{align}
W_2^2(Q,P)\leq MW_2^2(Q,P)
&:=\min_{\Gamma\in U(\omega,\pi)}
\sum_{i,j}\Gamma_{ij}W_2^2(q_i,p_j).
\label{eq:mg-mixture-ot}
\end{align}
Here $U(\omega,\pi)$ is the set of component couplings with marginals $\omega$ and $\pi$, and each pair cost is the Gaussian FD. For KL, the global velocity field is the difference of the score functions:
\begin{align}
v_{\text{global}}(z)
&=\nabla_z\log P(z)-\nabla_z\log Q(z)
=\sum_k\gamma_{P,k}(z)s_{p,k}(z)
-\sum_k\gamma_{Q,k}(z)s_{q,k}(z),
\label{eq:mg-global-kl}
\end{align}
where $s_{r,k}(z)=\nabla_z\log r_k(z) = \Sigma_{r,k}^{-1}(\mu_{r,k}-z)$ for $r\in\{p,q\}$, $\gamma_{P,k}=\pi_kp_k/P$, and $\gamma_{Q,k}=\omega_kq_k/Q$. The score of each mixture is a posterior-responsibility-weighted sum of its component scores.

However, representing multiple modes does not ensure correct mode proportions.
The global KL field can provide weak inter-mode updates when modes are well separated.
Consider $P=\sum_k\pi_kr_k$ and $Q=\sum_k\omega_kr_k$ with the same separated components $r_k$ but different positive weights. Correcting this weight mismatch requires movement between modes, not merely local refinement within each mode. However, in a region dominated by component $k$, the density ratio is approximately the constant $\omega_k/\pi_k$; the score difference can be small even when the component weights differ substantially:
\begin{align}
\log\frac{Q(z)}{P(z)}\approx\log\frac{\omega_k}{\pi_k},\quad
    v_{\text{global}}(z)=-\nabla_z\log\frac{Q(z)}{P(z)}\approx0
\end{align}

Thus, a mismatch can remain visible in the density ratio while producing only a weak feature-update signal in the high-density regions. This is analogous to the score-blindness phenomenon studied by~\citet{wenliang2020blindness}, and does not contradict the descent interpretation in Section~\ref{sec:method-unified}: descent does not guarantee effective transport between modes when the score field is weak. In the top-right state of Figure~\ref{fig:method-toy-pairing}, the two Q-components \(q_1\) and \(q_2\) carry 85.1\% and 14.9\% of the mass, respectively; the global velocity field alone provides too little inter-mode transport to correct this imbalance. The same issue can arise for Gaussian-KDE fields when smoothing leaves modes well separated. For score-based matching, \emph{mixture expressivity does not by itself provide an effective mechanism for correcting mode proportions.} We therefore connect mass allocation explicitly to the feature updates.

\Needspace{9\baselineskip}
\subsection{Mass-Constrained Allocation and Paired Transport}
\label{sec:method-paired}

\begin{wraptable}{r}{0.50\textwidth}
  \centering
  \vspace{-10pt}
  \setlength{\belowcaptionskip}{2pt}
  \caption{\small\textbf{Ablation on component assignment and transportation.}
  JiT-B trains in Inception feature space for 10 epochs with $K=4$. Timing uses 8 H200 GPUs.}
  \label{tab:method-lp}
  \label{tab:method-score}
  \fontsize{8}{9.5}\selectfont
  \setlength{\tabcolsep}{2pt}
  \begin{adjustbox}{Clip=0pt 0pt 0pt 0pt}
  \begin{tabularx}{\linewidth}{@{}l>{\centering\arraybackslash}Xc>{\centering\arraybackslash}Xc@{}}
    \toprule
    & \multicolumn{2}{c}{KL} & \multicolumn{2}{c}{$W_2$} \\
    \cmidrule(lr){2-3}\cmidrule(l){4-5}
    & $\text{FDr}^6\!\downarrow$ & s/step
    & $\text{FDr}^6\!\downarrow$ & s/step \\
    \midrule
    Posterior & 173.56 & \textbf{0.197} & 26.78 & 0.699 \\
    LP-global & 146.92 & 0.264 & 23.73 & 0.825 \\
    \rowcolor{tableOursTint}
    LP-paired & \textbf{23.27} & 0.263 & \textbf{23.42} & \textbf{0.392} \\
    \bottomrule
  \end{tabularx}
  \end{adjustbox}

  \vspace{2pt}
\end{wraptable}

\begin{figure}[t]
  \centering
  \includegraphics[width=\linewidth]{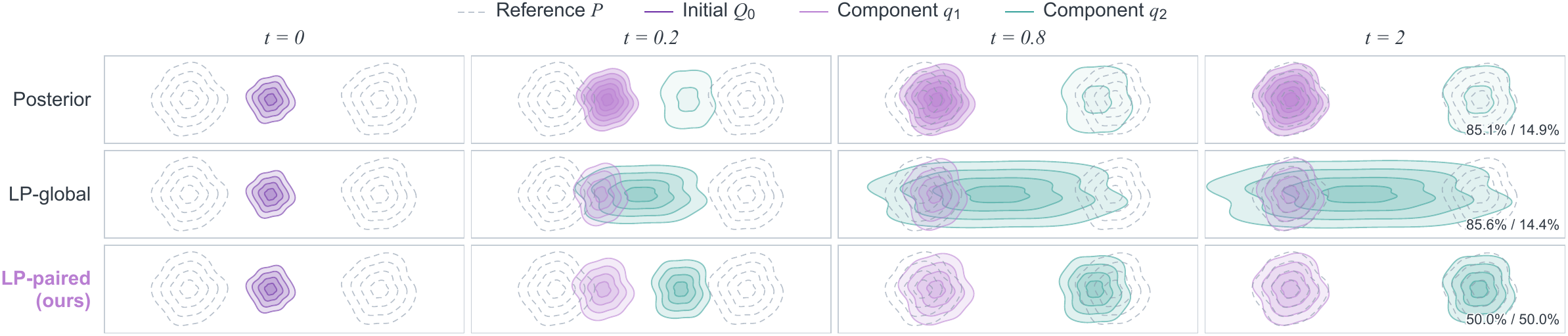}
  \vspace{-3mm}
  \caption{\small\textbf{Toy experiment with component assignment and transport.}
  Score matching with $K=2$ moves 2048 particles. Dashed contours show the reference $P$. Shading and contour counts indicate weights. LP fixes component weights, but global velocity collapses modes. Only LP-paired achieves correct transport. Detailed settings are given in Appendix~\ref{app:toy-settings}.}
  \label{fig:method-toy-pairing}
  \vspace{-3mm}
\end{figure}

\paragraph{LP-based component allocation.}
The reference mixture $P$ is fitted offline and kept fixed, whereas $Q$ must track the changing generator throughout training. For a single Gaussian, each generated batch contributes to one set of global moments. A GM instead requires component-specific moments, so updating $Q$ requires allocating new features to components.
A natural choice is posterior soft assignment: each generated sample
$z_n$ contributes to component $k$ with weight $\gamma_{Q,k}(z_n) = \omega_k q_k(z_n)/Q(z_n)$, its posterior responsibility under the current $Q$. However, due to dynamical reasons, posterior soft assignment cannot transport the correct amount of mass to each component, as shown in Figure~\ref{fig:method-toy-pairing}.

We instead assign generated features to the fixed reference components $p_k$ while constraining the mass allocated to component $k$
to $B\pi_k$ for a batch of $B$ features. To achieve this, we solve the
capacity-constrained linear program (LP) for batch-to-component allocation:
\begin{align}
  R^* \in \operatorname*{arg\,min}_{R\geq0}
  \sum_{n,k}R_{nk}[-\log(\pi_k p_k(z_n))],
  \quad\text{subject to}\quad
  R\mathbf{1}_K=\mathbf{1}_B,\quad
  R^{\mathsf T}\mathbf{1}_B=B\pi.
  \label{eq:method-batch-lp}
\end{align}
Here $R\in\mathbb R^{B\times K}$, where $R_{nk}$ denotes the assignment from sample $n$ to component $k$, which may be fractional.
The likelihood cost favors compatible reference components, while the capacity constraints enforce their prescribed masses.
Using detached $R^*$, we compute the weighted mean and raw second moment for each generated component, maintain these statistics across training batches with EMA as in FD-Loss~\citep{yang2026fdloss}, and recover the component covariances. We set $\omega=\pi$.

\paragraph{Paired OT matching.}
With matched weights, $\Gamma=\operatorname{diag}(\pi)$ is feasible in
Eq.~\eqref{eq:mg-mixture-ot}. Empirically, in a pMF-H training
run with this objective, \emph{the optimal component transport matrix was diagonal at every recorded plan refresh} (Appendix~\ref{app:diagonal-training}). Motivated by this observation, we fix this correspondence and calculate only the diagonal terms in the Mixture Wasserstein distance, optimizing:
\begin{equation}
  \mathcal{L}_{\text{pair-}W_2}
  =\sum_{k=1}^{K}\pi_k W_2^2(q_k,p_k)\ge MW_2^2(Q,P)\ge W_2^2(Q,P).
  \label{eq:method-mb-pair}
\end{equation}
This provides an upper bound for $W_2^2$. A sufficient condition for diagonal optimality is given in Appendix~\ref{app:method-diagonal}.
Fixed pairing reduces the number of Gaussian pair costs from $K^2$ to $K$.
We differentiate this objective through the current batch's contribution to the EMA statistics of the generated components, keeping historical statistics and the LP assignments $R^*$ detached.

\paragraph{Paired score matching.}
For the KL discrepancy, pairing is more than a matter of efficiency, as constraining the component statistics is not sufficient to constrain the feature updates. 
The global field in Eq.~\eqref{eq:mg-global-kl} still weights reference and generated scores by separate mixture responsibilities, rather than the LP assignments. 
As demonstrated in the second row of Figure~\ref{fig:method-toy-pairing}, the global velocity cannot distinguish the target component for each sample, resulting in mode collapse. To preserve the assignment in the update direction, we instead match corresponding components using
\begin{align}
    \mathcal{F}_{\text{pair}}=\sum_k\pi_k\KL(q_k\Vert p_k)\ge\KL(Q\Vert P)
\end{align}
which also upper-bounds the marginal KL. Appendix~\ref{app:method-paired-kl} gives the proof and further analysis. To avoid the heavy computational overhead of differentiating the closed-form
KL between Gaussians through the estimated covariances, we use the WGF-induced velocity.
Reusing $R^*$ for these fields gives the regularized update
\begin{equation}
  v_{\text{pair},n}=\sum_{k=1}^{K}R^*_{nk}
       [s_{p,k}^{\lambda}(z_n)-s_{q,k}^{\lambda}(z_n)],
  \quad
  s_{r,k}^{\lambda}(z)=(\Sigma_{r,k}+\lambda I)^{-1}(\mu_{r,k}-z).
  \label{eq:method-paired-score}
\end{equation}
Here $\lambda>0$ stabilizes covariance inversion.
The same assignment now weights both scores, so the update retains
the correspondence used to estimate $q_k$.
We evaluate the field using detached, pre-update statistics and train
the generator by detached-target regression:
\begin{align}
  \mathcal L_\text{pair-KL}
  &=\frac{1}{2B}\sum_{n=1}^{B}
  \left\|z_n-\operatorname{sg}[z_n+\eta v_{\text{pair},n}]\right\|_2^2,
  \qquad \eta>0.
  \label{eq:mg-regression}
\end{align}
The entire target is detached, and the EMA state is committed after
the generator step. The third row in Figure~\ref{fig:method-toy-pairing} shows that this formulation correctly transports samples from the initial distribution to the reference distribution. An ablation study in Table~\ref{tab:method-lp} proves that LP-based component allocation uniformly achieves better results than posterior assignments. Paired transport is crucial especially when training multi-step JiT models with score-based discrepancies. This is possibly because the large gap between the initial one-step generated distribution and the target distribution induces abnormal component allocation. Detailed training and evaluation settings are given in Appendix~\ref{app:assignment-ablation-settings}.

\subsection{Training MGFlow.}
As discussed above, we train two variants of MGFlow:
\emph{MGFlow-$W_2$} with the paired OT-based objective in
Equation~\ref{eq:method-mb-pair}, and \emph{MGFlow-KL} with paired
score-based updates in Equation~\ref{eq:mg-regression}.
For multiple encoders, we use fixed, discrepancy-specific normalizers for the encoder losses:
\begin{align}
    \mathcal L_{\text{MGFlow-}W_2}
    = \sum_e
    \frac{\mathcal L_{\text{pair-}W_2}^e}{W_2^2(R_e,V_e)},
    \quad
    \mathcal L_{\text{MGFlow-KL}}
    = \sum_e
    \frac{\mathcal{L}_{\text{pair-KL}}^{e}}
         {\KL(R_e\|V_e)}.
    \label{eq:mgflow-multi-encoder}
\end{align}
Here $\mathcal L_{\text{pair-}W_2}^e$ and $\mathcal L_{\text{pair-KL}}^e$ sum the respective
branch losses for encoder $e$, while $R_e$ and $V_e$ are single-Gaussian
fits to its real training and validation features.
For KL calibration, the same score ridge is added to both covariances.
Unlike FD-Loss's real--generated $(R,G)$ normalization, these fixed
$(R,V)$ scales avoid downweighting harder-to-match feature spaces
merely because their current discrepancies are large (Appendix~\ref{app:loss-weighting}).

\Needspace{12\baselineskip}
\section{Experiments}
\label{sec:experiments}
\subsection{Increasing Gaussian Components}
\begin{wraptable}{r}{0.50\textwidth}
  \centering
  \setlength{\belowcaptionskip}{6pt}
  \vspace{-20pt}
  \caption{\small\textbf{Component-count ablation for MGFlow-KL.} All configurations use pMF-B trained in Inception feature space for 10 epochs. $\text{FDr}^{5}$ evaluates five held-out encoders.}
  \label{tab:gmm-component-ablation}
  \begingroup
  \fontsize{8.5}{10}\selectfont
  \setlength{\tabcolsep}{3pt}
  \renewcommand{\arraystretch}{1.15}
  \begin{tabular*}{\linewidth}{@{\extracolsep{\fill}}lccc@{}}
    \toprule
    Components & \MetricHighlight{metricFDrSixTint}{$\text{FDr}^6\!\downarrow$}
      & \MetricHighlight{metricFDrHeldoutTint}{$\text{FDr}^5\!\downarrow$}
      & \MetricHighlight{metricFIDTint}{FID $\!\downarrow$} \\
    \midrule
    $K=1$ & 12.73 & 15.04 & 1.95 \\
    $K=4$ & 12.28 & 14.54 & 1.65 \\
    $K=16$ & 11.32 & 13.41 & \textbf{1.46} \\
    \midrule
    $K=1+4$ & 11.51 & 13.59 & 1.89 \\
    $K=1+4+16$ & \textbf{11.05} & \textbf{13.08} & 1.51 \\
    \bottomrule
  \end{tabular*}
  \endgroup
\vspace{-12pt}
\end{wraptable}

We first examine how distributional granularity affects ImageNet
post-training. We compare individual Gaussian mixture models with
$K\in\{1,4,16\}$ components and their combinations.
Beyond selecting a single GM, we consider \emph{hierarchical
resolution}: for a set of component counts $\mathcal{K}$, we sum the
corresponding training losses with equal weights,
\begin{align}
    \mathcal{L}_{\mathcal{K}}
    = \sum_{K\in\mathcal{K}} \mathcal{L}_K,
    \label{eq:hierarchical-matching}
\end{align}
where $\mathcal{L}_K$ denotes the MGFlow loss using $K$-component
distribution models.
Thus, $K=1+4+16$ combines global moment matching with progressively
finer componentwise matching, forming a coarse-to-fine hierarchy
of objectives.

Table~\ref{tab:gmm-component-ablation} compares component counts and their
combinations for MGFlow-KL on pMF-B.
Among the same $256\times256$ resolution configurations, increasing $K$ from $1$ to $16$ reduces
FDr$^6$ from $12.73$ to $11.32$ and FID from $1.95$ to $1.46$, demonstrating that distributional models with finer granularity enhance generation.
Hierarchical resolution further improves generalization, lowering FDr$^6$ and FDr$^5$. These results suggest that coarse and fine distributional
constraints provide complementary supervision, motivating
multi-resolution matching in the subsequent experiments. By default we use
$K=1+4+16$ for MGFlow-KL and $K=1+4$ for MGFlow-$W_2$.
Larger mixtures are limited by the cost of Gaussian Wasserstein matching
for MGFlow-$W_2$ and the samples available to accurately estimate each component's mean and covariance
for MGFlow-KL (see Appendix~\ref{app:component-scaling}).

\begin{table*}[t]
  \centering
  \caption{\small\textbf{Post-training JiT and pMF on ImageNet $\bm{256\times256}$.} All post-training runs use SIM encoder spaces for 100 epochs.
  ``Model'' denotes feature-distribution modeling. Baseline results are from
  FD-Loss~\citep{yang2026fdloss}, AdvFD~\citep{gao2026advfd},
  and AMFD~\citep{liu2026amfd}. 
  $^\dagger$ denotes the full-CFG NFE upper bound for interval CFG;
  dashes denote unavailable results.
  Best and second-best results per backbone are bold and underlined, respectively.
  More baselines are in Table~\ref{tab:imagenet256-additional-baselines}.}
  \label{tab:imagenet256-system-comparison}
  \begingroup
  \fontsize{8.2}{9.2}\selectfont
  \setlength{\tabcolsep}{2.8pt}
  \renewcommand{\arraystretch}{1.15}
  \begin{adjustbox}{Clip=0pt 0pt 0pt 0pt}
  \begin{tabularx}{\linewidth}{@{}p{0.20\linewidth}cccc*{4}{>{\centering\arraybackslash}X}@{}}
    \toprule
    Method & Model & Discrepancy & NFE & \#Params
      & \MetricHighlight{metricFDrSixTint}{$\text{FDr}^6\!\downarrow$}
      & \MetricHighlight{metricFDrHeldoutTint}{$\text{FDr}^3\!\downarrow$}
      & \MetricHighlight{metricFIDTint}{$\text{FID}\!\downarrow$} & $\text{IS}\!\uparrow$ \\
    \midrule
    \rowcolor{tableSectionTint}
    \multicolumn{9}{@{}l}{\textit{Pixel-space models, multi-step backbones}} \\
    JiT-B & -- & --
      & \textcolor{black!35}{$50{\times}2{\times}2^\dagger$} & \textcolor{black!35}{131M} & 15.65 & 15.06 & 3.71 & 269.0 \\
    \rowcolor{tablePriorTint}
    \quad $+$ FD-Loss & Gaussian & $W_2$
      & \textcolor{black!35}{1} & \textcolor{black!35}{131M} & 5.53 & 8.45 & 1.00 & \textbf{344.6} \\
    \rowcolor{tablePriorTint}
    \quad\quad $+$ AdvFD & Gaussian & $W_2$
      & \textcolor{black!35}{1} & \textcolor{black!35}{131M} & 3.92 & \underline{6.03} & \textbf{0.79} & -- \\
    \rowcolor{tablePriorTint}
    \quad\quad $+$ AMFD-C & Gaussian & Amortized $W_2$
      & \textcolor{black!35}{1} & \textcolor{black!35}{131M} & 4.75 & 6.71 & \underline{0.95} & 319.4 \\ %
    \rowcolor{tablePriorTint}
    \quad\quad $+$ AMFD-U & Gaussian & Amortized $W_2$
      & \textcolor{black!35}{1} & \textcolor{black!35}{131M} & \underline{3.91} & 6.13 & \underline{0.95} & \underline{325.2} \\ %
    \rowcolor{tableOursTint}
    \quad $+$ MGFlow-$W_2$ & GM & $W_2$
      & \textcolor{black!35}{1} & \textcolor{black!35}{131M} & 4.43 & 7.36 & 1.45 & 313.9 \\ %
    \rowcolor{tableOursTint}
    \quad $+$ MGFlow-KL & GM & KL
      & \textcolor{black!35}{1} & \textcolor{black!35}{131M} & \textbf{3.70} & \textbf{5.70} & 1.28 & 308.5 \\
    \addlinespace[0.15em]
    JiT-L & -- & --
      & \textcolor{black!35}{$50{\times}2{\times}2^\dagger$} & \textcolor{black!35}{459M} & 10.73 & 10.27 & 2.59 & 288.5 \\
    \rowcolor{tablePriorTint}
    \quad $+$ FD-Loss & Gaussian & $W_2$
      & \textcolor{black!35}{1} & \textcolor{black!35}{459M} & 3.24 & 5.46 & \underline{0.77} & 317.3 \\
    \rowcolor{tablePriorTint}
    \quad\quad $+$ AdvFD & Gaussian & $W_2$
      & \textcolor{black!35}{1} & \textcolor{black!35}{459M} & \underline{2.01} & 3.20 & \textbf{0.73} & -- \\
    \rowcolor{tablePriorTint}
    \quad\quad $+$ AMFD-C & Gaussian & Amortized $W_2$
      & \textcolor{black!35}{1} & \textcolor{black!35}{459M} & 2.67 & 3.86 & 0.87 & \textbf{325.1} \\ %
    \rowcolor{tablePriorTint}
    \quad\quad $+$ AMFD-U & Gaussian & Amortized $W_2$
      & \textcolor{black!35}{1} & \textcolor{black!35}{459M} & 2.02 & \underline{3.12} & 0.85 & \underline{319.8} \\ %
    \rowcolor{tableOursTint}
    \quad $+$ MGFlow-$W_2$ & GM & $W_2$
      & \textcolor{black!35}{1} & \textcolor{black!35}{459M} & 2.68 & 4.54 & 1.07 & 314.9 \\ %
    \rowcolor{tableOursTint}
    \quad $+$ MGFlow-KL & GM & KL
      & \textcolor{black!35}{1} & \textcolor{black!35}{459M} & \textbf{1.92} & \textbf{2.96} & 1.00 & 304.7 \\
    \addlinespace[0.15em]
    JiT-H & -- & --
      & \textcolor{black!35}{$50{\times}2{\times}2^\dagger$} & \textcolor{black!35}{953M} & 7.66 & 9.07 & 1.97 & 296.0 \\
    \rowcolor{tablePriorTint}
    \quad $+$ FD-Loss & Gaussian & $W_2$
      & \textcolor{black!35}{1} & \textcolor{black!35}{953M} & 2.65 & 4.44 & \underline{0.75} & \underline{313.0} \\
    \rowcolor{tablePriorTint}
    \quad\quad $+$ AdvFD & Gaussian & $W_2$
      & \textcolor{black!35}{1} & \textcolor{black!35}{953M} & 1.80 & 2.93 & \textbf{0.72} & -- \\
    \rowcolor{tablePriorTint}
    \quad\quad $+$ AMFD-C & Gaussian & Amortized $W_2$
      & \textcolor{black!35}{1} & \textcolor{black!35}{953M} & 2.15 & 3.10 & 0.85 & \textbf{328.1} \\ %
    \rowcolor{tablePriorTint}
    \quad\quad $+$ AMFD-U & Gaussian & Amortized $W_2$
      & \textcolor{black!35}{1} & \textcolor{black!35}{953M} & \underline{1.79} & \underline{2.78} & 0.83 & 312.2 \\ %
    \rowcolor{tableOursTint}
    \quad $+$ MGFlow-$W_2$ & GM & $W_2$
      & \textcolor{black!35}{1} & \textcolor{black!35}{953M} & 2.55 & 4.42 & 0.99 & 307.8 \\ %
    \rowcolor{tableOursTint}
    \quad $+$ MGFlow-KL & GM & KL
      & \textcolor{black!35}{1} & \textcolor{black!35}{953M} & \textbf{1.64} & \textbf{2.57} & 0.94 & 301.9 \\

    \midrule
    \rowcolor{tableSectionTint}
    \multicolumn{9}{@{}l}{\textit{Pixel-space models, one-step}} \\
    pMF-B & -- & --
      & \textcolor{black!35}{1} & \textcolor{black!35}{118M} & 13.70 & 11.82 & 3.31 & 254.6 \\
    \rowcolor{tablePriorTint}
    \quad $+$ FD-Loss & Gaussian & $W_2$
      & \textcolor{black!35}{1} & \textcolor{black!35}{118M} & 3.50 & 4.49 & \underline{0.85} & \underline{331.4} \\
    \rowcolor{tablePriorTint}
    \quad\quad $+$ AdvFD & Gaussian & $W_2$
      & \textcolor{black!35}{1} & \textcolor{black!35}{118M} & 3.32 & 4.22 & \textbf{0.81} & -- \\
    \rowcolor{tablePriorTint}
    \quad\quad $+$ AMFD-C & Gaussian & Amortized $W_2$
      & \textcolor{black!35}{1} & \textcolor{black!35}{118M} & 3.94 & 4.66 & 0.95 & 315.9 \\ %
    \rowcolor{tablePriorTint}
    \quad\quad $+$ AMFD-U & Gaussian & Amortized $W_2$
      & \textcolor{black!35}{1} & \textcolor{black!35}{118M} & 3.43 & 4.20 & 0.92 & 310.1 \\ %
    \rowcolor{tableOursTint}
    \quad $+$ MGFlow-$W_2$ & GM & $W_2$
      & \textcolor{black!35}{1} & \textcolor{black!35}{118M} & \textbf{3.03} & \underline{4.02} & 1.65 & \textbf{332.1} \\ %
    \rowcolor{tableOursTint}
    \quad $+$ MGFlow-KL & GM & KL
      & \textcolor{black!35}{1} & \textcolor{black!35}{118M} & \underline{3.09} & \textbf{3.76} & 1.56 & 323.0 \\
    \addlinespace[0.15em]
    pMF-L & -- & --
      & \textcolor{black!35}{1} & \textcolor{black!35}{410M} & 9.09 & 7.62 & 2.72 & 261.7 \\
    \rowcolor{tablePriorTint}
    \quad $+$ FD-Loss & Gaussian & $W_2$
      & \textcolor{black!35}{1} & \textcolor{black!35}{410M} & 2.09 & 2.72 & \underline{0.78} & 309.2 \\
    \rowcolor{tablePriorTint}
    \quad\quad $+$ AdvFD & Gaussian & $W_2$
      & \textcolor{black!35}{1} & \textcolor{black!35}{410M} & 1.89 & 2.57 & \textbf{0.77} & -- \\
    \rowcolor{tablePriorTint}
    \quad\quad $+$ AMFD-C & Gaussian & Amortized $W_2$
      & \textcolor{black!35}{1} & \textcolor{black!35}{410M} & 2.25 & 2.75 & 0.88 & \underline{321.8} \\ %
    \rowcolor{tablePriorTint}
    \quad\quad $+$ AMFD-U & Gaussian & Amortized $W_2$
      & \textcolor{black!35}{1} & \textcolor{black!35}{410M} & 2.01 & 2.57 & 0.86 & 306.7 \\ %
    \rowcolor{tableOursTint}
    \quad $+$ MGFlow-$W_2$ & GM & $W_2$
      & \textcolor{black!35}{1} & \textcolor{black!35}{410M} & \underline{1.80} & \underline{2.47} & 1.20 & \textbf{322.4} \\ %
    \rowcolor{tableOursTint}
    \quad $+$ MGFlow-KL & GM & KL
      & \textcolor{black!35}{1} & \textcolor{black!35}{410M} & \textbf{1.74} & \textbf{2.24} & 1.16 & 314.8 \\
    \addlinespace[0.15em]
    pMF-H & -- & --
      & \textcolor{black!35}{1} & \textcolor{black!35}{935M} & 6.87 & 6.09 & 2.29 & 267.2 \\
    \rowcolor{tablePriorTint}
    \quad $+$ FD-Loss & Gaussian & $W_2$
      & \textcolor{black!35}{1} & \textcolor{black!35}{935M} & 1.89 & 2.69 & \underline{0.77} & 310.1 \\
    \rowcolor{tablePriorTint}
    \quad\quad $+$ AdvFD & Gaussian & $W_2$
      & \textcolor{black!35}{1} & \textcolor{black!35}{935M} & 1.74 & 2.50 & \textbf{0.74} & -- \\
    \rowcolor{tablePriorTint}
    \quad\quad $+$ AMFD-C & Gaussian & Amortized $W_2$
      & \textcolor{black!35}{1} & \textcolor{black!35}{935M} & 1.93 & 2.43 & 0.86 & \textbf{323.0} \\ %
    \rowcolor{tablePriorTint}
    \quad\quad $+$ AMFD-U & Gaussian & Amortized $W_2$
      & \textcolor{black!35}{1} & \textcolor{black!35}{935M} & 1.75 & 2.30 & 0.85 & 307.3 \\ %
    \rowcolor{tableOursTint}
    \quad $+$ MGFlow-$W_2$ & GM & $W_2$
      & \textcolor{black!35}{1} & \textcolor{black!35}{935M} & \underline{1.50} & \underline{2.05} & 1.09 & \underline{316.1} \\ %
    \rowcolor{tableOursTint}
    \quad $+$ MGFlow-KL & GM & KL
      & \textcolor{black!35}{1} & \textcolor{black!35}{935M} & \textbf{1.45} & \textbf{1.88} & 1.07 & 311.0 \\
    \bottomrule
  \end{tabularx}
  \end{adjustbox}
  \endgroup
  \vspace{-1mm}
\end{table*}

\subsection{Class-Conditioned ImageNet Generation}

We post-train the B, L, and H variants of JiT~\citep{li2026jit} and
pMF~\citep{lu2026pmf} from their official pretrained weights on
ImageNet~\citep{russakovsky2015imagenet} $256\times256$.
All use frozen SigLIP~\citep{tschannen2025siglip2},
Inception~\citep{szegedy2016inception}, and MAE~\citep{he2022mae}
encoders (SIM) and a global batch
of 1,024, and are trained for 100 epochs.
We evaluate post-trained models with 50,000 one-step generated samples.
FD-Loss~\citep{yang2026fdloss} motivates
\MetricHighlight{metricFDrSixTint}{FDr$^6$} as a more robust metric for
ImageNet generation than \MetricHighlight{metricFIDTint}{FID} alone,
reducing the blind spots of a single representation.
We report \MetricHighlight{metricFDrSixTint}{FDr$^6$} across
six representation spaces and \MetricHighlight{metricFDrHeldoutTint}{FDr$^3$}
across the three encoders not used for training, together with
\MetricHighlight{metricFIDTint}{FID}~\citep{heusel2017fid} and
Inception Score (IS)~\citep{salimans2016improved}.
Optimization, sampling, and reference-fitting details are given in
Appendices~\ref{app:setup} and~\ref{app:references}; metric definitions
are in Appendix~\ref{app:metrics}.

Table~\ref{tab:imagenet256-system-comparison} compares MGFlow with
FD-Loss~\citep{yang2026fdloss}, AdvFD~\citep{gao2026advfd}, and
AMFD~\citep{liu2026amfd}. Both MGFlow-$W_2$ and MGFlow-KL surpass the FD-Loss baseline on every model size.
MGFlow-KL achieves \emph{state-of-the-art} $\mathbf{1.45}$ FDr$^6$ on pMF-H and $\mathbf{1.64}$ on JiT-H, improving over the FD-Loss baseline with \textbf{23\%} and \textbf{38\%} margins, respectively. In particular, its gains on the three held-out encoders indicate better generalization beyond the representations used for training, demonstrating that MGFlow can truly match distributions in feature spaces rather than just overfit to the first and second moments.
Additionally, MGFlow-KL uses KL-based matching without any FD loss, yet
outperforms all the FD-based baselines on both FDr$^6$ and held-out FDr$^3$,
showing that its gains do not rely on directly optimizing these evaluation
metrics.
See Appendix~\ref{app:encoder-results} for per-encoder results.
Qualitative results are demonstrated in Appendix~\ref{app:imagenet-qualitative}.

\FloatBarrier
\begin{table*}[!t]
  \centering
  \caption{\small\textbf{Text-to-image generation with FLUX.2 [klein] 4B}
  \citep{blackforestlabs2026flux2klein} at $\bm{512\times512}$.
  We report GenEval~\citep{ghosh2023geneval} and
  PickScore~\citep{kirstain2023pick}; PickScore is evaluated on 499 Pick-a-Pic prompts. Higher is better.
  Bold and underlining mark the best and second-best scores, respectively.
  Baselines are from AMFD~\citep{liu2026amfd}; dashes denote unavailable results.}
  \label{tab:text-to-image-comparison}
  \begingroup
  \fontsize{8.5}{10}\selectfont
  \setlength{\tabcolsep}{2.3pt}
  \renewcommand{\arraystretch}{1.1}
  \begin{adjustbox}{Clip=0pt 0pt 0pt 0pt}
  \begin{tabular}{@{}l>{\hspace{-2.6pt}}c<{\hspace{2.6pt}}*{8}{c}@{}}
    \toprule
    Method & NFE & Single & Two & Count & Colors
      & Position & Color attr. & Overall & PickScore \\
    \midrule
    FLUX.2 [klein] 4B
      & \textcolor{black!35}{4} & 0.994 & 0.904 & 0.791 & 0.880 & 0.575 & 0.623 & 0.794 & 21.85 \\
    DMD2~\citep{yin2024dmd2}
      & \textcolor{black!35}{1} & \underline{0.997} & 0.894 & 0.806 & 0.864 & 0.603 & 0.660 & 0.804 & -- \\
    FD-SIM~\citep{yang2026fdloss}
      & \textcolor{black!35}{1} & \textbf{1.000} & 0.944 & 0.716 & 0.878 & 0.618 & 0.658 & 0.802 & 21.62 \\
    iRDM~\citep{feng2026rdm}
      & \textcolor{black!35}{1} & 0.994 & 0.924 & 0.756 & \underline{0.923} & 0.650 & 0.708 & 0.826 & 21.82 \\
    AMFD-U-SIM~\citep{liu2026amfd}
      & \textcolor{black!35}{1} & 0.994 & 0.929 & 0.741 & 0.902 & 0.638 & 0.678 & 0.813 & 21.77 \\
    AMFD-C-SIM~\citep{liu2026amfd}
      & \textcolor{black!35}{1} & \underline{0.997} & 0.955 & 0.791 & \underline{0.923} & 0.670 & 0.740 & 0.846 & 21.85 \\
    AMFD-C-10 enc.~\citep{liu2026amfd}
      & \textcolor{black!35}{1} & \textbf{1.000} & 0.957 & 0.778 & 0.920 & \underline{0.680} & 0.733 & 0.845 & 21.82 \\
      \midrule
    \rowcolor{tableOursTint}
    MGFlow (image-only)
      & \textcolor{black!35}{1} & \underline{0.997} & \underline{0.967} & \underline{0.844} & \textbf{0.926} & 0.633 & \underline{0.770} & \underline{0.856} & \underline{21.86} \\
    \rowcolor{tableOursTint}
    MGFlow (joint)
      & \textcolor{black!35}{1} & \textbf{1.000} & \textbf{0.982} & \textbf{0.891} & \textbf{0.926} & \textbf{0.783} & \textbf{0.818} & \textbf{0.900} & \textbf{21.98} \\
    \bottomrule
  \end{tabular}
  \end{adjustbox}
  \endgroup
\end{table*}

\subsection{Text-to-Image Generation}

We initialize from the distilled FLUX.2 [klein] 4B model
\citep{blackforestlabs2026flux2klein} and train MGFlow-KL with $K=1+4$ for 1,000 steps with a batch size of 1,024. Following the exact protocol adopted in prior works~\citep{liu2026amfd,feng2026rdm}, we use reconstructed
COCO~\citep{lin2014microsoft} reference images augmented
with GenEval images. We use two versions of MGFlow. For the joint version, we concatenate image features with frozen SigLIP2~\citep{tschannen2025siglip2} text features~\citep{feng2026rdm}, matching the joint distribution $p(x,c)$ (for further analysis, see Appendix~\ref{app:joint-score-math}) instead of the marginal distribution $p(x)$ for better text-image alignment. By contrast,
the image-only variant uses image features alone, matching the marginal distribution.
The model is trained to generate $512\times512$ images in one step.
We report GenEval \citep{ghosh2023geneval} and PickScore on Pick-a-Pic
\citep{kirstain2023pick}. Reference construction and training details are
in Appendix~\ref{app:t2i-comparison-settings}.

With the COCO reference, joint matching reaches a GenEval score of
$\mathbf{0.900}$ in Table~\ref{tab:text-to-image-comparison}, compared with $0.794$ for the four-step
backbone, $0.826$ for iRDM, and $0.846$ for AMFD-C-SIM.
It also achieves the highest PickScore of $\mathbf{21.98}$ in the table.
iRDM~\citep{feng2026rdm} uses a batch size of 10,240 for 180 steps,
whereas AMFD~\citep{liu2026amfd} uses 1,024 for 1,500 steps.
MGFlow achieves higher GenEval and PickScore scores with \textbf{44\%} and \textbf{33\%} fewer
generated training samples, respectively.
Qualitative text-to-image examples are provided in Appendix~\ref{app:additional-t2i-example}.

\FloatBarrier
\section{Related Work}
One-step generation can be learned through trajectory consistency, adversarial training, 
average velocities, or diffusion distillation~\citep{song2023consistency,zhang2025voicebridge,geng2025meanflow,yin2024dmd2}.
Recent methods instead supervise generated populations through
global feature-space discrepancies~\citep{yang2026fdloss,feng2026rdm} or  construct updates from attraction--repulsion or
transport between batches~\citep{deng2026drifting,han2026wflow}.
We propose the unified theoretical framework of distributional training that recovers all these works.
MGFlow uses a novel Gaussian Mixture Model for distribution approximation that differs from all the above methods.
For a detailed discussion of related work, see Appendix~\ref{app:related-work}.

\Needspace{12\baselineskip}
\section{Conclusion}
We presented a unified view of distributional training that separates
sampling, encoding, distribution modeling, and matching discrepancy.
Wasserstein gradient flow connects global objectives to feature updates,
recovering FD-Loss and Gaussian-kernel Drifting as specific choices.
MGFlow combines Gaussian mixtures with
mass-constrained allocation and paired OT or KL updates.
Experiments on ImageNet and text-to-image generation show improvements
across training and held-out representations, with KL-based matching
improving Fr\'{e}chet metrics without directly optimizing them.
Both modeling granularity and component
correspondence matter for one-step generation. More broadly, our findings point toward richer distribution modeling and principled distribution matching as promising directions for advancing one-step generative models.

\label{main-text-end}

\bibliography{references}
\bibliographystyle{plainnat}

\appendix

\clearpage
\addtocontents{toc}{\protect\AppendixContentsStart}
\begingroup
\makeatletter
\hypersetup{linktoc=all}
\pdfbookmark[1]{Appendix Contents}{appendix-contents}
\section*{Appendix Contents}
\vspace{2pt}
\noindent\hfill{\small\textcolor{black!55}{Page}}\par
\vspace{2pt}
\hrule height 0.4pt
\vspace{5pt}

\c@tocdepth=2\relax
\setlength{\parskip}{0pt}
\fontsize{9.5}{15}\selectfont
\renewcommand{\@pnumwidth}{1.8em}
\renewcommand{\@tocrmarg}{2.8em}
\renewcommand{\@dotsep}{4.5}
\renewcommand*{\l@section}[2]{%
  \addvspace{11pt}%
  {\bfseries\@dottedtocline{1}{0em}{1.8em}{#1}{#2}}}
\renewcommand*{\l@subsection}[2]{%
  \@dottedtocline{2}{1.8em}{2.8em}{#1}{#2}}

\newif\ifAppendixContentsActive
\AppendixContentsActivefalse
\def\AppendixContentsStart{\AppendixContentsActivetrue}
\let\AppendixOriginalContentsLine\contentsline
\renewcommand{\contentsline}[4]{%
  \ifAppendixContentsActive
    \AppendixOriginalContentsLine{#1}{#2}{#3}{#4}%
  \fi}
\@starttoc{toc}
\makeatother
\endgroup
\clearpage

\section{Related Work}
\label{app:related-work}

\subsection{One-step generation and distribution matching}

One-step generation can be learned by approximating a sampling trajectory
or by directly matching the output distribution. The technique is widely useful in multimodal generation~\citep{blackforestlabs2026flux2klein}, editing~\citep{liu2026elasticttt}, and interaction~\citep{zhang2026interacvid}. Consistency models
\citep{song2023consistency} map points on the same probability-flow
trajectory to a common endpoint and support both distillation and
training without a teacher. MeanFlow~\citep{geng2025meanflow} instead
learns an average velocity over a time interval, using its relation to
the instantaneous velocity as a training target. Improved Mean Flows
\citep{geng2026imf} and pixel Mean Flows~\citep{lu2026pmf} further develop
this approach. These methods determine how a generator produces an image
in one step. Distributional post-training optimizes the distribution of
those images and can be applied to an already trained one-step model.
Our experiments use both pMF~\citep{lu2026pmf} and a one-step initialization from JiT
\citep{li2026jit}.

Distribution Matching Distillation (DMD)~\citep{yin2024dmd} expresses a
reverse-KL gradient through the difference between real and generated
scores at noisy image distributions. A pretrained diffusion model
provides the real score, while a separate diffusion model estimates the
generated score. Its original formulation also uses a regression loss
on teacher-generated pairs. DMD2~\citep{yin2024dmd2} removes this regression
requirement, uses two-timescale updates to improve the generated-score
estimate, and incorporates an adversarial loss on real images.
MGFlow also uses a difference of scores for KL matching, but computes
these scores from explicit Gaussian mixtures in frozen representation
spaces. It does not train diffusion score networks for the matching loss.

Kernel distribution matching predates recent one-step diffusion models.
Generative Moment Matching Networks~\citep{li2015gmmn} train a generator
with maximum mean discrepancy (MMD), a kernel two-sample criterion
\citep{gretton2012kernel}. A characteristic kernel can distinguish
distributions beyond their first two moments, whereas a Gaussian
Fr\'{e}chet objective depends only on means and covariances. Our GM
representation retains componentwise moments and assignments, allowing
distributional training to distinguish modes that a single Gaussian
cannot separate.

\subsection{Drifting and its gradient-flow interpretations}

Drifting~\citep{deng2026drifting} constructs a field from attraction to
real features and repulsion from generated features. The generator is
trained to regress its features toward detached field-shifted targets.
The iterative distribution update takes place during training; inference
uses a single generator evaluation. This separates training-time movement
of a distribution from the denoising trajectory used by a diffusion
sampler.

Several works analyze the relation between these fields and scores.
\citet{lai2026unifieddrifting} connect Gaussian-kernel mean shifts to
scores of smoothed densities and analyze the residual for more general
radial kernels. \citet{turan2026secretly} develop a spectral and variational
view, relating Gaussian-kernel Drifting to score differences and studying
the effect of bandwidth. \citet{cao2026gradientflowdrifting} formulate
Drifting through Wasserstein gradient flows of KDE-approximated
divergences, including extensions beyond KL. The choice of kernel and
normalization matters: \citet{franz2026nonconservative} show that general
normalized Drifting fields need not be conservative, with the Gaussian
kernel providing an exception. Our KDE--KL connection uses this
Gaussian-kernel setting; it does not identify every Drifting variant with
the same KL gradient flow.

Wasserstein gradient flows describe steepest descent of a distributional
energy under transport geometry~\citep{jordan1998variational,peyre2019computational}.
W-Flow~\citep{han2026wflow} applies this perspective to one-step generation
using the Sinkhorn divergence between empirical measures. Its field
subtracts the self-transport barycentric projection from the
cross-distribution projection. W-Flow and Gaussian-kernel Drifting thus
use different discrepancies even though both construct fields from
sample interactions. Table~\ref{tab:distribution-model-comparison} groups
them as sample-based methods, with empirical measures for W-Flow and KDE for
Gaussian-kernel Drifting. MGFlow instead estimates a finite collection of
component statistics and evaluates transport or score fields from them.

\subsection{Representation-based distributional post-training}

FD-Loss~\citep{yang2026fdloss} directly optimizes the Fr\'{e}chet distance
between real and generated feature statistics. The Gaussian distance has
a closed form~\citep{dowson1982frechet}, so training requires no learned
critic. Real statistics can be precomputed, while queues or exponential
moving averages provide generated statistics beyond a single batch.
Gradients pass through the current batch's contribution. Using several
frozen encoders extends the objective beyond a single representation.
Our Gaussian OT case recovers this moment-based objective. Replacing OT
by KL gives a different field with the same Gaussian representation;
using a GM changes the representation of the distribution itself.

Representation Distribution Matching (RDM)~\citep{feng2026rdm} organizes
visual generation around discrepancies between feature pushforward
distributions. Its iRDM implementation uses a Nystr\"om MMD estimator,
fixed reference statistics, and joint image--text matching for conditional
generation. 
Elastic Forcing~\citep{zhang2026ef} further expand this approach to videos, presenting a hybrid estimator balancing the computational-statistical tradeoff.
These works make clear that the representation, reference,
estimator, and discrepancy all affect training. Our study focuses on the
distribution model and its update: we compare Gaussian, GM, and
sample-based constructions, then address component assignment and
correspondence for GM training. For text-to-image generation, we follow
iRDM in concatenating image and text features for joint matching.

AdvFD~\citep{gao2026advfd} augments frozen representations with a learned
representation that maximizes the Fr\'{e}chet discrepancy while the
generator minimizes it. Whitening real features constrains the learned
representation and prevents a trivial increase through feature scaling.
Its adversarial update adapts the representation to the current generator.
MGFlow keeps the encoders fixed and instead refines the distribution model
within each representation.

Amortized Moment Matching (AMFD)~\citep{liu2026amfd} represents moment
matching through affine denoising operators and amortizes these operators
with neural networks. Its formulation supports representation spaces and
native generative spaces without explicitly maintaining all covariance
matrices in the loss. MGFlow uses explicit full-covariance components and
closed-form Gaussian scores or pair costs. The main additional problem is
then to assign new generated samples to components while keeping them
matched to the reference mixture.

Frozen visual representations are also used inside generative models.
REPA~\citep{yu2025repa} aligns diffusion-transformer hidden states with
pretrained features, while representation autoencoders
\citep{zheng2026rae} use pretrained representations as a generative latent
space. These uses differ from matching the distribution of generated
outputs in several feature spaces. The latter lets us change the training
objective without changing the generator's sampling architecture.

\subsection{Mixture transport and componentwise scores}

The Wasserstein distance between two Gaussians has a closed form, but the
same is not true for arbitrary Gaussian mixtures. \citet{delon2020gmmwasserstein}
define $MW_2$ by restricting the coupling to Gaussian component pairs.
The squared cost upper-bounds $W_2^2$ between the mixtures. MGFlow builds
on this discrepancy, but its LP acts at the sample level: it allocates
generated features to components with prescribed reference masses.
This differs from transporting mass between two already fitted mixtures.

For KL, mixture scores can be insensitive to mixing proportions
when components are well separated~\citep{wenliang2020blindness}.
MGFlow combines explicit mass constraints with paired component scores,
rather than relying on independently weighted marginal scores.
Appendix~\ref{app:method-paired-kl} relates this update to labelled KL.

\FloatBarrier

\section{Derivations and Proofs}
\label{app:proofs}
\subsection{Derivation of the KDE--KL interpretation of Drifting}
\label{app:drifting-derivation}

We derive the mean-shift--score identity and its KL interpretation in
Eq.~\eqref{eq:mg-instances}.

\paragraph{From the mean-shift field to the KDE score.}
For a fixed bandwidth $h>0$, let $k_h(z,y)=\mathcal N(z;y,h^2I)$ and define
\begin{equation}
  r_h(z)=\int k_h(z,y)\,\mathrm dr(y),
  \qquad r\in\{p,q_\theta\}.
  \label{eq:kde-definitions}
\end{equation}
Differentiating with respect to the query $z$, while holding $r$ fixed, gives
\begin{align}
  \nabla_z k_h(z,y)&=\frac{y-z}{h^2}k_h(z,y),\nonumber\\
  \nabla_z\log r_h(z)
  &=\frac{\int\nabla_z k_h(z,y)\,\mathrm dr(y)}
          {\int k_h(z,y)\,\mathrm dr(y)}\nonumber\\
  &=\frac{1}{h^2}
    \frac{\E_{y\sim r}[k_h(z,y)(y-z)]}{\E_{y\sim r}[k_h(z,y)]}
   =\frac{a_r(z)}{h^2}.
  \label{eq:mean-shift-score}
\end{align}
The Gaussian kernel and its first derivatives are bounded, so the
derivative can pass through the integral. The Gaussian-kernel drifting
field $v_{\mathrm{drift}}=a_p-a_{q_\theta}$ therefore satisfies
\begin{equation}
  v_{\mathrm{drift}}(z)
  =h^2\bigl[\nabla_z\log p_h(z)-\nabla_z\log q_h(z)\bigr]
\end{equation}
\citep{lai2026unifieddrifting,turan2026secretly,franz2026nonconservative}.

\paragraph{From KL to the score difference.}
For smooth positive densities $\rho$ and fixed $p_h$, write
$\mathcal E(\rho)=\int\rho\log(\rho/p_h)\,\mathrm dz$.
A mass-preserving perturbation $\rho+\varepsilon\eta$, with
$\int\eta\,\mathrm dz=0$, gives
\begin{align}
  \left.\frac{\mathrm d}{\mathrm d\varepsilon}
    \mathcal E(\rho+\varepsilon\eta)\right|_{\varepsilon=0}
  &=\int\left(\log\frac{\rho(z)}{p_h(z)}+1\right)\eta(z)\,\mathrm dz,
    \nonumber\\
  \frac{\delta\mathcal E}{\delta\rho}(z)
  &=\log\rho(z)-\log p_h(z)+1.
  \label{eq:app-kl-first-variation}
\end{align}
Taking the negative spatial gradient gives the Wasserstein velocity
\begin{equation}
  v_\rho(z)=-\nabla_z\frac{\delta\mathcal E}{\delta\rho}(z)
  =\nabla_z\log p_h(z)-\nabla_z\log\rho(z)
  \label{eq:app-kl-velocity}
\end{equation}
\citep{jordan1998variational}. Evaluating at $\rho=q_h$ proves
$v_{\mathrm{drift}}=h^2v_{\mathrm{KL}}$
\citep{cao2026gradientflowdrifting}.

This is the KL field evaluated at the smoothed densities. For comparison,
varying the composite functional $\mathcal G(q)=\KL(k_h*q\Vert p_h)$
with respect to $q$ gives
\begin{equation}
  \frac{\delta\mathcal G}{\delta q}(x)
  =\int k_h(z,x)\left(\log\frac{q_h(z)}{p_h(z)}+1\right)\,\mathrm dz.
  \label{eq:app-smoothed-kl-variation}
\end{equation}
Here $\delta q_h(z)=\int k_h(z,x)\,\delta q(x)\,\mathrm dx$ introduces
an additional smoothing operation; its negative gradient is not generally
the field in Eq.~\eqref{eq:app-kl-velocity}.

\paragraph{Detached-target regression.}
Drifting uses the regression loss \citep{deng2026drifting}
\begin{equation}
  \mathcal L_{\mathrm{drift}}
  =\frac{1}{B}\sum_i
    \left\|z_i-\operatorname{sg}[z_i+v_{\mathrm{drift}}(z_i)]\right\|_2^2.
  \label{eq:drifting-loss}
\end{equation}
The general calculation in Appendix~\ref{app:energy-descent} gives
$\nabla_{z_i}\mathcal L_{\mathrm{drift}}=-2v_{\mathrm{drift}}(z_i)/B$.

\subsection{Derivation of the Gaussian OT interpretation of FD-Loss}
\label{app:fd-derivation}

We derive the Gaussian transport cost, its feature gradient, and the
finite-sample factors underlying the FD field in Eq.~\eqref{eq:mg-instances}.

\paragraph{The Gaussian optimal transport cost and map.}
Let $q_{\mathrm G}=\mathcal N(\mu_q,\Sigma_q)$ and
$p_{\mathrm G}=\mathcal N(\mu_p,\Sigma_p)$, with positive-definite
covariances. All matrix square roots below are symmetric positive definite.
The optimal Gaussian transport map is
\begin{equation}
  T(z)=\mu_p+A(z-\mu_q),\qquad
  A=\Sigma_q^{-1/2}
    (\Sigma_q^{1/2}\Sigma_p\Sigma_q^{1/2})^{1/2}\Sigma_q^{-1/2}.
  \label{eq:gaussian-ot-map}
\end{equation}
Indeed, $A\Sigma_q A=\Sigma_p$, so $T$ sends $q_{\mathrm G}$ to
$p_{\mathrm G}$. Since $A$ is positive definite, $T$ is the gradient of
a convex quadratic and is optimal for quadratic transport
\citep{peyre2019computational}. Writing $z-T(z)=(\mu_q-\mu_p)+(I-A)(z-\mu_q)$,
we obtain
\begin{align}
  W_2^2(q_{\mathrm G},p_{\mathrm G})
  &=\E_{q_{\mathrm G}}\|z-T(z)\|_2^2\nonumber\\
  &=\|\mu_q-\mu_p\|_2^2+
    \operatorname{tr}\bigl((I-A)\Sigma_q(I-A)\bigr)\nonumber\\
  &=\|\mu_q-\mu_p\|_2^2+
    \operatorname{tr}(\Sigma_q+\Sigma_p-2A\Sigma_q)\nonumber\\
  &=\|\mu_q-\mu_p\|_2^2+
    \operatorname{tr}\!\left(\Sigma_q+\Sigma_p
      -2(\Sigma_q^{1/2}\Sigma_p\Sigma_q^{1/2})^{1/2}\right)
   =\mathcal D_{\mathrm{FD}}.
  \label{eq:fd-is-w2}
\end{align}
The third line uses $A\Sigma_q A=\Sigma_p$; the last uses cyclicity
of the trace. This is the Gaussian Fr\'echet formula
\citep{dowson1982frechet}.

\paragraph{Differentiating the mean and covariance.}
Set $F=\tfrac12\mathcal D_{\mathrm{FD}}$, keeping the reference moments fixed.
The mean term gives $\nabla_{\mu_q}F=\mu_q-\mu_p$.
For the covariance term, use the equivalent cross term
$\operatorname{tr}(M^{1/2})$, where
$M=\Sigma_p^{1/2}\Sigma_q\Sigma_p^{1/2}$.
The two cross-term matrices are $BB^{\mathsf T}$ and $B^{\mathsf T}B$
for $B=\Sigma_q^{1/2}\Sigma_p^{1/2}$, so they have the same eigenvalues.
To differentiate its trace, write $S=M^{1/2}$ and differentiate $S^2=M$:
\begin{equation}
  S\,\mathrm dS+\mathrm dS\,S=\mathrm dM
  \quad\Longrightarrow\quad
  \mathrm d\operatorname{tr}(M^{1/2})
  =\tfrac12\operatorname{tr}(M^{-1/2}\mathrm dM).
\end{equation}
The implication follows by multiplying by $S^{-1}$ and taking the trace;
it does not require $M$ and $\mathrm dM$ to commute. Therefore
\begin{align}
  \mathrm d_{\Sigma_q}F
  &=\tfrac12\operatorname{tr}(\mathrm d\Sigma_q)
    -\tfrac12\operatorname{tr}
      (M^{-1/2}\Sigma_p^{1/2}\,\mathrm d\Sigma_q\,\Sigma_p^{1/2})
      \nonumber\\
  &=\tfrac12\operatorname{tr}\bigl((I-A)\,\mathrm d\Sigma_q\bigr).
\end{align}
Here $\Sigma_p^{1/2}M^{-1/2}\Sigma_p^{1/2}=A$: both are positive-definite
solutions of $X\Sigma_qX=\Sigma_p$, whose unique solution follows by
squaring $\Sigma_q^{1/2}X\Sigma_q^{1/2}$. Thus
\begin{equation}
  \nabla_{\mu_q}F=\mu_q-\mu_p,\qquad
  \nabla_{\Sigma_q}F=\tfrac12(I-A).
  \label{eq:fd-moment-gradient}
\end{equation}

\paragraph{From moment derivatives to feature movement.}
Let $q$ be any generated distribution with finite second moments and
positive-definite covariance. Perturb its features by
$z_\varepsilon=z+\varepsilon u(z)$, and denote their distribution by
$q_\varepsilon$. Differentiating the mean and
$\Sigma_q=\E_q[zz^{\mathsf T}]-\mu_q\mu_q^{\mathsf T}$ at $\varepsilon=0$ gives
\begin{align}
  \dot\mu_q&=\E_q[u(z)],\nonumber\\
  \dot\Sigma_q
  &=\E_q[u(z)z^{\mathsf T}+zu(z)^{\mathsf T}]
    -\dot\mu_q\mu_q^{\mathsf T}-\mu_q\dot\mu_q^{\mathsf T}\nonumber\\
  &=\E_q[u(z)(z-\mu_q)^{\mathsf T}+(z-\mu_q)u(z)^{\mathsf T}].
\end{align}
Combining these with Eq.~\eqref{eq:fd-moment-gradient} yields
\begin{align}
  \left.\frac{\mathrm d}{\mathrm d\varepsilon}
    F(\mu_{q_\varepsilon},\Sigma_{q_\varepsilon})\right|_0
  &=(\mu_q-\mu_p)^{\mathsf T}\dot\mu_q
    +\tfrac12\operatorname{tr}((I-A)\dot\Sigma_q)\nonumber\\
  &=\E_q\!\left[
    \bigl(\mu_q-\mu_p+(I-A)(z-\mu_q)\bigr)^{\mathsf T}u(z)\right]\nonumber\\
  &=\E_q[(z-T(z))^{\mathsf T}u(z)].
  \label{eq:fd-feature-variation}
\end{align}
The two covariance terms combine because $I-A$ is symmetric.
Hence the moment objective has descent field $T(z)-z$.
When $q=q_{\mathrm G}$, this is the Wasserstein velocity of
$\tfrac12W_2^2(q,p_{\mathrm G})$:
\begin{equation}
  v_{\mathrm{FD},t}(z)=T_{q_{\mathrm G,t}\to p_{\mathrm G}}(z)-z,\qquad
  \partial_tq_{\mathrm G,t}+\nabla\cdot(q_{\mathrm G,t}v_{\mathrm{FD},t})=0.
  \label{eq:fd-wgf}
\end{equation}
The affine field preserves Gaussianity. For non-Gaussian $q$, the same
feature gradient differentiates the moment objective, but $T$ need only
match the reference mean and covariance, not the full distribution.

Setting $v=v_{\mathrm{FD}}$ and $\eta=1$ in the regression formula of
Appendix~\ref{app:energy-descent} gives the target $\operatorname{sg}[T(z_i)]$
and gradient $(z_i-T(z_i))/B$. This recovers the same field; using
$\mathcal D_{\mathrm{FD}}=2F$ instead of $F$ multiplies its gradient by two.

\paragraph{Finite-sample and EMA factors.}
For a feature $z_i$ contributing weight $c_i$ to the estimated mean
and raw second moment, while all other contributions are fixed,
\begin{equation}
  \mathrm d\mu_q=c_i\,\mathrm dz_i,\qquad
  \mathrm d\Sigma_q=c_i\bigl[
    \mathrm dz_i(z_i-\mu_q)^{\mathsf T}
    +(z_i-\mu_q)\mathrm dz_i^{\mathsf T}\bigr].
\end{equation}
Substitution into Eq.~\eqref{eq:fd-moment-gradient} gives
\begin{equation}
  \nabla_{z_i}F
  =c_i\bigl[\mu_q-\mu_p+(I-A)(z_i-\mu_q)\bigr]
  =c_i(z_i-T(z_i)).
  \label{eq:app-fd-estimator-gradient}
\end{equation}
For moments averaged over $N$ features, $c_i=1/N$; detached queue
entries affect the moments but receive no gradient. For EMA updates
of the mean and raw second moment with decay $\beta$ and current
batch size $B$, $c_i=(1-\beta)/B$, with $T$ evaluated at the updated
moments \citep{yang2026fdloss}.

If the empirical covariance instead uses
$\Sigma_q=(N-1)^{-1}\sum_j(z_j-\mu_q)(z_j-\mu_q)^{\mathsf T}$,
its derivative has coefficient $1/(N-1)$, while the mean retains $1/N$:
\begin{equation}
  \nabla_{z_i}F
  =\frac{\mu_q-\mu_p}{N}
   +\frac{(I-A)(z_i-\mu_q)}{N-1}.
\end{equation}
Here the terms from differentiating the centering cancel because
$\sum_j(z_j-\mu_q)=0$.

\subsection{Global KL Velocity for Gaussian Mixtures}
\label{app:method-derivation}
\label{app:method-global-kl}

Let $P$ and $Q$ have positive mixture weights and positive-definite
component covariances. Applying Eq.~\eqref{eq:app-kl-velocity} to these densities
gives $v_{\mathrm{global}}=\nabla\log P-\nabla\log Q$.
To evaluate each mixture score, differentiate
$Q(z)=\sum_k\omega_kq_k(z)$:
\begin{align}
  \nabla_z\log Q(z)
  &=\frac{\sum_k\omega_k\nabla_zq_k(z)}{Q(z)}\nonumber\\
  &=\sum_k\frac{\omega_kq_k(z)}{Q(z)}\nabla_z\log q_k(z)
   =\sum_k\gamma_{Q,k}(z)s_{q,k}(z).
  \label{eq:app-mixture-score}
\end{align}
For $q_k=\mathcal N(\mu_{q,k},\Sigma_{q,k})$,
\begin{equation}
  s_{q,k}(z)
  =-\tfrac12\nabla_z\bigl[(z-\mu_{q,k})^{\mathsf T}
                          \Sigma_{q,k}^{-1}(z-\mu_{q,k})\bigr]
  =\Sigma_{q,k}^{-1}(\mu_{q,k}-z).
\end{equation}
Applying the same calculation to $P$ yields Eq.~\eqref{eq:mg-global-kl}.
The two scores use their respective posterior responsibilities
$\gamma_{Q,k}=\omega_kq_k/Q$ and $\gamma_{P,k}=\pi_kp_k/P$.

\subsection{Component Matching and Paired Updates}
\label{app:method-paired-derivation}

\paragraph{Mixture transport and its paired upper bound.}
For each component pair, let $\gamma_{ij}$ be an optimal coupling of
$q_i$ and $p_j$, with cost $C_{ij}=W_2^2(q_i,p_j)$.
For any $\Gamma\in U(\omega,\pi)$,
$\gamma=\sum_{i,j}\Gamma_{ij}\gamma_{ij}$ has marginals
$\sum_i\omega_iq_i=Q$ and $\sum_j\pi_jp_j=P$.
Its transport cost is $\sum_{i,j}\Gamma_{ij}C_{ij}$, so minimizing over
$\Gamma$ gives $W_2^2(Q,P)\leq MW_2^2(Q,P)$
\citep{delon2020gmmwasserstein}.
When $\omega=\pi$, the feasible coupling $\operatorname{diag}(\pi)$ gives
\begin{equation}
  W_2^2(Q,P)\leq MW_2^2(Q,P)\leq\sum_k\pi_kC_{kk},
\end{equation}
which proves the two bounds in Eq.~\eqref{eq:method-mb-pair}.

\paragraph{When is diagonal component transport optimal?}
\label{app:method-diagonal}
\label{app:diagonal-training}
Suppose $C_{ii}\leq C_{ij}$ for every $i,j$ and both mixture weights are
$\pi$. For any feasible $\Gamma$, its row sums give
\begin{equation}
  \sum_{i,j}\Gamma_{ij}C_{ij}-\sum_i\pi_iC_{ii}
  =\sum_{i,j}\Gamma_{ij}(C_{ij}-C_{ii})\geq0.
\end{equation}
The diagonal coupling attains equality and is therefore optimal.
It is the unique optimum if $C_{ii}<C_{ij}$ for all $j\ne i$.
Matched weights alone do not imply this condition: swapping the labels
of two equally weighted, distinct Gaussians makes off-diagonal pairing
optimal. 

Empirically, we inspected a pMF-H run using SIM encoders, $K=1+4$, and a global
batch of 1,024. For each encoder, the training code forms all 16
Gaussian costs and solves the component transport LP every 100 updates.
The logs cover 34.8 epochs and contain 435 distinct plan refreshes per
encoder. Every recorded plan has diagonal mass $1$ and exactly four
nonzero entries, giving $\Gamma=\operatorname{diag}(\pi)$ in all three
feature spaces.

\paragraph{Paired KL and the marginal KL bound.}
\label{app:method-paired-kl}
For fixed positive weights $\pi_k$ summing to one, introduce the labelled
distributions $\widetilde Q(k,z)=\pi_kq_k(z)$ and
$\widetilde P(k,z)=\pi_kp_k(z)$. Since the weights cancel inside the ratio,
\begin{equation}
  \KL(\widetilde Q\Vert\widetilde P)
  =\sum_k\int\pi_kq_k(z)\log\frac{\pi_kq_k(z)}{\pi_kp_k(z)}\,\mathrm dz
  =\sum_k\pi_k\KL(q_k\Vert p_k)
  =\mathcal F_{\mathrm{pair}}.
  \label{eq:method-labelled-kl}
\end{equation}
To relate this to marginal KL, factor
$\widetilde Q(k,z)=Q(z)\gamma_{Q,k}(z)$ and
$\widetilde P(k,z)=P(z)\gamma_{P,k}(z)$. Then
\begin{align}
  \mathcal F_{\mathrm{pair}}
  &=\sum_k\int Q(z)\gamma_{Q,k}(z)
     \left[\log\frac{Q(z)}{P(z)}
          +\log\frac{\gamma_{Q,k}(z)}{\gamma_{P,k}(z)}\right]\,\mathrm dz
     \nonumber\\
  &=\KL(Q\Vert P)
    +\E_{z\sim Q}\!\left[
      \sum_k\gamma_{Q,k}(z)\log
        \frac{\gamma_{Q,k}(z)}{\gamma_{P,k}(z)}\right]\nonumber\\
  &=\KL(Q\Vert P)
    +\E_{z\sim Q}\!\left[
      \KL\!\left(\gamma_Q(\cdot\mid z)\Vert\gamma_P(\cdot\mid z)\right)
      \right]\geq\KL(Q\Vert P).
  \label{eq:method-kl-chain-rule}
\end{align}
The second line uses $\sum_k\gamma_{Q,k}=1$.
Equality holds exactly when the posterior label distributions agree
$Q$-almost everywhere.

\paragraph{From the paired energy to component fields.}
Let $u_k$ move component $q_k$, and set
$g_k=\nabla_z\log(q_k/p_k)$.
Using the KL variation in Eq.~\eqref{eq:app-kl-first-variation} and
$\partial_tq_k=-\nabla\cdot(q_ku_k)$ gives
\begin{equation}
  \frac{\mathrm d}{\mathrm dt}\mathcal F_{\mathrm{pair}}
  =\sum_k\pi_k\E_{q_k}[g_k^{\mathsf T}u_k],
\end{equation}
with vanishing boundary terms. The squared product Wasserstein distance
$\sum_k\pi_kW_2^2(q_k,q'_k)$ corresponds to squared speed
$\sum_k\pi_k\E_{q_k}\|u_k\|_2^2$. Its steepest-descent velocity minimizes
\begin{align}
  \sum_k\pi_k\E_{q_k}
    \left[g_k^{\mathsf T}u_k+\tfrac12\|u_k\|_2^2\right]
  =\frac12\sum_k\pi_k\E_{q_k}
    \left[\|u_k+g_k\|_2^2-\|g_k\|_2^2\right].
\end{align}
The minimum is attained at
\begin{equation}
  v_k(z)=-g_k(z)=\nabla_z\log p_k(z)-\nabla_z\log q_k(z).
  \label{eq:method-component-kl-field}
\end{equation}
Thus the common weights in the energy and metric do not introduce an
extra factor $\pi_k$ into the component velocity
\citep{jordan1998variational,peyre2019computational}.

The regularized scores in Eq.~\eqref{eq:method-paired-score} are those of
$r_k^\lambda=\mathcal N(\mu_{r,k},\Sigma_{r,k}+\lambda I)$.
Training combines their differences with the detached LP assignments
$R^*_{nk}$ and applies the regression gradient derived in
Appendix~\ref{app:energy-descent}. Unlike the marginal KL velocity in
Eq.~\eqref{eq:mg-global-kl}, this update uses $R^*_{nk}$ for both scores
rather than separate mixture posteriors.

\subsection{Energy descent and detached-target regression}
\label{app:energy-descent}

\paragraph{Continuous distributional descent.}
Let $\phi_t=\left.\delta\mathcal F/\delta Q\right|_{Q=Q_t}$, where
$t$ denotes continuous flow time. For a smooth density satisfying
$\partial_tQ_t+\nabla\cdot(Q_tv_t)=0$, assume sufficient decay or
no-flux boundary conditions so that the boundary term below vanishes.
The chain rule and integration by parts give
\citep{jordan1998variational,peyre2019computational}
\begin{align}
  \frac{\mathrm d}{\mathrm dt}\mathcal F(Q_t)
  &=\int\phi_t(z)\,\partial_tQ_t(z)\,\mathrm dz\nonumber\\
  &=-\int\phi_t(z)\nabla\cdot(Q_t(z)v_t(z))\,\mathrm dz\nonumber\\
  &=\int Q_t(z)\nabla\phi_t(z)^{\mathsf T}v_t(z)\,\mathrm dz.
\end{align}
Substituting $v_t=-\nabla\phi_t$ proves Eq.~\eqref{eq:mg-wgf}:
\begin{equation}
  \frac{\mathrm d}{\mathrm dt}\mathcal F(Q_t)
  =-\E_{z\sim Q_t}\|v_t(z)\|_2^2\leq0.
\end{equation}

\paragraph{Applying a field to generator training.}
For $z_n=E(G_\theta(\xi_n,c_n),c_n)$, define
$\bar z_n=\operatorname{sg}[z_n+\eta v_n]$ and
$\mathcal L_v=(2B)^{-1}\sum_n\|z_n-\bar z_n\|_2^2$.
The entire target is constant during differentiation, so
\begin{equation}
  \nabla_{z_n}\mathcal L_v
  =\frac{z_n-\bar z_n}{B}=-\frac{\eta}{B}v_n,\qquad
  \nabla_\theta\mathcal L_v
  =-\frac{\eta}{B}\sum_n J_n^{\mathsf T}v_n,
  \quad J_n=\frac{\partial z_n}{\partial\theta}.
  \label{eq:app-regression-gradient}
\end{equation}
No derivative of the estimated field enters this gradient and the regression transfers the field through the encoder and generator
Jacobians.

\subsection{Conditional image scores in a joint Gaussian}
\label{app:joint-score-math}

Let $p(z,u)$ be a joint Gaussian over image features $z$ and text
features $u$, with
\begin{equation}
  \mu=\begin{bmatrix}\mu_z\\\mu_u\end{bmatrix},\qquad
  \Sigma=\begin{bmatrix}
    \Sigma_{zz}&\Sigma_{zu}\\
    \Sigma_{uz}&\Sigma_{uu}
  \end{bmatrix}\succ0.
\end{equation}
Write $a=z-\mu_z$, $b=u-\mu_u$, and
$[\xi_z;\xi_u]=\Sigma^{-1}[a;b]$.
The Gaussian image score is $-\xi_z$. Instead of forming the full inverse,
solve the block equations:
\begin{align}
  \Sigma_{zz}\xi_z+\Sigma_{zu}\xi_u&=a,\nonumber\\
  \Sigma_{uz}\xi_z+\Sigma_{uu}\xi_u&=b.
\end{align}
The second equation gives
$\xi_u=\Sigma_{uu}^{-1}(b-\Sigma_{uz}\xi_z)$.
Substituting into the first yields
\begin{equation}
  \underbrace{(\Sigma_{zz}-\Sigma_{zu}\Sigma_{uu}^{-1}\Sigma_{uz})}
    _{\Sigma_{z\mid u}}\xi_z
  =a-\Sigma_{zu}\Sigma_{uu}^{-1}b.
\end{equation}
Define
$\mu_{z\mid u}=\mu_z+\Sigma_{zu}\Sigma_{uu}^{-1}(u-\mu_u)$.
We obtain
\begin{equation}
  \nabla_z\log p(z,u)
  =-\xi_z=\Sigma_{z\mid u}^{-1}(\mu_{z\mid u}-z)
  =\nabla_z\log p(z\mid u).
  \label{eq:app-joint-precision}
\end{equation}
The last equality also follows from
$\log p(z,u)=\log p(z\mid u)+\log p(u)$, since $u$ is fixed.
For paired GM matching, we apply this calculation to each joint Gaussian
component and combine the image-score differences with $R^*_{nk}$.
Image--text cross-covariance therefore affects the image update even
though text features receive no gradient.

\FloatBarrier
\clearpage
\section{Implementation}
\label{app:setup}

\subsection{Algorithms}
\label{app:training-loop}

We fit the real-data distributions once, initialize generated statistics
from the pretrained generator, and then update the generator and its
statistics jointly. Each encoder and component count has a separate
reference and moment state. A $K=1+4$ configuration therefore contains
one Gaussian branch and one four-component branch.
Algorithms~\ref{alg:reference-fit} and~\ref{alg:statistics-init} are run
once for each encoder--branch pair; Algorithm~\ref{alg:paired-training}
is repeated at each training step. We write $E(x,c)$ for the features
being matched: image features for image-only training, or concatenated
image and text features for joint matching.

\paragraph{Reference fitting.}
Algorithm~\ref{alg:reference-fit} gives the full-data EM update
\citep{dempster1977em}. Features are cached and processed in blocks, but
parameters are updated only after a complete pass. The saved covariance
is the raw centered second moment.

\begin{algorithm}[!htbp]
\caption{Full-covariance reference GM fitting}
\label{alg:reference-fit}
\small
\begin{algorithmic}[1]
\Require Cached real features $Y=\{y_n\}_{n=1}^N$, component count $K$,
initial parameters $(\pi,\mu,\Sigma)$, maximum EM iterations $T_{\max}$
\Ensure Frozen reference $P=\sum_{k=1}^K\pi_k\mathcal N(\mu_k,\Sigma_k)$
\If{$K=1$}
  \State $\mu_1\gets N^{-1}\sum_n y_n$;
  $\Sigma_1\gets N^{-1}\sum_n y_ny_n^{\mathsf T}-\mu_1\mu_1^{\mathsf T}$
  \State \Return $P=\mathcal N(\mu_1,\Sigma_1)$, with $\pi_1=1$
\EndIf
\For{$t=1,\ldots,T_{\max}$}
  \State $(\pi^{\mathrm{old}},\mu^{\mathrm{old}},\Sigma^{\mathrm{old}})\gets(\pi,\mu,\Sigma)$
  \State Set $N_k=0$, $b_k=0$, $S_k=0$ for all $k$ \Comment{Reset full-data accumulators}
  \For{each feature block with indices $\mathcal I$}
    \State For all $n\in\mathcal I$ and $k$, compute the E-step responsibilities:
    \Statex \hspace{\algorithmicindent}\hspace{\algorithmicindent}\hspace{\algorithmicindent}
    $r_{nk}\gets
    \displaystyle\frac{\pi_k^{\mathrm{old}}\mathcal N(y_n;\mu_k^{\mathrm{old}},\Sigma_k^{\mathrm{old}})}
    {\sum_j\pi_j^{\mathrm{old}}\mathcal N(y_n;\mu_j^{\mathrm{old}},\Sigma_j^{\mathrm{old}})}$
    \State For all $k$, accumulate $N_k\gets N_k+\sum_{n\in\mathcal I}r_{nk}$
    \State $b_k\gets b_k+\sum_{n\in\mathcal I}r_{nk}y_n$;
    $S_k\gets S_k+\sum_{n\in\mathcal I}r_{nk}y_ny_n^{\mathsf T}$ for all $k$
  \EndFor
  \State For all $k$, perform the M-step after processing all $N$ features:
  \Statex \hspace{\algorithmicindent}\hspace{\algorithmicindent}
  $\pi_k\gets N_k/N$;
  $\mu_k\gets b_k/N_k$;
  $\Sigma_k\gets S_k/N_k-\mu_k\mu_k^{\mathsf T}$
  \State Compute parameter changes relative to $(\pi^{\mathrm{old}},\mu^{\mathrm{old}},\Sigma^{\mathrm{old}})$
  \If{the convergence test below passes}
    \State \textbf{break}
  \EndIf
\EndFor
\State \Return frozen reference $P=\sum_k\pi_k\mathcal N(\mu_k,\Sigma_k)$
\end{algorithmic}
\end{algorithm}

For $K>1$, we initialize centers by $k$-means++ on 65,536 sampled features
with seed 3407, then run Lloyd updates on the full feature bank for
8--20 iterations, stopping when the maximum relative center shift is
below $10^{-4}$. The resulting hard assignments initialize component
weights, means, and full covariances. We allow at most 96 full-data EM
iterations for ImageNet and 384 for the T2I references.
We check maximum absolute weight change and relative RMS changes in means
and covariances, using thresholds $0.002$, $0.006$, and $0.020$,
respectively. The test must pass on two consecutive updates without an
increase in these changes. ImageNet fits are additionally checked by one
further full-data EM update. We use the terminal seed-3407 fits as fixed
references. The retained Inception $K=16$ fit meets the parameter-change
thresholds but has a maximum-to-minimum weight ratio of $58.62$, exceeding
the fitting diagnostic threshold of 10.

\paragraph{Generated statistics.}
\label{app:method-statistics}
For assignments $R_{nk}$, define the batch statistics
$\widehat a_k=B^{-1}\sum_nR_{nk}$,
$\widehat b_k=B^{-1}\sum_nR_{nk}z_n$, and
$\widehat S_k=B^{-1}\sum_nR_{nk}z_nz_n^{\mathsf T}$.
The stored state is $H=\{a_k,b_k,S_k\}_k$, from which
$\mu_{q,k}=b_k/a_k$ and
$\Sigma_{q,k}=S_k/a_k-\mu_{q,k}\mu_{q,k}^{\mathsf T}$.
Algorithm~\ref{alg:statistics-init} initializes this state without a
generator update. Subsequent updates use an EMA, following FD-Loss
\citep{yang2026fdloss}.

For the LP assignments $R^*$ in Eq.~\eqref{eq:method-batch-lp},
the capacity constraint fixes $a_k=\widehat a_k=\pi_k$.
The conditional batch mean and raw second moment are therefore
\begin{equation}
  \widehat\mu_k=\frac{1}{B\pi_k}\sum_n R^*_{nk}z_n,
  \qquad
  \widehat M_k=\frac{1}{B\pi_k}\sum_n R^*_{nk}z_nz_n^{\mathsf T}.
  \label{eq:method-batch-moments}
\end{equation}
Writing $M_{q,k}=S_k/a_k$, the EMA with decay $\beta$ gives
\begin{equation}
  \mu_{q,k}^{+}=\beta\mu_{q,k}+(1-\beta)\widehat\mu_k,
  \qquad
  M_{q,k}^{+}=\beta M_{q,k}+(1-\beta)\widehat M_k,
  \qquad
  \Sigma_{q,k}^{+}=M_{q,k}^{+}-\mu_{q,k}^{+}(\mu_{q,k}^{+})^{\mathsf T}.
  \label{eq:method-ema-moments}
\end{equation}

\begin{algorithm}[!htbp]
\caption{Warm-starting generated component statistics}
\label{alg:statistics-init}
\small
\begin{algorithmic}[1]
\Require Pretrained generator $G_{\theta_0}$, frozen encoder $E$,
reference $P=\sum_k\pi_kp_k$, sample budget $N_0$, assignment batch size $B_0$
\Ensure Initial moment state $H=\{a_k,b_k,S_k\}_{k=1}^K$; no generator update
\State Disable gradient recording; set processed count $m=0$
\State Set totals $A_k=0$, $U_k=0$, $V_k=0$ for all $k$
\While{$m<N_0$}
  \State $B\gets\min(B_0,N_0-m)$ \Comment{Include the final partial batch}
  \State Sample $\{(\xi_n,c_n)\}_{n=1}^B$ using the training sampling scheme
  \State Compute $z_n=E(G_{\theta_0}(\xi_n,c_n),c_n)$ for all $n$
  \If{$K=1$}
    \State Set $R_{n1}=1$ for every $n$
  \Else
    \State $C_{nk}\gets-\log(\pi_kp_k(z_n))$ for all $n,k$
    \State Solve $R\in\arg\min_{R\geq0}\sum_{n,k}R_{nk}C_{nk}$
    \Statex \hspace{\algorithmicindent}\hspace{\algorithmicindent}\hspace{\algorithmicindent}
    subject to $R\mathbf1_K=\mathbf1_B$ and $R^{\mathsf T}\mathbf1_B=B\pi$
  \EndIf
  \State For all $k$, accumulate $A_k\gets A_k+\sum_nR_{nk}$;
  $U_k\gets U_k+\sum_nR_{nk}z_n$
  \State $V_k\gets V_k+\sum_nR_{nk}z_nz_n^{\mathsf T}$ for all $k$;
  $m\gets m+B$
\EndWhile
\State For all $k$, set $(a_k,b_k,S_k)\gets(A_k,U_k,V_k)/N_0$
\State \Return $H=\{a_k,b_k,S_k\}_k$ \Comment{$a_k=\pi_k$ by the LP constraints}
\end{algorithmic}
\end{algorithm}

We use $B_0=1{,}024$ and $N_0=50{,}000$. Encoder microbatches are
collected into these assignment batches before solving the LP.
Initialization averages the accumulated moments.

\paragraph{LP-paired training.}
We solve the global-batch assignment LP with SciPy's HiGHS backend
(\texttt{linprog}, \texttt{method='highs'}). The FP64 solve runs on rank
zero with primal and dual feasibility tolerances of $10^{-10}$;
the plan is then broadcast to all ranks.
Before solving, we subtract each row's minimum cost and divide by the
median positive cost. We require maximum absolute row- and column-sum
residuals below $10^{-8}$ in the original assignment units.
Moment accumulation, EMA states, covariance matrices, and their
factorizations use FP64, independently of neural-network mixed precision.

Algorithm~\ref{alg:paired-training} distinguishes the two gradient paths.
OT differentiates through the current batch in the candidate EMA state.
KL evaluates the field using the stored state before its update and
detaches the regression target. In both cases the assignment is fixed
during backpropagation, and the EMA is committed once per global batch.
The Gaussian OT cost uses raw covariances, clipping numerically negative
eigenvalues to zero when evaluating matrix square roots.
Here $\operatorname{sg}$ stops gradients without changing values.
Encoder parameters remain fixed, but gradients pass through their image
inputs to the generator. Within each encoder--branch iteration below,
we omit the indices $(e,K)$ on component parameters and moment entries.

\begin{algorithm}[!htbp]
\caption{One generator update with LP-paired OT or KL}
\label{alg:paired-training}
\label{fig:app-algorithm}
\small
\begin{algorithmic}[1]
\Require Generator $G_\theta$ and its optimizer; frozen encoders $E_e$ and references $P_{e,K}$
\Require Stored states $H_{e,K}$, branch counts $K\in\mathcal K$, weights $w_{e,K}$,
global batch size $B$, EMA decay $\beta_t$
\Require Matching objective (OT or KL); for KL, score ridges $\lambda_{e,K}$ and field scale $\eta$
\Ensure Updated generator parameters and one EMA update per encoder--branch pair
\State Clear generator gradients; set $\mathcal L\gets0$
\State Sample $\{(\xi_n,c_n)\}_{n=1}^B$; generate $x_n=G_\theta(\xi_n,c_n)$ once
\State For each encoder $e$, compute $z_{e,n}=E_e(x_n,c_n)$ for all $n$
\For{each encoder $e$ and branch $K$}
  \State Set $z_n=z_{e,n}$ and read the fixed reference $P_{e,K}=\sum_k\pi_kp_k$
  \If{$K=1$}
    \State Set $R_{n1}=1$ for every $n$
  \Else
    \State Compute $C_{nk}=-\log(\pi_kp_k(\operatorname{sg}[z_n]))$ and solve Eq.~\eqref{eq:method-batch-lp}
  \EndIf
  \State $R\gets\operatorname{sg}[R]$ \Comment{Reuse this assignment for moments and paired fields}
  \State $\widehat a_k\gets B^{-1}\sum_nR_{nk}$;
  $\widehat b_k\gets B^{-1}\sum_nR_{nk}z_n$ for all $k$
  \State $\widehat S_k\gets B^{-1}\sum_nR_{nk}z_nz_n^{\mathsf T}$ for all $k$
  \State $H^+_{e,K}\gets\beta_t\operatorname{sg}[H_{e,K}]+(1-\beta_t)(\widehat a,\widehat b,\widehat S)$
  \Statex \hspace{\algorithmicindent}\hspace{\algorithmicindent}
  \textit{Keep $H_{e,K}$ unchanged until after the optimizer step.}
  \If{OT matching}
    \State $\mu^+_{q,k}\gets b_k^+/a_k^+$;
    $\Sigma^+_{q,k}\gets S_k^+/a_k^+-\mu^+_{q,k}(\mu^+_{q,k})^{\mathsf T}$ for all $k$
    \State $\ell_{e,K}\gets\sum_k\pi_k
    W_2^2(\mathcal N(\mu^+_{q,k},\Sigma^+_{q,k}),p_k)$
  \Else \Comment{KL matching}
    \State From stored $H_{e,K}$, set $\mu_{q,k}=b_k/a_k$ and
    $\Sigma_{q,k}=S_k/a_k-\mu_{q,k}\mu_{q,k}^{\mathsf T}$
    \State Without gradients, solve $(\Sigma_{r,k}+\lambda_{e,K}I)s_{r,k,n}=\mu_{r,k}-z_n$
    \Statex \hspace{\algorithmicindent}\hspace{\algorithmicindent}\hspace{\algorithmicindent}
    for $r\in\{p,q\}$, all $k,n$, using Cholesky factors shared across samples
    \State $v_n\gets\sum_kR_{nk}(s_{p,k,n}-s_{q,k,n})$;
    $\bar z_n\gets\operatorname{sg}[z_n+\eta v_n]$
    \State $\ell_{e,K}\gets(2B)^{-1}\sum_n\|z_n-\bar z_n\|_2^2$
  \EndIf
  \State $\mathcal L\gets\mathcal L+w_{e,K}\ell_{e,K}$
\EndFor
\State Backpropagate $\mathcal L$ to $\theta$ and take one optimizer step
\State Commit $H_{e,K}\gets\operatorname{sg}[H^+_{e,K}]$ for every encoder and branch
\end{algorithmic}
\end{algorithm}

\paragraph{Global-batch gradients with limited memory.}
When a global batch spans several devices or accumulation passes, we first
collect detached features and solve the LP on the complete batch.
We compute the loss gradient with respect to these features, then replay
each generator microbatch with the same noise, conditions, and random
state. Sequential encoder vector--Jacobian products propagate its slice
of the feature gradient to the generator~\citep{gao2021scaling}. The feature loss is already
normalized by the global batch size; parameter gradients are summed
without a second batch-size normalization. The assignment and candidate
EMA state remain fixed across replay passes.

\subsection{Weighting multiple representation losses}
\label{app:loss-weighting}

The discrepancies have different scales across encoders. FD-Loss
\citep{yang2026fdloss} divides each encoder's FD by its detached current
value plus a constant, giving a weight
$[\operatorname{sg}\{\mathrm{FD}(R_e,G_e)\}+c]^{-1}$.
Here $R_e$ and $G_e$ denote real and generated feature statistics.
This weight changes with the generator. We instead use the discrepancy
between real training features $R_e$ and real validation features $V_e$
to set a fixed scale for each encoder.

For OT, we use $w_e=1/\mathrm{FD}(R_e,V_e)$, with the same encoder weight
for every component-count branch. In SigLIP/Inception/MAE order, the
denominators are $(0.62469,1.67956,0.04212)$.
Table~\ref{tab:app-fd-weighting} compares the published FD-Loss results
with our aligned single-Gaussian FD runs. The aligned recipe uses this
fixed normalization and the statistics schedule in
Appendix~\ref{app:statistics-ema}, without a GM branch.
It reduces FDr$^6$ on pMF-B.

\begin{table}[!htbp]
\centering\small
\caption{\textbf{Single-Gaussian FD with fixed reference normalization.}
$\text{FDr}^6\downarrow$ after 100 epochs with SIM encoders.
FD-Loss results are from \citet{yang2026fdloss};
the aligned runs use our statistics-update recipe.}
\label{tab:app-fd-weighting}
\begin{adjustbox}{Clip=0pt 0pt 0pt 0pt}
\begin{tabular}{@{}llc@{}}
\toprule
Training recipe & Encoder normalization & pMF-B \\
\midrule
FD-Loss & $\operatorname{sg}\{\mathrm{FD}(R,G)\}+c$ & 3.50 \\
\rowcolor{tableOursTint}
Aligned Gaussian FD & $\mathrm{FD}(R,V)$ & \textbf{3.17} \\
\bottomrule
\end{tabular}
\end{adjustbox}
\end{table}

For KL, we similarly use $w_e=1/\KL(R_e\Vert V_e)$, where $R_e$ and $V_e$
are single-Gaussian fits. These calibration distributions are distinct from the
training objective $\KL(Q\Vert P)$.
The resulting SigLIP/Inception/MAE weights are $(0.327,0.110,0.276)$ for
the ImageNet reference. Each weight is reused across the encoder's
Gaussian and mixture branches; we do not estimate a separate marginal
GM KL to normalize each branch.

For multiple branches, the final loss weight is $w_{e,K}=w_e\alpha_K$.
All multi-branch configurations use $\alpha_K=1$ for every included
branch. Text-to-image weights and their calibration split are given in
Appendix~\ref{app:t2i-comparison-settings}.

\subsection{Training configurations}
\label{app:configurations}
\subsubsection{ImageNet}
\label{app:imagenet-settings}

Table~\ref{tab:app-training-config} summarizes the common ImageNet
settings for the six JiT/pMF backbones. We use frozen
SigLIP2~\citep{tschannen2025siglip2}, Inception-v3~\citep{szegedy2016inception},
and MAE~\citep{he2022mae} encoders, abbreviated as SIM.
All six models use AdamW~\citep{loshchilov2019decoupled} and a global batch
size of 1,024, and are trained for 100 epochs. One epoch corresponds to 1,250 optimizer updates.
Learning rates are $10^{-5}$ for JiT and $10^{-6}$ for pMF, with five
warmup epochs followed by cosine decay.
We use no gradient clipping or dropout.

\begin{table}[!htbp]
\centering
\caption{\textbf{ImageNet post-training configurations.}
Slash-separated pMF sampling values follow B/L/H order.}
\label{tab:app-training-config}
\label{tab:app-sampling}
\begingroup
\small
\setlength{\tabcolsep}{6pt}
\renewcommand{\arraystretch}{1.12}
\begin{adjustbox}{Clip=0pt 0pt 0pt 0pt}
\begin{tabularx}{\linewidth}{@{}l*{2}{>{\centering\arraybackslash}X}@{}}
\toprule
Configuration & JiT~\citep{li2026jit} & pMF~\citep{lu2026pmf} \\
\midrule
\rowcolor{tableSectionTint}
\multicolumn{3}{@{}l}{\textbf{Model and representations}} \\
Model sizes & B, L, H & B, L, H \\
Image resolution & \multicolumn{2}{c}{$256\times256$} \\
Initialization & \multicolumn{2}{c}{Official pretrained weights} \\
Frozen encoders & \multicolumn{2}{c}{SigLIP2, Inception-v3, MAE} \\
\midrule
\rowcolor{tableSectionTint}
\multicolumn{3}{@{}l}{\textbf{Training}} \\
Training epochs & \multicolumn{2}{c}{100} \\
Global batch size & \multicolumn{2}{c}{1,024} \\
Optimizer & \multicolumn{2}{c}{AdamW, $(\beta_1,\beta_2)=(0.9,0.95)$} \\
Learning rate & $10^{-5}$ & $10^{-6}$ \\
Learning-rate schedule & \multicolumn{2}{c}{Cosine decay} \\
Warmup epochs & \multicolumn{2}{c}{5} \\
Weight decay & \multicolumn{2}{c}{0} \\
Precision & \multicolumn{2}{c}{BF16 mixed precision} \\
\midrule
\rowcolor{tableSectionTint}
\multicolumn{3}{@{}l}{\textbf{Statistics estimation}} \\
Estimator & \multicolumn{2}{c}{EMA} \\
Warm-start samples & \multicolumn{2}{c}{50,000 from the pretrained generator} \\
EMA decay & \multicolumn{2}{c}{$0.995\to0.999$, linear over epochs 8--32} \\
\midrule
\rowcolor{tableSectionTint}
\multicolumn{3}{@{}l}{\textbf{Sampling and evaluation}} \\
NFE & \multicolumn{2}{c}{1} \\
CFG & 1.0 & 8.5 / 7.0 / 7.0 \\
Noise scale & 1.0 & 1.0 / 1.0 / 2.0 \\
Evaluation samples & \multicolumn{2}{c}{50,000} \\
\bottomrule
\end{tabularx}
\end{adjustbox}
\endgroup
\end{table}

The guidance intervals~\citep{ho2022cfg,kynkaanniemi2024guidance} for pMF-B/L/H are $[0.1,0.7]$, $[0.2,0.7]$, and
$[0.2,0.6]$; JiT uses $[0.1,1.0]$.
Evaluation uses 50,000 one-step samples.
Each component count in a configuration has its own fitted reference
and generated statistics. Encoder and branch weights are specified in
Appendix~\ref{app:loss-weighting}.

\subsubsection{Text-to-image generation}
\label{app:t2i-comparison-settings}

Table~\ref{tab:text-to-image-comparison} uses two checkpoints initialized
from the official distilled FLUX.2 [klein] 4B model~\citep{blackforestlabs2026flux2klein}. All use
MGFlow-KL with equally weighted $K=1$ and $K=4$
branches and frozen SigLIP, MAE, and Inception image encoders.
The component-count choice is discussed in
Appendix~\ref{app:t2i-component-count}.
Following iRDM~\citep{feng2026rdm}, the joint variant concatenates image features with scaled, normalized
SigLIP2~\citep{tschannen2025siglip2} text features and retains the full
image--text cross-covariance.
For each caption $c$, we normalize its text embedding as
$\widehat t(c)=t(c)/\|t(c)\|_2$ and form the joint feature
$[f_e(x);\beta_e\widehat t(c)]$, without additional normalization of
the image feature $f_e(x)$.
We set $\beta_e=0.25\,m_e/m_t$, where $m_e$ and $m_t$ are the square roots
of the median pairwise squared distances of image and normalized text
features on the same 2,000 reference samples (seed 3407), excluding self-pairs.
In SigLIP/Inception/MAE order, $\beta_e\approx(3.76114,4.11487,0.51504)$.
These scales remain fixed during reference fitting and training and are
shared by the $K=1$ and $K=4$ branches.
The image-only variant omits text from the matching objective, not from
the generator conditioning.

\paragraph{Reference data.}
We reproduce the reference-data construction procedure of
iRDM~\citep{feng2026rdm} and AMFD~\citep{liu2026amfd}.
The teacher is the official distilled FLUX.2 [klein] 4B model, sampled
in four steps at $512\times512$ with guidance scale 1 and a 512-token
generator text context. For each of the 82,783 COCO~\citep{lin2014microsoft}
train2014 images, we select the first caption in annotation order and
retain the three highest-PickScore images from 24 teacher-generated
candidates, yielding 248,349 reconstructed images.
For each of the 553 GenEval prompts, we initially generate 150 candidates
and extend to at most 1,000 if fewer than 100 pass the official GenEval
correctness test. We retain the first 100 passing candidates in seed order,
or all passing candidates when fewer are available. Acceptance requires
the objects, counts, colors, and spatial relations specified by the prompt;
the detector, counting, and position thresholds are $0.3$, $0.9$, and $0.1$.
This yields 53,357 GenEval images. Both variants use the resulting
301,706-image reference set, sampled uniformly by image.

\paragraph{Encoder-weight calibration.}
Using seed 3407, we sample 100,000 reference rows without replacement and
split them into two disjoint 50,000-row subsets $R$ and $V$, shared across
encoders and variants. For each variant, we fit a single Gaussian to each
subset and set $w_e=1/\KL(R_e\Vert V_e)$, adding the encoder's $K=1$
score ridge to both covariance matrices. In SigLIP/Inception/MAE order,
the weights are $(0.08206932,0.04604280,0.19601964)$ for joint matching and
$(0.38121410,0.10227116,0.72123389)$ for image-only matching.
Each weight is reused for both $K=1$ and $K=4$, without further
normalization across encoders or branches.

\paragraph{Training and evaluation.}
Table~\ref{tab:app-t2i-config} lists the optimizer, statistics, and sampling
settings.
Evaluation uses one Euler step. GenEval~\citep{ghosh2023geneval} averages
its six category scores equally, using the same prompt set as the GenEval
reference subset. PickScore~\citep{kirstain2023pick} is evaluated on
499 Pick-a-Pic prompts. All evaluation protocols are aligned with AMFD~\citep{liu2026amfd}.

\begin{table}[!htbp]
\centering
\fontsize{9}{10}\selectfont
\caption{\textbf{Text-to-image post-training configurations.}
Both variants use reconstructed COCO and GenEval reference images.}
\label{tab:app-t2i-config}
\begingroup
\setlength{\tabcolsep}{4pt}
\renewcommand{\arraystretch}{1.12}
\begin{adjustbox}{Clip=0pt 0pt 0pt 0pt}
\begin{tabularx}{\linewidth}{@{}l*{2}{>{\centering\arraybackslash}X}@{}}
\toprule
Configuration & Joint & Image-only \\
\midrule
\rowcolor{tableSectionTint}
\multicolumn{3}{@{}l}{\textbf{Model and reference}} \\
Initialization & \multicolumn{2}{c}{Distilled FLUX.2 [klein] 4B} \\
Image resolution & \multicolumn{2}{c}{$512\times512$} \\
Image encoders & \multicolumn{2}{c}{SigLIP2, MAE, Inception-v3} \\
Text features & Frozen SigLIP2 & -- \\
Reference images & \multicolumn{2}{c}{301,706, including 53,357 GenEval images} \\
Distribution branches & \multicolumn{2}{c}{$K=1+4$} \\
Branch weights & \multicolumn{2}{c}{$1:1$} \\
\midrule
\rowcolor{tableSectionTint}
\multicolumn{3}{@{}l}{\textbf{Optimization and statistics}} \\
Training steps & \multicolumn{2}{c}{1,000} \\
Global batch size & \multicolumn{2}{c}{1,024} \\
Optimizer & \multicolumn{2}{c}{AdamW~\citep{loshchilov2019decoupled}} \\
Learning rate & \multicolumn{2}{c}{$5\times10^{-6}$} \\
Learning-rate schedule & \multicolumn{2}{c}{150-step warmup, then constant} \\
Statistics EMA decay & \multicolumn{2}{c}{0.99} \\
\midrule
\rowcolor{tableSectionTint}
\multicolumn{3}{@{}l}{\textbf{Sampling and evaluation}} \\
NFE / CFG & \multicolumn{2}{c}{1 / 1} \\
Text context length & \multicolumn{2}{c}{512 tokens} \\
GenEval samples & \multicolumn{2}{c}{2,212 (four per prompt)} \\
PickScore prompts & \multicolumn{2}{c}{499 Pick-a-Pic prompts} \\
\bottomrule
\end{tabularx}
\end{adjustbox}
\endgroup
\end{table}

\subsection{Statistics EMA schedule}
\label{app:statistics-ema}

An EMA averages out batch noise but also delays the response to a changing
generator. We use decay $0.995$ for the first eight ImageNet epochs,
increase it linearly to $0.999$ over epochs 8--32, and keep it fixed
thereafter. The smaller initial decay lets the component statistics
follow early distribution changes; the larger final decay smooths the
estimates later in training. This EMA updates feature statistics, not
generator parameters.

Table~\ref{tab:app-statistics-ema} compares the schedule with a constant
$0.999$ decay for JiT-B in the aligned single-Gaussian FD implementation.
After 50 epochs, the schedule reduces FDr$^6$ from $4.99$ to $4.87$.

\begin{table}[!htbp]
\centering\small
\caption{\textbf{Statistics EMA decay.}
JiT-B trained for 50 epochs with aligned Gaussian-FD, fixed
$\mathrm{FD}(R,V)$ normalization, and SIM encoders.}
\label{tab:app-statistics-ema}
\begin{adjustbox}{Clip=0pt 0pt 0pt 0pt}
\begin{tabular}{@{}lc@{}}
\toprule
Statistics decay & $\text{FDr}^6\downarrow$ \\
\midrule
Constant $0.999$ & 4.99 \\
\rowcolor{tableOursTint}
$0.995\to0.999$ & \textbf{4.87} \\
\bottomrule
\end{tabular}
\end{adjustbox}
\end{table}

\FloatBarrier
\section{Additional ImageNet Results}
\label{app:results}

Table~\ref{tab:imagenet256-additional-baselines} supplements the JiT/pMF
comparison in Table~\ref{tab:imagenet256-system-comparison} with other
discrete-, latent-, and pixel-space generators.
\begin{table}[!htbp]
  \centering
  \caption{\textbf{Additional ImageNet $256{\times}256$ baselines.}
  Published results from FD-Loss~\citep{yang2026fdloss},
  AdvFD~\citep{gao2026advfd}, and AMFD~\citep{liu2026amfd}.
  FDr$^6$ averages normalized Fr\'{e}chet distances over six encoders;
  FDr$^3$ excludes SigLIP, Inception, and MAE.
  $^\dagger$ denotes the full-CFG NFE upper bound for interval CFG;
  dashes denote unavailable results.
  JiT/pMF results are in
  Table~\ref{tab:imagenet256-system-comparison}.}
  \label{tab:imagenet256-additional-baselines}
  \begingroup
  \fontsize{7.4}{8.2}\selectfont
  \setlength{\tabcolsep}{3.7pt}
  \renewcommand{\arraystretch}{1.0}
  \begin{adjustbox}{Clip=0pt 0pt 0pt 0pt}
  \begin{tabularx}{\textwidth}{@{}Xccccccc@{}}
    \toprule
    Method & NFE & Space & \#Params
      & $\text{FDr}^6\!\downarrow$
      & $\text{FDr}^3\!\downarrow$
      & $\text{FID}\!\downarrow$ & $\mathrm{IS}\!\uparrow$ \\
    \midrule

    \rowcolor{tableSectionTint}
    \multicolumn{8}{@{}l}{\textit{Reference (real images)}} \\
    \textcolor{gray}{50k validation images}
      & \textcolor{gray}{N/A} & \textcolor{gray}{N/A}
      & \textcolor{gray}{N/A} & \textcolor{gray}{1.00}
      & \textcolor{gray}{1.00} & \textcolor{gray}{1.68} & \textcolor{gray}{232.2} \\

    \midrule
    \rowcolor{tableSectionTint}
    \multicolumn{8}{@{}l}{\textit{Discrete-space models}} \\
    VAR-d30~\citep{tian2024var}
      & $10{\times}2$ & discrete & 2B & 6.70 & 6.77 & 1.97 & 304.6 \\
    BAR-L~\citep{yu2026bar}
      & $256{\times}2{\times}4$ & discrete & 1.1B & 3.57 & 3.35 & 1.01 & 281.9 \\

    \midrule
    \rowcolor{tableSectionTint}
    \multicolumn{8}{@{}l}{\textit{Latent-space models, multi-step}} \\
    \multicolumn{8}{@{}l}{\quad\textit{without semantic distillation}} \\
    SiT-XL/2~\citep{ma2024sit}
      & $250{\times}2$ & latent & 675M & 8.44 & 9.20 & 2.12 & 256.7 \\
    MAR-L~\citep{li2024mar}
      & $256{\times}2{\times}100$ & latent & 478M & 6.68 & 7.32 & 1.80 & 293.4 \\
    FlowAR-H~\citep{ren2025flowar}
      & $50{\times}2^\dagger$ & latent & 1.9B & 6.13 & 6.16 & 1.68 & 274.1 \\
    MAR-H~\citep{li2024mar}
      & $256{\times}2{\times}100$ & latent & 942M & 5.61 & 6.31 & 1.56 & 299.5 \\
    MAR-L, DeTok~\citep{yang2026detok}
      & $256{\times}2{\times}100$ & latent & 478M & 5.49 & 6.05 & 1.39 & 306.2 \\
    \multicolumn{8}{@{}l}{\quad\textit{with semantic distillation}} \\
    REG~\citep{wu2025reg}
      & $250{\times}2^\dagger$ & latent & 685M & 4.64 & 5.15 & 1.54 & 302.9 \\
    SiT-XL/2-REPA~\citep{yu2025repa}
      & $250{\times}2^\dagger$ & latent & 675M & 5.45 & 6.05 & 1.42 & 306.1 \\
    LightningDiT~\citep{yao2025lightningdit}
      & $250{\times}2$ & latent & 675M & 4.57 & 5.02 & 1.42 & 294.3 \\
    DDT-XL~\citep{wang2026ddt}
      & $250{\times}2$ & latent & 675M & 5.70 & 6.38 & 1.26 & 309.3 \\
    REPA-E~\citep{leng2025repae}
      & $250{\times}2^\dagger$ & latent & 676M & 3.04 & 3.33 & 1.17 & 298.3 \\
    RAE-XL~\citep{zheng2026rae}
      & $50{\times}2^\dagger$ & latent & 839M & 3.26 & 3.92 & 1.16 & 261.0 \\

    \midrule
    \rowcolor{tableSectionTint}
    \multicolumn{8}{@{}l}{\textit{Latent-space models, one-step}} \\
    Drift-L (latent)~\citep{deng2026drifting}
      & 1 & latent & 463M & 10.92 & 11.32 & 1.53 & 257.2 \\
    iMF-XL~\citep{geng2026imf}
      & 1 & latent & 610M & 8.39 & 8.72 & 1.82 & 278.9 \\
    iMF-XL~\citep{geng2026imf}
      & 2 & latent & 610M & 7.48 & 7.79 & 1.61 & 289.1 \\

    \midrule
    \rowcolor{tableSectionTint}
    \multicolumn{8}{@{}l}{\textit{Pixel-space models}} \\
    PixNerd-XL~\citep{wang2026pixnerd}
      & $100{\times}2$ & pixel & 1.0B & 5.01 & -- & 2.10 & 318.8 \\
    Drift-L (pixel)~\citep{deng2026drifting}
      & 1 & pixel & 465M & 10.51 & 11.18 & 1.43 & 305.8 \\
    \bottomrule
  \end{tabularx}
  \end{adjustbox}
  \endgroup
\end{table}

\FloatBarrier

\section{Evaluation and Comparison Protocols}
\label{app:evaluation-protocol}

\subsection{The 10-epoch distribution-model comparison}
\label{app:short-comparison}

\paragraph{Shared setup.}
All six methods in Table~\ref{tab:distribution-model-comparison} start
from the same pretrained pMF-B and use frozen Inception features on
ImageNet $256\times256$. We train for 10 epochs with a global batch size
of 1,024 and a learning rate of $10^{-6}$. Sampling uses one step, CFG 8.5,
a guidance interval of $[0.1,0.7]$, and a noise scale of 1. Class labels are drawn
uniformly, and each feature objective matches the global class-marginal
distribution. We evaluate 50,000 generated images using
FDr$^6$ under the protocol in Appendix~\ref{app:metrics}.

The Gaussian and GM methods fit full-covariance reference distributions
to real features and keep them fixed during training. Generated-distribution
statistics are initialized from 50,000 samples of the pretrained generator
and updated by EMA of first and second moments. GM methods additionally
use the capacity-constrained LP in Eq.~\eqref{eq:method-batch-lp} to assign
each batch to components with the reference weights.

\paragraph{FD-Loss.}
The aligned FD-Loss~\citep{yang2026fdloss} baseline uses a single Gaussian
and minimizes its squared Wasserstein distance to the reference.
It retains the Gaussian reference, statistics estimator, and fixed
real-training/validation normalization of our OT implementation,
with no mixture term. Gradients pass through the current batch's
contribution to the EMA statistics, while historical statistics are detached.

\paragraph{Gaussian KL.}
This baseline uses $K=1$ and the Gaussian score difference
$v_{\mathrm{KL}}(z)=s_p(z)-s_q(z)$.
We train with detached-target regression in Eq.~\eqref{eq:mg-regression},
using $\eta=1$, and update the generated statistics after the generator step.

\paragraph{MGFlow-$W_2$.}
We use a single $K=4$ branch with LP assignments and minimize the
weighted sum of paired Gaussian costs in Eq.~\eqref{eq:method-mb-pair}.
The loss uses the same fixed normalization as the aligned FD-Loss baseline.
We differentiate through the current batch statistics, keeping the LP
assignments and historical statistics detached.

\paragraph{MGFlow-KL.}
We use a single $K=16$ branch. The LP assignments weight the paired
component score differences in Eq.~\eqref{eq:method-paired-score}.
Training uses the same detached-target regression and statistics-update
order as Gaussian KL. Neither GM method includes an additional Gaussian
branch or another component count.

\paragraph{W-Flow.}
W-Flow~\citep{han2026wflow} uses the released quadratic-cost Sinkhorn objective with
$\epsilon=0.05$, ten iterations, and its feature and force normalization.
Each update uses 1,024 gradient-carrying queries, an independent batch
of 1,024 generated support samples, and 1,024 real feature-bank samples.

\paragraph{Gaussian-kernel Drifting.}
Gaussian-kernel Drifting~\citep{deng2026drifting,turan2026secretly} uses the current generated batch as
detached negative support and 1,024 real feature-bank samples.
Its fixed bandwidths satisfy $2\sigma^2\in\{0.02,0.05,0.2\}$ in normalized
feature coordinates, with the self-interaction diagonal masked and
per-scale force normalization. Both sample-based methods gather the
complete global batch before computing their fields. Their sample
interactions replace the moment estimator, so they require no generated
statistics initialization or EMA.

\subsection{Evaluation in training and held-out representations}
\label{app:metrics}

\paragraph{Representation encoders.}
Table~\ref{tab:evaluation-encoders} lists the six frozen encoders used for
ImageNet evaluation. SIM uses SigLIP2, Inception-v3, and MAE for training;
ConvNeXt-v2, DINOv2, and CLIP are held out. We use the pooled or CLS
features without the classification or contrastive projection head.
The five \texttt{timm} encoders use bicubic resizing and their pretrained
input normalization; Inception-v3 uses TensorFlow-compatible FID preprocessing.

\begin{table}[ht]
  \centering
  \caption{\textbf{ImageNet representation encoders.}
  Input denotes the resized image side length before patch padding.
  The first three encoders form SIM; the last three define
  FDr$^3$. The fixed denominator $b_e$ is used to compute
  $\text{FDr}_e=F_e/b_e$.}
  \label{tab:evaluation-encoders}
  \small
  \renewcommand{\arraystretch}{1.1}
  \begin{tabular*}{\linewidth}{@{}l@{\extracolsep{\fill}}lrrlr@{}}
    \toprule
    Encoder & Variant & Input & Dimension & Feature & $b_e$ \\
    \midrule
    SigLIP2~\citep{tschannen2025siglip2}
      & ViT-SO400M/16 & 224 & 1,152 & Attention pool & 0.60 \\
    Inception-v3~\citep{szegedy2016inception}
      & FID weights & 299 & 2,048 & Global average pool & 1.68 \\
    MAE~\citep{he2022mae}
      & ViT-L/16 & 224 & 1,024 & CLS token & 0.04 \\
    \midrule
    ConvNeXt-v2~\citep{woo2023convnextv2}
      & Base, IN-22K $\to$ IN-1K & 224 & 1,024 & Global average pool & 56.87 \\
    DINOv2~\citep{oquab2023dinov2}
      & ViT-L/14 & 256 & 1,024 & CLS token & 14.19 \\
    CLIP~\citep{radford2021clip}
      & ViT-L/14, OpenAI & 256 & 1,024 & CLS token & 5.60 \\
    \bottomrule
  \end{tabular*}
\end{table}

\paragraph{Metric aggregation.}
We evaluate 50,000 generated images using FDr$^6$
\citep{yang2026fdloss}. For an evaluation encoder
$e$, let $F_e$ be the Fr\'{e}chet distance between generated and real
features, and let $b_e$ be the corresponding real-validation reference
value. For Inception features, the unnormalized distance $F_e$ is FID
(FD-Inception)~\citep{heusel2017fid}. The normalized score is $\text{FDr}_e=F_e/b_e$. We use the
arithmetic mean across the six evaluation spaces,
\begin{equation}
  \text{FDr}^6=\frac{1}{6}\sum_{e\in\mathcal H_6}\text{FDr}_e,
  \qquad
  \mathcal H_6=\{\mathrm{Inception},\mathrm{ConvNeXt},\mathrm{DINOv2},
  \mathrm{MAE},\mathrm{SigLIP},\mathrm{CLIP}\}.
\end{equation}
For a single training encoder $e_{\rm train}$, we additionally report
\begin{equation}
  \text{FDr}^5
  =\frac{1}{5}\sum_{e\in\mathcal H_6\setminus\{e_{\rm train}\}}
    \text{FDr}_e.
\end{equation}
For SIM training, the corresponding held-out score is
\begin{equation}
  \text{FDr}^3=\frac{1}{3}
  \left(\text{FDr}_{\rm ConvNeXt}
        +\text{FDr}_{\rm DINOv2}
        +\text{FDr}_{\rm CLIP}\right).
\end{equation}
IS~\citep{salimans2016improved} is the exponentiated mean KL divergence
between each generated image's Inception-v3 class probabilities and their
marginal over generated images.

\section{Reference GMs and Component Structure}
\label{app:references}
\label{app:semantics}

\begingroup
\setlength{\intextsep}{4pt plus 1pt minus 1pt}

Reference fitting is described in Algorithm~\ref{alg:reference-fit}.
This section examines the structure of the fitted components.

Figure~\ref{fig:app-reference-projections} visualizes the ImageNet training
references with $K=1,4,16$. The mixtures capture distinct feature groups that a single Gaussian
represents only through global moments, while several components still overlap in the
two-dimensional view.

\begin{figure}[!htbp]
\centering
\vspace{2mm}
\captionsetup{skip=4pt}
\includegraphics[width=0.90\linewidth,height=0.95\linewidth,keepaspectratio]{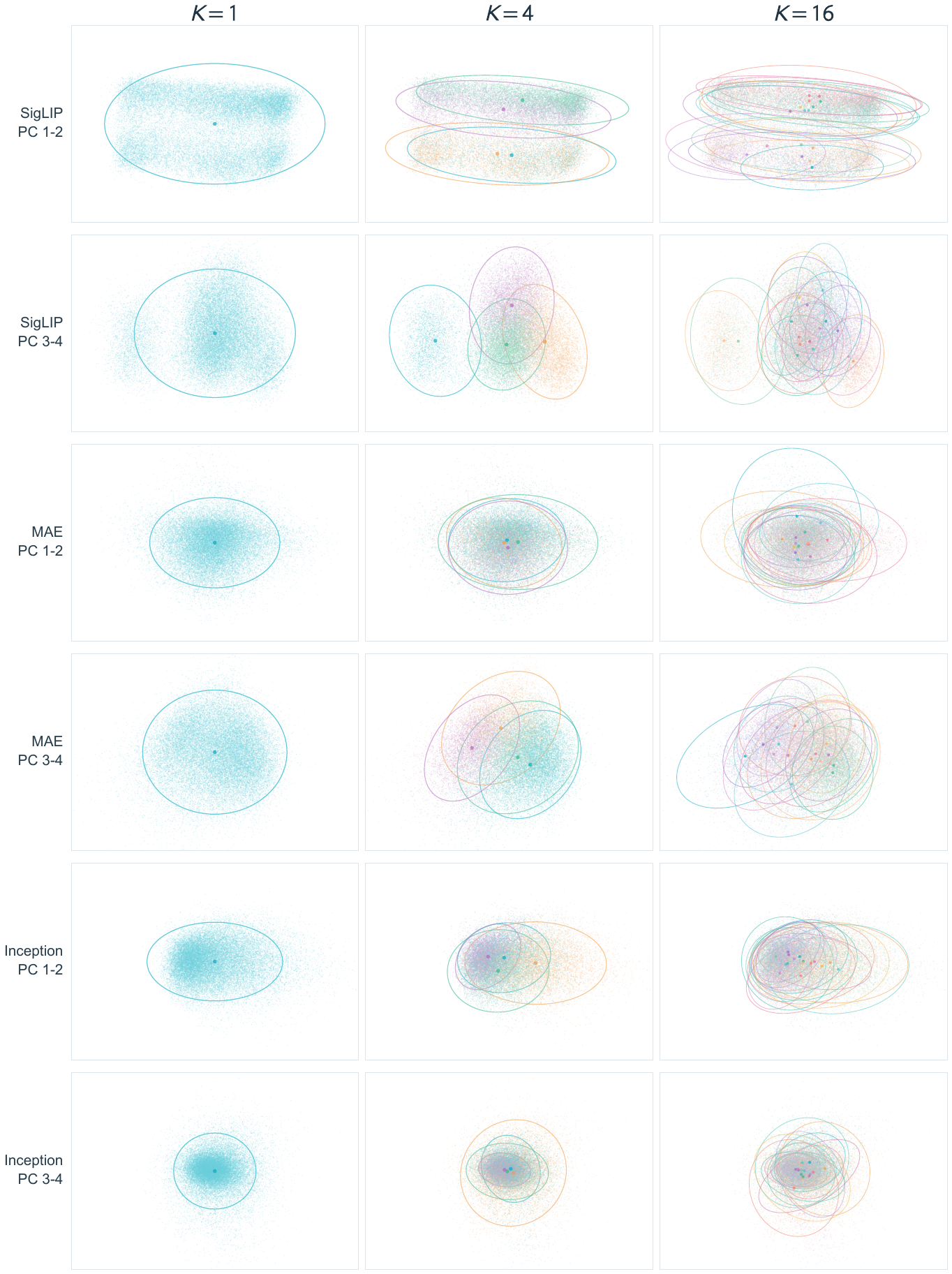}
\caption{\textbf{Geometry of the ImageNet reference distributions.}
Within each encoder, the two rows show PC1--2 and PC3--4 of the global
feature covariance. All three models share the PCA projection and axis
limits within each row.
Colors show maximum-posterior assignments in the original feature space
and are local to each mixture. Ellipses are 90\% probability contours of
the projected Gaussian components; the larger dots mark their means.}
\label{fig:app-reference-projections}
\vspace{1mm}
\end{figure}

We measure class--component association on 200 randomly sampled training
images per ImageNet class, for 200,000 images in total. For class $c$, let
\begin{equation}
  r_{ck}=\frac{1}{N_c}\sum_{n:y_n\in c}\gamma_k(y_n),\qquad
  q_c=\max_k r_{ck}.
\end{equation}
Here $q_c$ is the average mass assigned to the dominant component of
class $c$.

\begin{table}[!htb]
\centering\small
\setlength{\belowcaptionskip}{4pt}
\caption{\textbf{Class concentration versus image-level assignment.}
We use 200 images per ImageNet class. Class concentration $q_c$ is the
mean responsibility of class $c$ for its dominant component;
``Image peak'' averages the maximum responsibility of each image.}
\label{tab:app-class-concentration}
\begin{tabular}{@{}lrrrr@{}}
\toprule
Encoder & $K$ & Mean $q_c$ & Classes with $q_c\geq0.9$ & Image peak \\
\midrule
MAE & 4 & 81.24\% & 398 & 99.88\% \\
SigLIP & 4 & 91.12\% & 744 & 99.96\% \\
Inception & 4 & 82.23\% & 461 & 99.97\% \\
\midrule
MAE & 16 & 64.79\% & 113 & 99.79\% \\
SigLIP & 16 & 71.88\% & 229 & 99.94\% \\
Inception & 16 & 77.21\% & 305 & 99.96\% \\
\bottomrule
\end{tabular}
\end{table}

\begin{figure}[!htb]
\centering
\captionsetup{skip=4pt}
\includegraphics[width=0.90\linewidth]{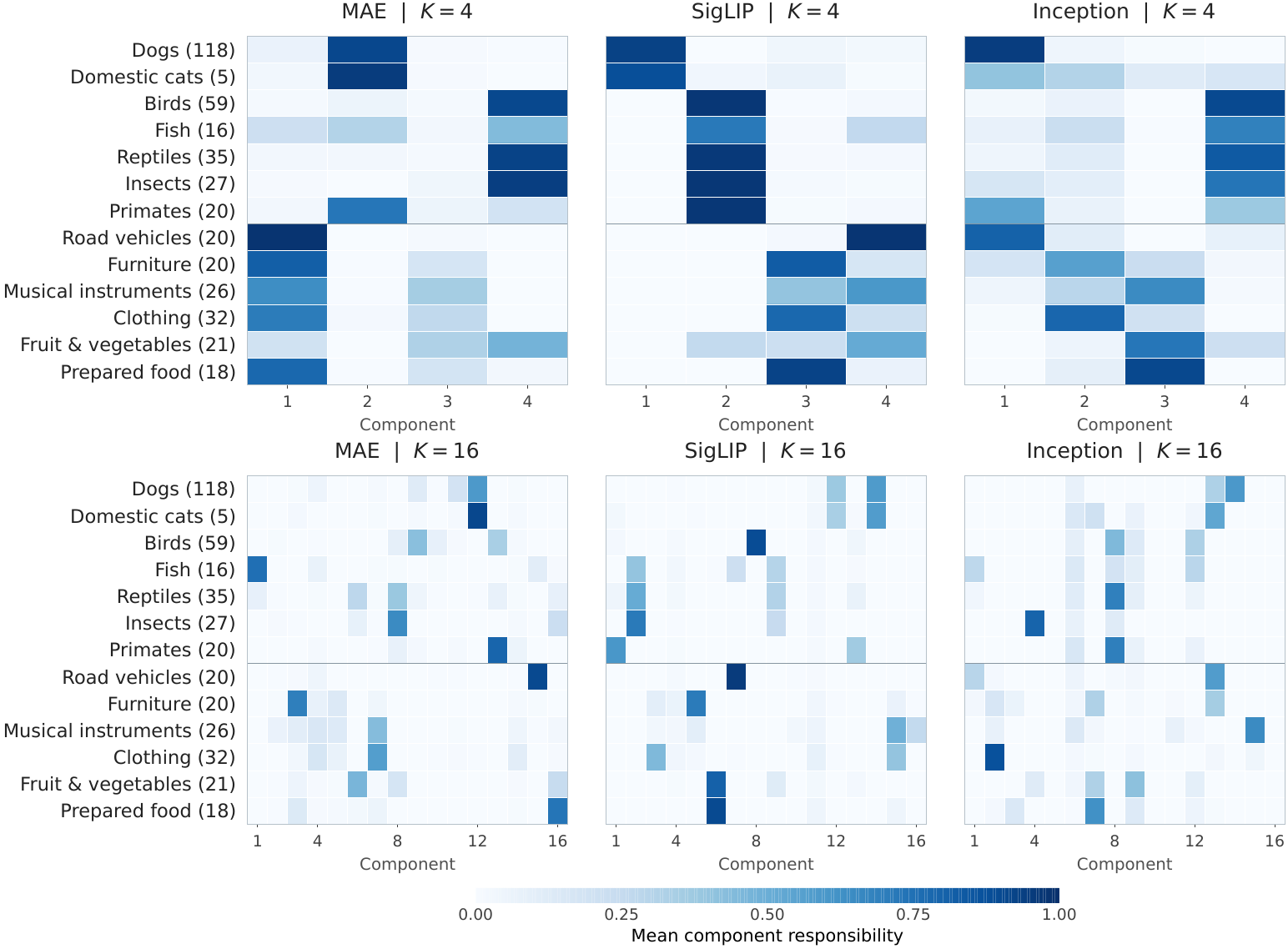}
\caption{\textbf{Semantic groups across independently fitted mixtures.}
Rows show mean posterior responsibilities for 13 label-defined groups
covering 417 ImageNet classes, using 200 images per class.
We average $r_{ck}$ equally over the classes in each group;
parentheses give the number of classes.
Component indices are local to each GM, not aligned across panels.}
\label{fig:app-semantic-groups}
\end{figure}

In Table~\ref{tab:app-class-concentration}, the mean image-level peak
remains above $99\%$, while class concentration decreases at $K=16$.
Individual images have sharp assignments, but images within the same
class are distributed across several components.

Figure~\ref{fig:app-semantic-groups} shows how the grouping of
semantic categories varies with $K$ and the encoder. For SigLIP, furniture and prepared food share a dominant
component at $K=4$ but peak in distinct components at $K=16$.
The grouping also depends on the encoder: at $K=4$, Inception places
dogs and road vehicles in the same dominant component, whereas MAE and
SigLIP separate them.

\endgroup
\FloatBarrier

\section{Score Ridge Settings}
\label{app:ridge}

We vary the score ridge $\lambda$ with $K=1$, training
pMF-B for 10 epochs with MAE, SigLIP, or Inception alone.
Table~\ref{tab:app-ridge-k1} reports the sweep results. These sweeps initialize the generated statistics with 5,000 samples,
retain a 5,000-sample prior, and use initial gradient-norm calibration.
The runs in Table~\ref{tab:distribution-model-comparison} instead use
50,000 initialization samples, no prior, and fixed loss scaling.

The lowest FDr$^6$ is obtained at $\lambda=0.001$ for MAE
and $\lambda=0.03$ for both SigLIP and Inception, reaching
$5.769$, $5.742$, and $10.720$, respectively.
A larger ridge value does not consistently improve the score.

\begin{table}[!ht]
\centering\small
\setlength{\tabcolsep}{7pt}
\renewcommand{\arraystretch}{1.05}
\caption{\textbf{Single-encoder ridge selection.}
pMF-B trained with MGFlow-KL, $K=1$, for 10 epochs.
All metrics are lower-is-better. Bold marks the lowest FDr$^6$
for each training encoder.}
\label{tab:app-ridge-k1}
\begin{adjustbox}{Clip=0pt 0pt 0pt 0pt}
\begin{tabular}{@{}lrrrrr@{}}
\toprule
Encoder & $\lambda$ & FDr$^6$ & FID & $\mathrm{FD}_{\mathrm{MAE}}$ & $\mathrm{FD}_{\mathrm{SigLIP}}$ \\
\midrule
MAE & $10^{-5}$ & 5.867 & 2.559 & 0.091 & 8.470 \\
 & $10^{-4}$ & 5.821 & 2.749 & 0.089 & 8.326 \\
 & $3\times10^{-4}$ & 5.800 & 2.867 & 0.086 & 8.260 \\
 & $5\times10^{-4}$ & 5.837 & 2.966 & 0.077 & 8.353 \\
\rowcolor{tableOursTint}
 & $10^{-3}$ & \textbf{5.769} & 3.207 & 0.076 & 8.107 \\
 & $2\times10^{-3}$ & 5.881 & 3.504 & 0.072 & 8.274 \\
 & $3\times10^{-3}$ & 5.902 & 3.876 & 0.067 & 8.380 \\
 & $10^{-2}$ & 6.312 & 5.288 & 0.064 & 8.780 \\
\midrule
SigLIP & $10^{-5}$ & 6.497 & 6.116 & 0.525 & 3.918 \\
 & $10^{-4}$ & 6.110 & 6.535 & 0.414 & 3.834 \\
 & $10^{-3}$ & 6.010 & 6.487 & 0.410 & 3.730 \\
 & $3\times10^{-3}$ & 6.103 & 7.311 & 0.416 & 3.707 \\
 & $5\times10^{-3}$ & 5.975 & 7.215 & 0.400 & 3.588 \\
 & $10^{-2}$ & 5.970 & 7.000 & 0.408 & 3.576 \\
\rowcolor{tableOursTint}
 & $3\times10^{-2}$ & \textbf{5.742} & 7.338 & 0.366 & 3.380 \\
 & $5\times10^{-2}$ & 5.831 & 7.605 & 0.368 & 3.406 \\
\midrule
Inception & $10^{-5}$ & 11.109 & 1.153 & 0.353 & 16.464 \\
 & $10^{-4}$ & 10.966 & 1.072 & 0.366 & 16.070 \\
 & $3\times10^{-4}$ & 10.957 & 1.087 & 0.355 & 16.118 \\
 & $10^{-3}$ & 11.064 & 1.139 & 0.351 & 16.496 \\
 & $2\times10^{-3}$ & 10.874 & 1.087 & 0.351 & 16.047 \\
 & $3\times10^{-3}$ & 10.828 & 1.096 & 0.352 & 15.793 \\
 & $5\times10^{-3}$ & 10.995 & 1.105 & 0.360 & 16.211 \\
 & $10^{-2}$ & 10.898 & 1.057 & 0.359 & 15.727 \\
\rowcolor{tableOursTint}
 & $3\times10^{-2}$ & \textbf{10.720} & 1.104 & 0.367 & 15.528 \\
 & $5\times10^{-2}$ & 10.836 & 1.105 & 0.354 & 15.731 \\
\bottomrule
\end{tabular}
\end{adjustbox}
\end{table}

\FloatBarrier

\paragraph{Ridge configuration.}
For ImageNet, the $K=1$ branches use $\lambda_1=(0.03,0.03,0.001)$
in SigLIP/\allowbreak Inception/\allowbreak MAE order.
Table~\ref{tab:app-ridge-imagenet} reports the reference covariance
eigenvalue quantiles and the percentile of each $\lambda_1$.

\begin{table}[!ht]
\centering\small
\setlength{\tabcolsep}{4pt}
\renewcommand{\arraystretch}{1.12}
\caption{\textbf{ImageNet covariance eigenvalue quantiles and score ridges.}
Quantiles of the $K=1$ reference spectra use the inverse empirical CDF.
Percentile is the fraction of eigenvalues at or below the training ridge
$\lambda_1$.}
\label{tab:app-ridge-imagenet}
\begin{adjustbox}{Clip=0pt 0pt 0pt 0pt}
\begin{tabular*}{\linewidth}{@{\extracolsep{\fill}}lrrrrr@{}}
\toprule
Encoder & 25th & 50th & 75th & Percentile & $\lambda_1$ \\
\midrule
SigLIP & $1.02\times10^{-3}$ & $1.99\times10^{-2}$ & $1.04\times10^{-1}$ & $56.34\%$ & $0.03$ \\
Inception & $6.78\times10^{-3}$ & $1.58\times10^{-2}$ & $7.54\times10^{-2}$ & $62.26\%$ & $0.03$ \\
MAE & $1.11\times10^{-4}$ & $3.18\times10^{-4}$ & $1.12\times10^{-3}$ & $73.24\%$ & $0.001$ \\
\bottomrule
\end{tabular*}
\end{adjustbox}
\end{table}

For text-to-image generation, we select the eigenvalues at these same
percentiles in the T2I $K=1$ reference spectra and use nearby rounded
values for $\lambda_1$ (Table~\ref{tab:app-ridge-t2i}).
Joint matching uses the full image--text covariance; image-only matching
uses the image covariance. We use $\lambda_4=3\lambda_1$ in both tasks
and $\lambda_{16}=9\lambda_1$ for ImageNet, with the same ridge added
to the reference and generated covariances.

\begin{table}[!ht]
\centering\small
\setlength{\tabcolsep}{4pt}
\renewcommand{\arraystretch}{1.12}
\caption{\textbf{T2I covariance eigenvalue quantiles and score ridges.}
The $K=1$ references use 301,706 COCO and GenEval images.
Matched denotes the eigenvalue at the encoder-specific ImageNet percentile;
$\lambda_1$ is the value used in training.}
\label{tab:app-ridge-t2i}
\begin{adjustbox}{Clip=0pt 0pt 0pt 0pt}
\begin{tabular*}{\linewidth}{@{\extracolsep{\fill}}lrrrrr@{}}
\toprule
Encoder & 25th & 50th & 75th & Matched & $\lambda_1$ \\
\midrule
\rowcolor{tableSectionTint}
\multicolumn{6}{@{}l}{\textit{Joint image--text}} \\
SigLIP & $1.10\times10^{-4}$ & $1.90\times10^{-3}$ & $1.71\times10^{-2}$ & $3.22\times10^{-3}$ & $0.003$ \\
Inception & $2.16\times10^{-3}$ & $5.60\times10^{-3}$ & $2.10\times10^{-2}$ & $9.44\times10^{-3}$ & $0.01$ \\
MAE & $1.06\times10^{-5}$ & $5.36\times10^{-5}$ & $2.14\times10^{-4}$ & $1.91\times10^{-4}$ & $0.0002$ \\
\midrule
\rowcolor{tableSectionTint}
\multicolumn{6}{@{}l}{\textit{Image-only}} \\
SigLIP & $6.90\times10^{-4}$ & $1.33\times10^{-2}$ & $6.88\times10^{-2}$ & $1.97\times10^{-2}$ & $0.02$ \\
Inception & $4.44\times10^{-3}$ & $1.00\times10^{-2}$ & $4.30\times10^{-2}$ & $1.87\times10^{-2}$ & $0.02$ \\
MAE & $5.35\times10^{-5}$ & $1.51\times10^{-4}$ & $5.42\times10^{-4}$ & $4.83\times10^{-4}$ & $0.0005$ \\
\bottomrule
\end{tabular*}
\end{adjustbox}
\end{table}

\FloatBarrier

\section{Component Matching}
\label{app:matching}

\paragraph{Toy-experiment settings.}
\label{app:toy-settings}
Figure~\ref{fig:method-toy-pairing} uses an equal-weight reference mixture
with means $(-3.5,0)$ and $(3.5,0)$ and covariance $0.7^2I_2$ for both
components. All variants start from the same 2,048 particles sampled from
$\mathcal{N}((-0.5,0)^{\mathsf T},0.45^2I_2)$ and model $Q$
with two full-covariance Gaussian components. Initially, the first component is fitted to all particles, while the
second has no assigned particles.
Posterior assigns particles using $Q$'s posterior responsibilities.
LP-global and LP-paired both use capacity-constrained LP assignments
to the fixed reference components, with equal mass allocated to each component.
Posterior and LP-global use the global KL field; LP-paired uses the paired
component field. Component masses and first and second moments are
recomputed from the pre-update particles at each iteration, without EMA
smoothing, and used to construct the next iteration's field.
Particles follow explicit Euler updates with step size $0.01$; the figure
shows steps 0, 20, 80, and 200, corresponding to $t=0,0.2,0.8,2$.
Ellipses show the fitted components, and the final percentages report the fractions of particles to the left
and right of $x=0$, respectively.

\paragraph{Assignment-ablation settings.}
\label{app:assignment-ablation-settings}
All configurations in Table~\ref{tab:method-lp} use $K=4$. We train for 10 epochs and evaluate with 50k images.
Training time is measured on eight H200 GPUs using
the protocol in Appendix~\ref{app:timing-protocol}.

The Posterior variant uses the generated mixture's posterior
responsibilities to update component statistics. LP-global instead
uses the capacity-constrained assignment in
Eq.~\eqref{eq:method-batch-lp}, while retaining the global KL field
or full component transport for $W_2$. LP-paired uses the same LP
assignment but replaces these updates with paired component scores
or diagonal component transport, respectively.

\Needspace{10\baselineskip}
\section{Computation and Training Efficiency}
\label{app:efficiency}

\subsection{Training-time comparison and protocol}
\label{app:timing-protocol}

Table~\ref{tab:app-training-speed} compares training times for the six
JiT/pMF backbones. MGFlow-KL runs faster per training step than the official
FD-Loss implementation on all L/H backbones, but slightly slower on the
two B backbones under the benchmark settings below.

\begin{table}[htbp]
  \centering\small
  \caption{\textbf{Training time per step.}
  Seconds per optimizer update on eight H200 GPUs with a global batch of 1,024.
  MGFlow-$W_2$ uses $K=1+4$; MGFlow-KL uses $K=1+4+16$.
  Training states and precision settings are specified below. Lower is better.}
  \label{tab:app-training-speed}
  \begin{adjustbox}{Clip=0pt 0pt 0pt 0pt}
  \begin{tabular*}{0.85\linewidth}{@{\extracolsep{\fill}}lccc@{}}
    \toprule
    Backbone & FD-Loss~\citep{yang2026fdloss} & MGFlow-$W_2$ & MGFlow-KL \\
    \midrule
    JiT-B & \textbf{0.799} & 0.969 & 0.846 \\
    JiT-L & 1.131 & 1.137 & \textbf{1.009} \\
    JiT-H & 2.192 & 1.368 & \textbf{1.247} \\
    \midrule
    pMF-B & \textbf{0.829} & 0.874 & 0.882 \\
    pMF-L & 1.660 & \textbf{1.103} & 1.112 \\
    pMF-H & 2.572 & \textbf{1.703} & 1.882 \\
    \bottomrule
  \end{tabular*}
  \end{adjustbox}
\end{table}

The training times in Tables~\ref{tab:method-lp}
and~\ref{tab:app-training-speed} are measured on the same node
with eight H200 GPUs and a global batch of 1,024. We use 128 samples per
GPU unless memory requires gradient accumulation, as listed in
Table~\ref{tab:app-timing-accum}. All assignment/objective ablations use
$a=1$.

We synchronize CUDA and record the slowest rank for each complete optimizer
step, including generation, all training encoders, the objective, backward
passes, communication, the optimizer, and native metric reduction.
Compilation, statistics initialization, checkpointing, evaluation, and
benchmark-log I/O are excluded. After at least ten warmup steps, we report
the first twenty-step mean with both coefficient of variation and relative
drift between its two halves at most $5\%$. GPUs are not shared with other jobs.

\paragraph{Training state and precision.}
The assignment/objective ablations, MGFlow-$W_2$ ($K=1+4$), and FD-Loss
are timed from their pretrained-base initial training states. The
MGFlow-KL ($K=1+4+16$) column instead reports plateau measurements from
60-epoch checkpoints, restoring the model, optimizer, and all
nine encoder--branch statistics queues. MGFlow uses BF16 neural forwards
with its original covariance precision. FD-Loss uses the official
implementation\footnote{\url{https://github.com/Jiawei-Yang/FD-Loss},
commit \texttt{5c03b8112fec}.}, retaining FP32 training, enabled TF32,
and the original FD-statistics precision, without added autocast.
These are end-to-end implementation timings, with the training states
and precision settings specified above.

\begin{table}[htbp]
\centering\small
\caption{\textbf{Gradient accumulation in the timing benchmark.}
Entries are $a$; the per-GPU microbatch is $128/a$ and the global batch
remains 1,024. Values above one follow a recorded CUDA OOM at the preceding
larger microbatch.}
\label{tab:app-timing-accum}
\begin{tabular*}{\linewidth}{@{\extracolsep{\fill}}lccc@{}}
\toprule
Backbone & MGFlow-$W_2$ & MGFlow-KL & Official FD-Loss \\
\midrule
JiT-B & 1 & 1 & 1 \\
JiT-L & 1 & 1 & 1 \\
JiT-H & 1 & 1 & 2 \\
pMF-B & 1 & 1 & 1 \\
pMF-L & 1 & 1 & 2 \\
pMF-H & 2 & 2 & 4 \\
\bottomrule
\end{tabular*}
\end{table}

\paragraph{Full-batch FD-Loss with accumulation.}
At $a=1$, we use the upstream training step. For $a>1$, we evaluate one
FD objective over the full global batch, then replay microbatches to
backpropagate its feature gradients, as in Appendix~\ref{app:training-loop}.
Each global batch performs one optimizer update and one statistics update;
replay overhead is included in the measured time.

\paragraph{Why use plateau KL measurements?}
LP solve time can change during training. On JiT-L, step time drops rapidly from about $1.8$ to $1.0$ seconds
early in training as the CPU LP solve becomes faster.
At the 60-epoch checkpoint used for timing, the CPU LP solve takes
$0.241$ seconds, compared with $1.031$ seconds at initialization.
The corresponding pMF-B/L times change little: $0.890/1.115$ seconds
initially and $0.882/1.112$ seconds after 60 epochs.

\subsection{Scaling the number of components}
\label{app:component-scaling}

\paragraph{Computational cost.}
Consider one encoder with feature dimension $d$, batch size $B$, and
$K$ full-covariance components. Both variants form the Gaussian assignment
costs and update component moments in $O(BKd^2)$ time.
The component moment state requires $O(Kd^2)$ storage per encoder.
The sample LP has
$BK$ variables and $B+K-1$ independent equality constraints; denote its
solve time by $T_{\mathrm{LP}}(B,K)$.

MGFlow-$W_2$ evaluates $K$ paired Gaussian costs. Each requires a dense
spectral decomposition for the matrix-square-root trace, giving
$O(Kd^3)$ work. Evaluating all component pairs instead
would require $O(K^2d^3)$ work before solving a $K\times K$ component
transport LP~\citep{delon2020gmmwasserstein}. Fixed pairing removes this
component-level LP, but retains the sample-assignment LP.
MGFlow-KL uses Cholesky factorizations in $O(Kd^3)$ time and evaluates
the Gaussian scores in $O(BKd^2)$ time. Global and paired KL scores have
the same leading dense cost: pairing reuses $R^*$ instead of computing
two mixture posteriors, but still evaluates $K$ component scores per
feature. Excluding the shared
neural-network computation, both paired variants therefore have cost
\begin{equation}
  O(BKd^2+Kd^3)+T_{\mathrm{LP}}(B,K).
\end{equation}
For joint resolutions, these costs sum over
$K\in\mathcal K$, while generated images and encoder features are shared.

\paragraph{Why KL is cheaper in practice.}
The cubic terms involve different matrix operations. For KL, we factor
the covariance used in score evaluation as $C_{q,k}=L_{q,k}L_{q,k}^{\top}$ and compute each score
by two triangular solves:
\begin{equation}
  L_{q,k}u=\mu_{q,k}-z,\qquad
  L_{q,k}^{\top}s_{q,k}(z)=u.
\end{equation}
Each generated-component factor is reused across the batch, and reference
factors are cached. Cholesky factorization has a smaller computational
constant than dense spectral decomposition; the subsequent solves need
neither eigenvectors nor an explicit covariance inverse.

For $W_2$, even with the reference square root cached, each update forms
$C_k=\Sigma_{p,k}^{1/2}\Sigma_{q,k}\Sigma_{p,k}^{1/2}$ through two dense
matrix multiplications and computes $\operatorname{tr}(C_k^{1/2})$ from
its eigenvalues. Gradients then pass through this spectral term, the
matrix products, and the current batch's contribution to the covariance.
KL instead computes the entire field with gradients stopped. Its backward
pass differentiates only the squared regression loss with respect to the
features, followed by the shared encoder and generator backward passes.
KL therefore saves the spectral computation, its backward pass, and the
memory needed to differentiate the current-batch covariance estimate.
The practical advantage lies in these cheaper matrix operations and the
shorter backward path, rather than a lower asymptotic order.

\paragraph{Larger Wasserstein objectives.}
We extend the timing protocol in Appendix~\ref{app:timing-protocol} to
MGFlow-$W_2$ with $K=16$ and $K=1+4+16$ on the same eight-H200 node,
using SIM features and a global batch size of 1,024. Runs start from pretrained
JiT-H and pMF-H weights. JiT-H uses a per-GPU batch of 128 without
accumulation; pMF-H uses 64 with two accumulation steps after an OOM at 128,
matching the accepted geometry in Table~\ref{tab:app-timing-accum}.
Table~\ref{tab:app-large-k-speed} compares these measurements with
Table~\ref{tab:app-training-speed}. Relative to $K=1+4$, the two larger
Wasserstein objectives take $1.68\times/1.95\times$ as long per step on
JiT-H and $1.26\times/1.40\times$ on pMF-H. This cost motivates our default
$K=1+4$ for MGFlow-$W_2$. MGFlow-KL uses $K=1+4+16$ at 1.247 and 1.882
seconds per step on the two backbones in the plateau benchmark above.

\begin{table}[!t]
\centering
\fontsize{9.5}{11}\selectfont
\setlength{\tabcolsep}{2pt}
\captionsetup{font={footnotesize,stretch=1.0},skip=4pt}
\begin{minipage}[t]{0.42\linewidth}
\vspace{0pt}
\caption{\textbf{Training cost of larger mixtures for MGFlow-$W_2$.}
Seconds per optimizer update on eight H200 GPUs with SIM features and
global batch size 1,024. All runs start from pretrained weights.
The $K=1+4$ result is from Table~\ref{tab:app-training-speed}.}
\label{tab:app-large-k-speed}
\begin{tabular*}{\linewidth}{@{\extracolsep{\fill}}lccc@{}}
\toprule
Backbone & $K=1+4$ & $K=16$ & $K=1+4+16$ \\
\midrule
JiT-H & 1.368 & 2.294 & 2.661 \\
pMF-H & 1.703 & 2.146 & 2.384 \\
\bottomrule
\end{tabular*}
\end{minipage}\hfill
\begin{minipage}[t]{0.55\linewidth}
\vspace{0pt}
\caption{\textbf{Component balance and sample support at $K=64$.}
Ranges across seeds 3407--3410. $N\pi_{\min}$ is the smallest component's
reference responsibility mass; $N_0\pi_{\min}$ and $B\pi_{\min}$ are its
LP allocation budgets for 50,000-sample initialization and a batch of
1,024. Passed denotes the number of fits with $\pi_{\max}/\pi_{\min}\leq10$.}
\label{tab:app-k64-support}
\begin{tabular*}{\linewidth}{@{\extracolsep{\fill}}lccc@{}}
\toprule
Encoder & Inception & MAE & SigLIP \\
\midrule
$\pi_{\max}/\pi_{\min}$ & 50.58--75.54 & 15.44--16.66 & 3.02--5.64 \\
$N\pi_{\min}$ & 978--1,358 & 2,766--3,112 & 7,767--10,146 \\
$N_0\pi_{\min}$ & 38.2--53.0 & 107.9--121.5 & 303.1--396.0 \\
$B\pi_{\min}$ & 0.78--1.09 & 2.21--2.49 & 6.21--8.11 \\
Passed & 0/4 & 0/4 & 4/4 \\
\bottomrule
\end{tabular*}
\end{minipage}
\end{table}

\paragraph{Sample support for component statistics.}
Increasing $K$ also divides the statistics budget among more components.
Under the LP constraints, component $k$ receives total assignment mass
$B\pi_k$ per batch and $N_0\pi_k$ over the $N_0$ initialization samples.
We fit $K=64$ full-covariance GMs to $N=1{,}281{,}167$ real features for
each encoder with seeds 3407--3410. All twelve EM runs reach the
parameter-change stopping criterion. No Inception or MAE fit
satisfies $\pi_{\max}/\pi_{\min}\leq10$, whereas all four SigLIP fits
do (Table~\ref{tab:app-k64-support}).

Inception is the most restrictive: its smallest component has only
978--1,358 units of reference responsibility mass in $d=2{,}048$
dimensions. With $N_0=50{,}000$ and $B=1{,}024$, the corresponding LP
budgets are only 38--53 at initialization and 0.78--1.09 per batch.
These small components provide weak support for estimating a full
covariance. EMA accumulates information over time, but extending the
averaging window also slows adaptation to the changing generator.
This sample-support limitation motivates retaining $K=16$ as the largest
branch in MGFlow-KL rather than using $K=64$ across the three encoders.

\FloatBarrier
\paragraph{Component count for text-to-image generation.}
\label{app:t2i-component-count}
Our text-to-image reference set contains only 301,706 images.
At $K=16$, the Inception fits have weight ratios
$\pi_{\max}/\pi_{\min}=79.61$ for joint matching and $72.31$ for
image-only matching. Their smallest components have reference
responsibility masses $N\pi_{\min}$ of only 699 and 683 in 3,200- and
2,048-dimensional feature spaces, respectively. These small components
provide insufficient sample support for stable full-covariance estimation.
At $K=4$, both weight ratios are below 6.2, and the smallest components
each have over 25,000 units of reference responsibility mass.
We therefore use $K=1+4$ for both text-to-image variants.

\Needspace{8\baselineskip}
\section{Results in Individual Evaluation Representations}
\label{app:encoder-results}

Table~\ref{tab:app-encoders} separates the training and held-out terms of
FDr$^6$ for the configurations in
Table~\ref{tab:imagenet256-system-comparison}. Each value is the
encoder's Fr\'{e}chet distance divided by its real-validation baseline.

\par\smallskip
\noindent\begin{minipage}{\linewidth}
\centering
\fontsize{8.5}{10}\selectfont
\captionsetup{type=table,font={stretch=1.0},textfont={small},labelfont={bf,color=meowPurple},skip=4pt}
\setlength{\tabcolsep}{1.3pt}
\caption{\textbf{Evaluation in training and held-out representations.}
Normalized Fr\'echet distances after 100 epochs with SIM encoders;
lower is better. MGFlow-$W_2$ uses $K=1+4$, and MGFlow-KL uses
$K=1+4+16$. FD-Loss per-encoder results for JiT-H and pMF-L/H are
taken from \citet{yang2026fdloss}; for JiT-B/L and pMF-B, these were not
reported, so we evaluate the official SIM checkpoints using 50,000
generated images.}
\label{tab:app-encoders}
\begin{minipage}[t]{0.49\linewidth}
\begin{tabular*}{\linewidth}{@{\extracolsep{\fill}}llrrrrrr@{}}
\toprule
& & \multicolumn{3}{c}{Training}
& \multicolumn{3}{c}{Held-out} \\
\cmidrule(lr){3-5}\cmidrule(l){6-8}
Model & Method & Inc. & MAE & Sig. & Conv. & DINO & CLIP \\
\midrule
JiT-B & FD-Loss & 0.60 & 4.22 & 3.34 & 1.33 & 5.13 & 18.78 \\
 & MGFlow-$W_2$ & 0.87 & 1.63 & 2.05 & 1.44 & 4.97 & 15.65 \\
 & MGFlow-KL & 0.76 & 1.75 & 2.60 & 1.22 & 4.62 & 11.26 \\
\addlinespace[0.25em]
JiT-L & FD-Loss & 0.46 & 0.66 & 1.98 & 0.87 & 2.63 & 12.91 \\
 & MGFlow-$W_2$ & 0.63 & 0.58 & 1.24 & 1.00 & 2.63 & 10.00 \\
 & MGFlow-KL & 0.59 & 0.53 & 1.50 & 0.97 & 2.35 & 5.55 \\
\addlinespace[0.25em]
JiT-H & FD-Loss & 0.45 & 0.43 & 1.68 & 0.86 & 2.10 & 10.37 \\
 & MGFlow-$W_2$ & 0.59 & 0.40 & 1.08 & 0.80 & 2.05 & 10.39 \\
 & MGFlow-KL & 0.56 & 0.33 & 1.25 & 0.77 & 1.85 & 5.09 \\
\bottomrule
\end{tabular*}
\end{minipage}\hfill
\begin{minipage}[t]{0.49\linewidth}
\begin{tabular*}{\linewidth}{@{\extracolsep{\fill}}llrrrrrr@{}}
\toprule
& & \multicolumn{3}{c}{Training}
& \multicolumn{3}{c}{Held-out} \\
\cmidrule(lr){3-5}\cmidrule(l){6-8}
Model & Method & Inc. & MAE & Sig. & Conv. & DINO & CLIP \\
\midrule
pMF-B & FD-Loss & 0.51 & 1.86 & 5.37 & 0.77 & 4.10 & 8.51 \\
 & MGFlow-$W_2$ & 0.98 & 1.78 & 3.36 & 1.19 & 3.64 & 7.24 \\
 & MGFlow-KL & 0.93 & 1.99 & 4.35 & 1.09 & 3.59 & 6.58 \\
\addlinespace[0.25em]
pMF-L & FD-Loss & 0.47 & 0.56 & 3.03 & 0.57 & 2.21 & 5.68 \\
 & MGFlow-$W_2$ & 0.71 & 0.67 & 2.00 & 0.81 & 2.03 & 4.58 \\
 & MGFlow-KL & 0.69 & 0.64 & 2.40 & 0.65 & 1.99 & 4.08 \\
\addlinespace[0.25em]
pMF-H & FD-Loss & 0.46 & 0.35 & 2.46 & 0.57 & 1.74 & 5.77 \\
 & MGFlow-$W_2$ & 0.65 & 0.45 & 1.72 & 0.78 & 1.68 & 3.70 \\
 & MGFlow-KL & 0.64 & 0.40 & 2.03 & 0.68 & 1.61 & 3.35 \\
\bottomrule
\end{tabular*}
\end{minipage}
\end{minipage}

\clearpage
\section{Qualitative ImageNet Results}
\label{app:imagenet-qualitative}

\begin{figure}[!ht]
  \centering
  \includegraphics[width=\linewidth,height=0.83\textheight,keepaspectratio]{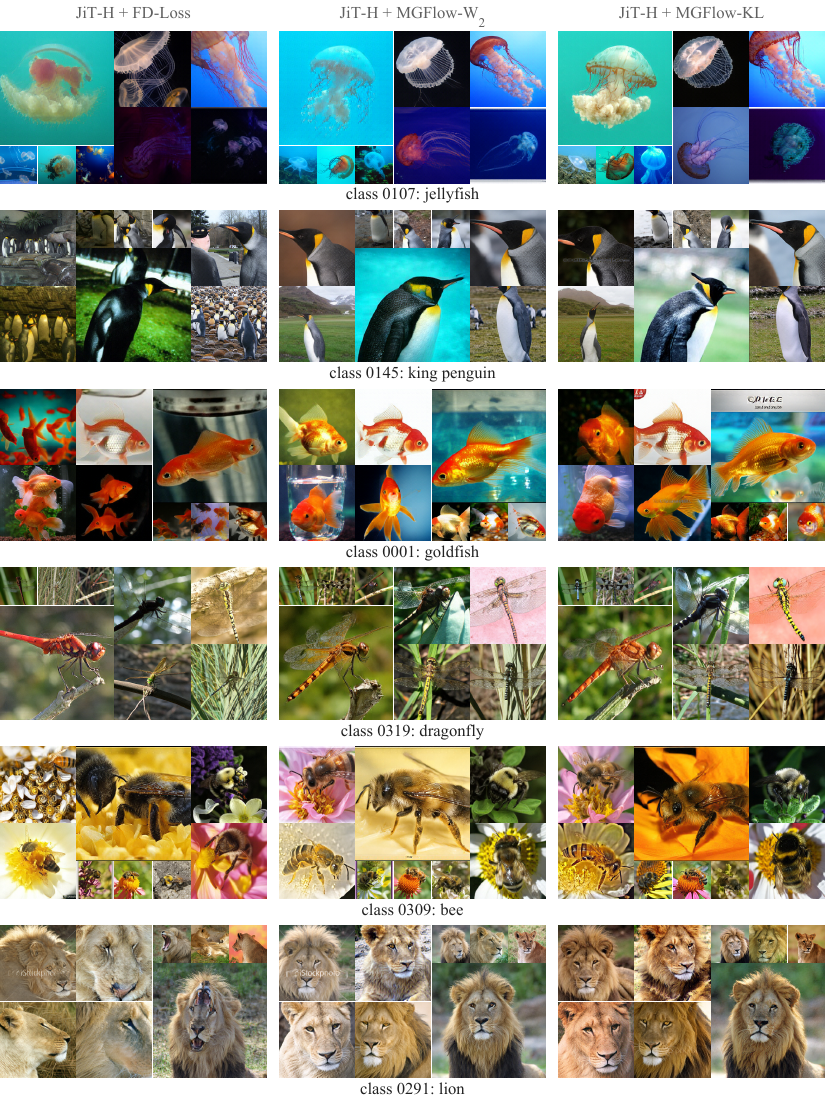}
  \caption{\textbf{Uncurated ImageNet $\bm{256\times256}$ samples from JiT-H.}
  FD-Loss~\citep{yang2026fdloss} (left), MGFlow-$W_2$ (middle), and
  MGFlow-KL (right), all with one NFE and identical input noise at
  corresponding positions.}
  \label{fig:app-jit-samples-1}
\end{figure}

\clearpage
\begin{figure}[!ht]
  \centering
  \includegraphics[width=\linewidth,height=0.83\textheight,keepaspectratio]{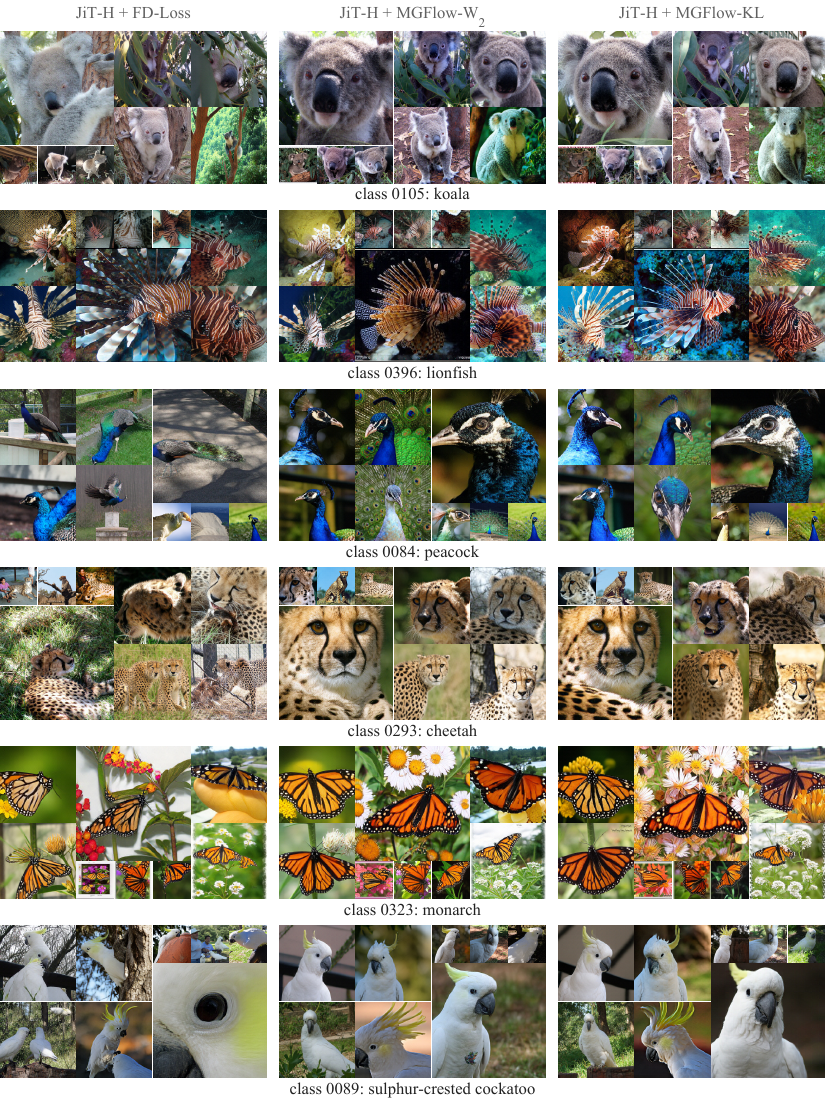}
  \caption{\textbf{Additional uncurated ImageNet $\bm{256\times256}$ samples from JiT-H.}
  FD-Loss (left), MGFlow-$W_2$ (middle), and MGFlow-KL (right), all with
  one NFE. Corresponding images use identical input noise.}
  \label{fig:app-jit-samples-2}
\end{figure}

\clearpage
\begin{figure}[!ht]
  \centering
  \includegraphics[width=\linewidth,height=0.83\textheight,keepaspectratio]{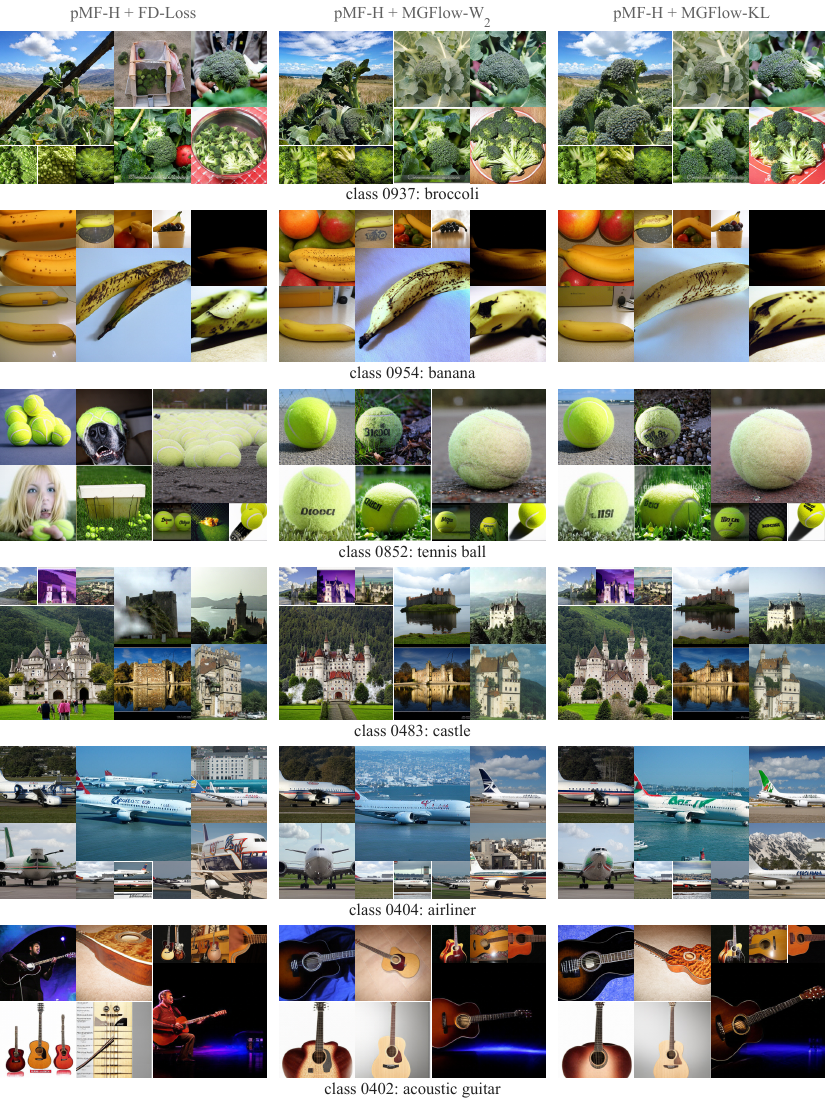}
  \caption{\textbf{Uncurated ImageNet $\bm{256\times256}$ samples from pMF-H.}
  FD-Loss~\citep{yang2026fdloss} (left), MGFlow-$W_2$ (middle), and
  MGFlow-KL (right), all with one NFE and identical input noise at
  corresponding positions.}
  \label{fig:app-pmf-samples-1}
\end{figure}

\clearpage
\begin{figure}[!ht]
  \centering
  \includegraphics[width=\linewidth,height=0.83\textheight,keepaspectratio]{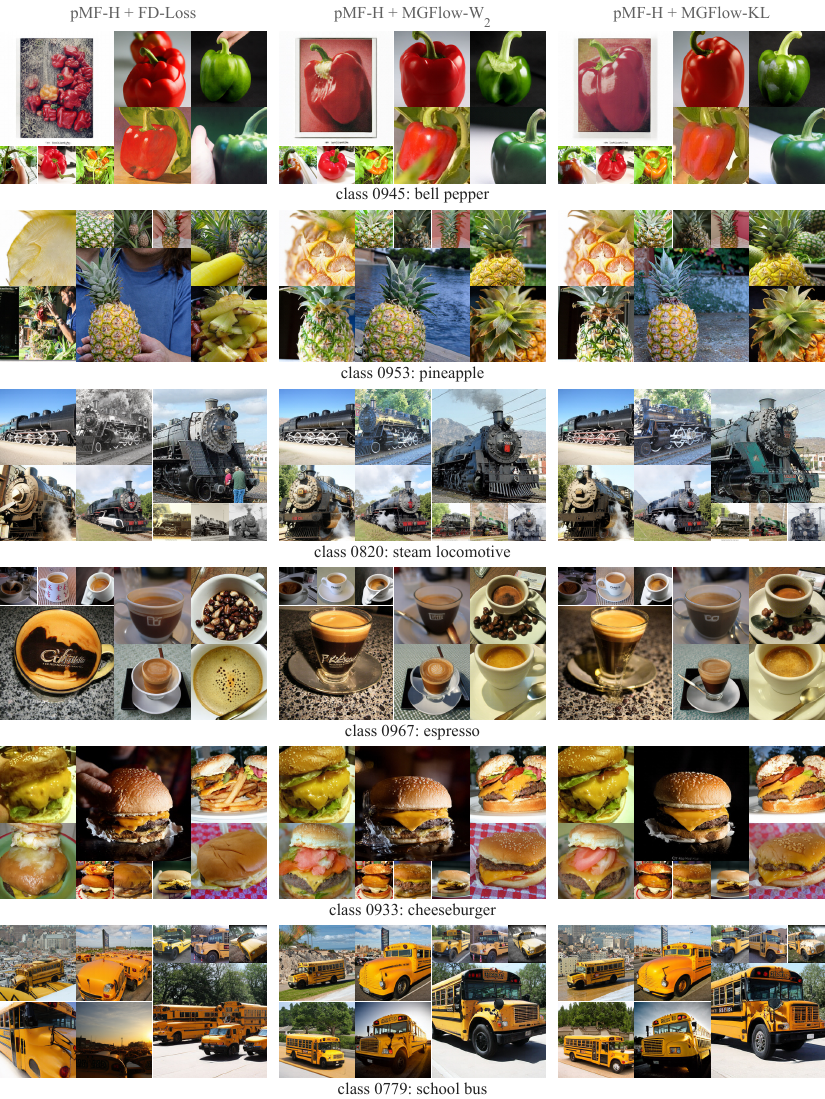}
  \caption{\textbf{Additional uncurated ImageNet $\bm{256\times256}$ samples from pMF-H.}
  FD-Loss (left), MGFlow-$W_2$ (middle), and MGFlow-KL (right), all with
  one NFE. Corresponding images use identical input noise.}
  \label{fig:app-pmf-samples-2}
\end{figure}
\clearpage

\clearpage
\section{Additional Text-to-Image Examples}
\label{app:additional-t2i-example}

\newcommand{\MGFlowTTwoICard}[2]{%
  \begin{minipage}[t]{0.325\linewidth}
    \vspace{0pt}
    \includegraphics[width=\linewidth]{#1}
    \par\smallskip
    \raggedright\fontsize{6.5}{7.5}\selectfont
    \textbf{Prompt.} #2
  \end{minipage}%
}

\begin{figure}[!ht]
  \centering
  \begin{minipage}[t]{\linewidth}
    \raggedright
    \MGFlowTTwoICard{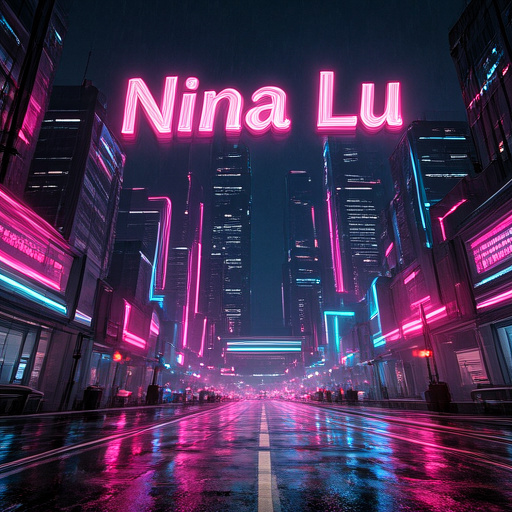}{%
      Giant glowing letters spelling ``Nina Lu'' rising above a neon-lit
      futuristic city skyline at night, low-angle wide shot, rain-slicked
      streets reflecting pink and blue light, cinematic sci-fi digital art.}%
    \hspace{0.011\linewidth}%
    \MGFlowTTwoICard{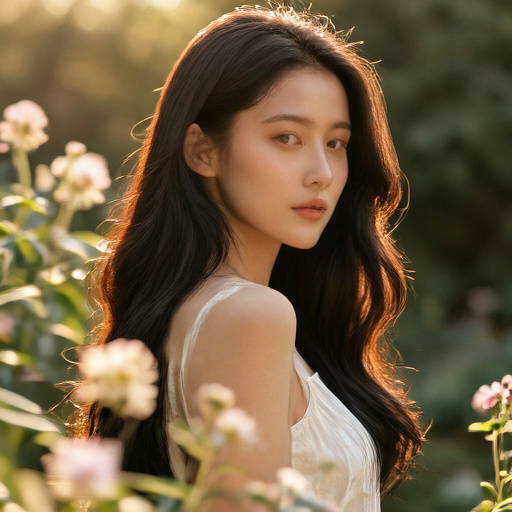}{%
      A beautiful young woman with long dark hair stands in a sunlit garden,
      soft golden light on her face, delicate flowers around her, shallow
      depth of field, elegant portrait photography.}%
    \hspace{0.011\linewidth}%
    \MGFlowTTwoICard{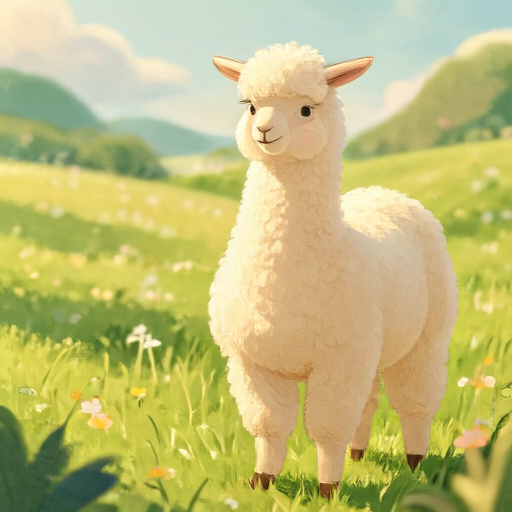}{%
      A fluffy cream-colored alpaca standing in a sunny green meadow, soft
      morning light, gentle expression, shallow depth of field, pastel
      color palette, whimsical children's book illustration style.}
    \par\vspace{6pt}\noindent
    \MGFlowTTwoICard{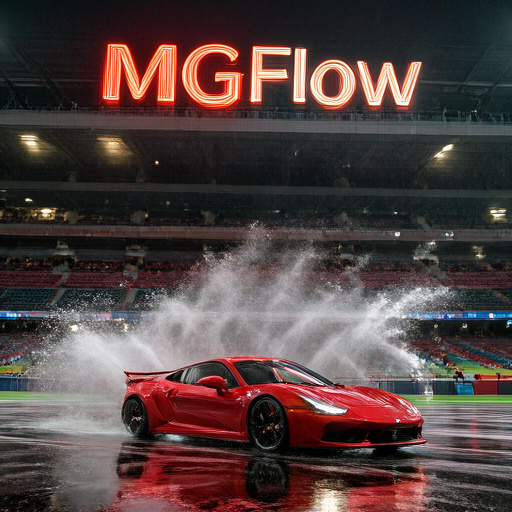}{%
      A red sports car drifts through a rain-soaked stadium beneath a
      blazing and neat MGFlow neon sign; flying spray, wet reflections,
      low-angle sports photograph}%
    \hspace{0.011\linewidth}%
    \MGFlowTTwoICard{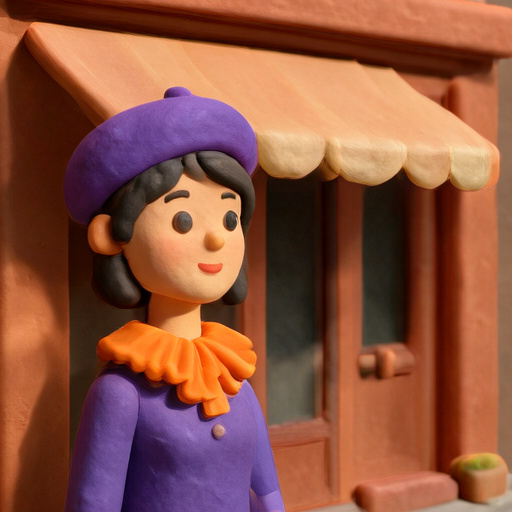}{%
      Claymation fashion portrait, woman in purple beret and orange ruffled
      collar posing beside a solid-colored awning, terracotta storefront,
      morning light casting soft shadows, medium close-up, stop-motion
      clay texture.}%
    \hspace{0.011\linewidth}%
    \MGFlowTTwoICard{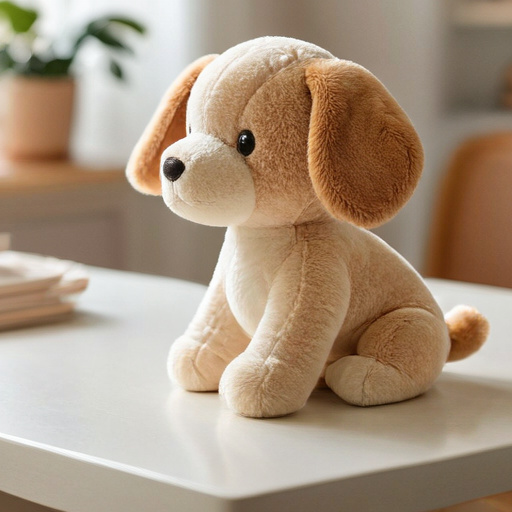}{%
      A cute plush puppy with two large floppy ears, sitting on a clean
      table, seen from a side angle, soft daylight, shallow depth of field,
      cozy still-life photograph.}
    \par\vspace{6pt}\noindent
    \MGFlowTTwoICard{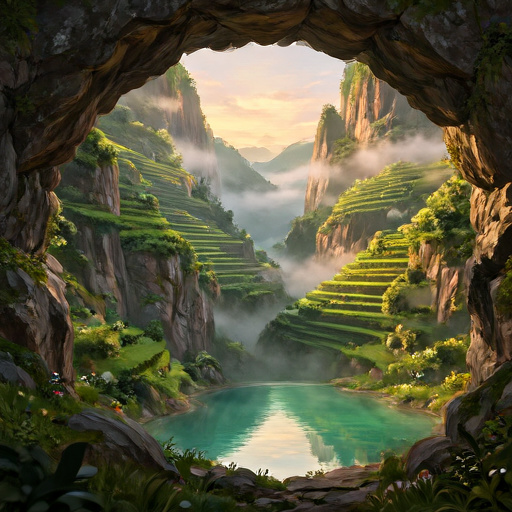}{%
      A hidden valley revealed through a rocky archway, lush green terraces,
      mist drifting between cliffs, a still turquoise pool reflecting soft
      dawn light, wide landscape view, painterly digital art.}%
    \hspace{0.011\linewidth}%
    \MGFlowTTwoICard{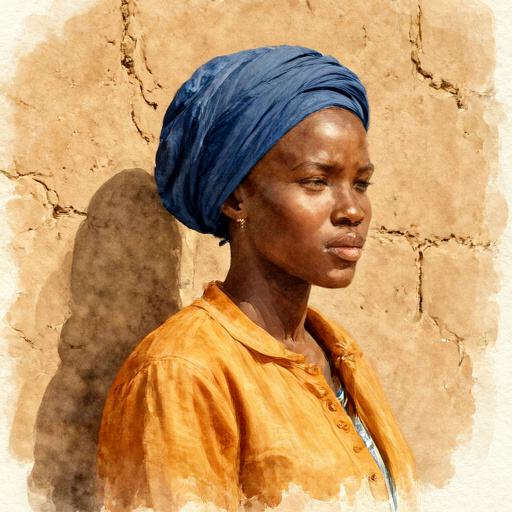}{%
      Fashion portrait in watercolor: a Somali model in an indigo headwrap
      and ochre linen jacket poses against a sunlit mud-brick wall,
      three-quarter view, soft dry-brush washes, warm dusty palette,
      loose paper texture.}%
    \hspace{0.011\linewidth}%
    \MGFlowTTwoICard{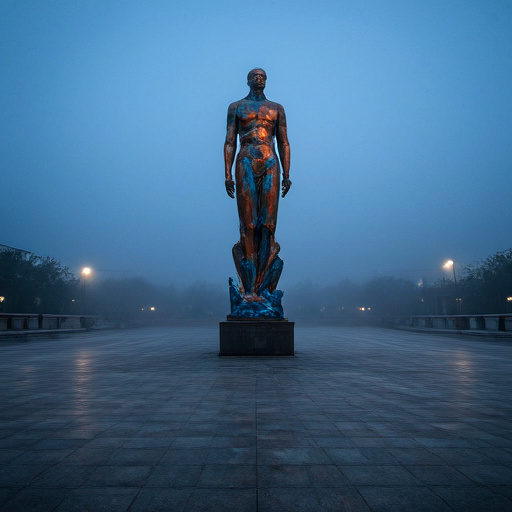}{%
      Wide cinematic still of a towering sculpture of weathered copper and
      blue resin standing in a misty deserted plaza at soft twilight,
      camera pulled far back to show the full statue small within the empty
      square, calm deep blue tones.}
  \end{minipage}
  \caption{\textbf{One-step text-to-image samples.}
  Selected $512\times512$ images from FLUX.2 [klein] 4B post-trained
  with MGFlow using joint image--text matching for 1,000 steps.}
  \label{fig:app-neon-city}
\end{figure}

\end{document}